\newif\ifconfver
\confvertrue        

\newif\ifonecoltab

\newif\ifplainver  

\ifplainver
    \confverfalse                         
\fi

\ifconfver
     \documentclass[10pt,journal]{IEEEtran}
\else
    \ifplainver
        \documentclass[11pt]{article}
        \usepackage{fullpage}
    \else
        \documentclass[12pt,draftcls,onecolumn]{IEEEtran}
    \fi
\fi

\usepackage{calc,amsfonts,amssymb,amsmath,bm,url,color,theorem,graphicx,cite,epstopdf,nicefrac,bbold}
\usepackage{psfrag,subfigure,float}
\usepackage[algoruled,linesnumbered]{algorithm2e}
\usepackage{multirow}
\usepackage[normalem]{ulem}
\usepackage{stmaryrd}
\usepackage{xcolor}
\usepackage{psfrag,framed,mdframed}
\usepackage{comment}
\definecolor{orange}{RGB}{255,107,0}

\newcommand{\X}{\boldsymbol{X}}
\newcommand{\C}{\boldsymbol{C}}

\newcommand{\U}{\boldsymbol{U}}

\newcommand{\A}{\boldsymbol{A}}
\newcommand{\B}{\boldsymbol{B}}

\renewcommand{\S}{\boldsymbol{S}}
\newcommand{\s}{\boldsymbol{s}}

\newcommand{\x}{\boldsymbol{x}}

\renewcommand{\c}{\boldsymbol{c}}
\newcommand{\y}{\boldsymbol{y}}
\renewcommand{\u}{\boldsymbol{u}}

\renewcommand{\a}{\boldsymbol{a}}

\newcommand{\T}{{\!\top\!}}

\DeclareMathOperator*{\minimize}{\textrm{minimize}}
\DeclareMathOperator*{\maximize}{\textrm{maximize}}

\DeclareMathOperator{\tr}{Tr}

\usepackage{xcolor}
\usepackage{psfrag,framed}
\usepackage{lipsum}
\usepackage{chapterbib}
\PassOptionsToPackage{normalem}{ulem}
\usepackage{ulem}
\definecolor{shadecolor}{RGB}{220,220,220}

\newtheorem{Lemma}{Lemma}
\newtheorem{Prop}{Proposition}
\newtheorem{Theorem}{Theorem}
\newtheorem{Def}{Definition}
\newtheorem{Corollary}{Corollary}

\theorembodyfont{\rmfamily}

\newtheorem{Remark}{Remark}

\definecolor{orange}{RGB}{255,107,0}

\begin{document}

\newcommand{\papertitle}{
Nonlinear Multiview Analysis:
Identifiability and Neural Network-assisted Implementation
}

\newcommand{\paperabstract}{%
Multiview analysis aims at extracting shared latent components from data samples that are acquired in different domains, e.g., image, text, and audio.
Classic multiview analysis, e.g., {\it canonical correlation analysis} (CCA), tackles this problem via matching the linearly transformed views in a certain latent domain. More recently, powerful nonlinear learning tools such as kernel methods and neural networks are utilized for enhancing the classic CCA.
However, unlike linear CCA whose theoretical aspects are clearly understood, nonlinear CCA approaches are largely intuition-driven.
In particular, it is unclear under what conditions the shared latent components across the {views} can be identified---while identifiability plays an essential role in many applications. In this work, we revisit nonlinear multiview analysis and address both the theoretical and computational aspects. {Our work leverages a useful nonlinear model, namely, the post-nonlinear model, from the nonlinear mixture separation literature. Combining with multiview data, we take a nonlinear multiview mixture learning viewpoint}, which is a natural extension of the classic generative models for linear CCA.
From there, we derive a learning {criterion}. We show that minimizing this criterion leads to identification of the latent shared components up to certain ambiguities, under reasonable conditions. Our derivation and formulation also offer new insights and interpretations to existing deep neural network-based CCA formulations. On the computation side, we propose an effective algorithm with simple and scalable update rules. A series of simulations and real-data experiments corroborate our theoretical analysis.
}


\ifplainver

    \date{\today}

    \title{\papertitle}

    \author{
    Qi Lyu and Xiao Fu\\
    School of Electrical Engineering and Computer Science\\
    Oregon State University\\
    Email: (lyuqi, xiao.fu)@oregonstate.edu
    }

	\date{}

    \maketitle

\else
    \title{\papertitle}

    \ifconfver \else {\linespread{1.1} \rm \fi

\author{Qi Lyu and Xiao Fu
	
	\thanks{

    	This work is supported in part by National Science Foundation under Projects NSF ECCS 1608961 and ECCS 1808159, and in part by the Army Research Office (ARO) under projects ARO W911NF-19-1-0247 and ARO W911NF-19-1-0407.
		
		Q. Lyu and X. Fu are with the School of Electrical Engineering and Computer Science, Oregon State University, Corvallis, OR 97331, United States. email: (lyuqi, xiao.fu)@oregonstate.edu

	}
}

\maketitle

\ifconfver \else
\begin{center} \vspace*{-2\baselineskip}
\end{center}
\fi

\begin{abstract}
	\paperabstract
\end{abstract}

\begin{IEEEkeywords}\vspace{-0.0cm}%
	Unsupervised learning, multiview analysis, neural networks, identifiability, {post-nonlinear mixture model}
\end{IEEEkeywords}

    \ifconfver \else \IEEEpeerreviewmaketitle} \fi

 \fi

\ifconfver \else
    \ifplainver \else
        \newpage
\fi \fi
\section{Introduction}
Multiview analysis has been an indispensable tool in statistical signal processing, machine learning, and data analytics.
In the context of multiview learning, a view can be understood as measurements of data entities (e.g., a cat) in a certain domain (e.g., text, image, and audio). 
Most data entities naturally appear in different domains. Multiview analysis aims at extracting essential and common information from different views.
Compared with single-view analysis tools like {\it principal component analysis} (PCA) {\cite{wold1987principal}}, \textit{independent component analysis} (ICA) \cite{Comon1994}, and \textit{nonnegative matrix factorization} (NMF) {\cite{paatero1994positive,fu2018nonnegative},} 
multiview analysis tools such as {\it canonical correlation analysis} (CCA) \cite{hotelling1936relations} have an array of unique features. For example, CCA has been shown to be more robust to noise and view-specific strong interference \cite{bach2005probabilistic,ibrahim2019cell}.

The classic CCA has been extensively studied in the literature, ever since its proposal in statistics in the 1930s {\cite{hotelling1936relations,hardoon2004canonical}}.
In a nutshell, the classic CCA seeks linear transformations for the views. The transformations are supposed to `project' the views to a domain where the views share similar representations.
Interestingly, the formulated optimization problem, although being nonconvex, can be recast into a generalized eigendecomposition problem and solved efficiently {\cite{anderson1962introduction}}. 
{Since the 1990s,} many attempts have been made towards scaling up classic CCA to handle big datasets; {see  \cite{wang2016efficient,ma2015finding,ge2016efficient,lu2014large,golub1995canonical}.}
Beyond the classic two-view CCA, a series of generalized CCA (GCCA) formulations {\cite{kettenring1971canonical,carroll1968generalization}} for handling more views exist, whose scalable versions have also been considered;  {see, e.g., \cite{fu2017scalable,fu2018efficient,kanatsoulis2018structured}.}

Linear transformation-based CCA/GCCA algorithms are elegant in computation. However, from a modeling viewpoint, restricting the transformations to be linear makes the `modeling power' limited.
For decades, a lot of effort has been invested to extending the CCA/GCCA ideas to the nonlinear regime---via incorporating a variety of nonlinear transformations. For example, kernel CCA has been popular since the 2000s \cite{fukumizu2007statistical,bach2002kernel}.
More recently, together with the success of deep learning, deep neural networks are also used to enhance CCA for unsupervised representation learning \cite{wang2015deep,andrew13deep}.
Compared to kernel methods, deep neural networks are considered more flexible and more scalable.

Kernel CCA and deep CCA have demonstrated effectiveness in many real-world applications, e.g., image representation learning \cite{wang2015deep} and speech processing \cite{wang2015unsupervised}.
This is encouraging---it shows that incorporating nonlinearity in data analytics is indeed well-motivated. On the other hand, it is still largely unclear under what conditions these methods will work (fail) or how to improve existing schemes with theoretical supports.
In fact, unlike linear CCA whose generative models, parameter identifiability {properties}, and computational aspects are fairly well understood, nonlinear CCA formulations are largely intuition-driven. 

In this work, we revisit the nonlinear multiview analysis problem, and offer both theoretical understanding and theory-backed implementation.
Our work is motivated by a recent work in linear CCA \cite{ibrahim2019cell}, where different views are modeled as mixtures of shared components and view-specific components (which are interference).
This model is similar to the one considered in machine learning in the context of probabilistic CCA \cite{bach2005probabilistic}.
Under this model, interesting interpretation for the effectiveness of CCA can be obtained. Specifically, the work in \cite{ibrahim2019cell} shows that the classic linear CCA can extract the shared components' range space---even if the interference terms are much stronger than the shared components.
The model is fairly simple and succinct, yet the insight is significant: it offers an interesting interpretation for the effectiveness of CCA for extracting shared information across views under possibly strong interference.

Building upon the intriguing perspectives in \cite{ibrahim2019cell}, we take a step further and consider the following problem: If the acquired views are mixtures of shared and view-specific components distorted by {\it unknown nonlinear functions}, is it still possible to extract the same shared information as in the linear case?
This problem is well-motivated, since a large variety of acquired real data are subject to \textit{unknown} nonlinear distortion effects. {For example, in array processing, satellite and microwave communications, and biological signal processing, nonlinear distortions often happen; see the discussion in \cite{taleb1999source,jutten2004advances,Common2010} and the references therein.}

However, taking into consideration unknown nonlinear distortions makes the problem of interest much more challenging---both in theory and practice.  Nonlinearity removal from mixture models was considered under some limited cases, which are mostly single-view analysis problems. Some notable ones are 1) \textit{nonlinear independent component analysis} (nICA)\cite{hyvarinen1999nonlinear,hyvarinen2016unsupervised,taleb1999source,achard2005identifiability,oja1997nonlinear}, where the components of interest are statistically independent random processes, and 2) \textit{nonlinear mixture learning} (NML) \cite{yang2019learning}, where the sought latent components reside in a confined manifold, i.e., the probability simplex.
In both cases, the assumptions on the components of interest are leveraged on to come up with identification criteria.
However, both statistical independence and the probability-simplex type structure are considered special assumptions, which may not hold in general. In addition, both nICA and NML assume ``clean'' mixtures without interference present.  With multiple views available, can we circumvent these strong assumptions like independence or the probability simplex structure? More importantly, is fending against non-interesting strong interference components still possible under nonlinear settings?

\smallskip

\noindent
{\bf Contributions} Bearing these questions in mind, we address both the analytical and computational aspects of nonlinear multiview analysis.
Our detailed contributions are as follows:

\noindent
$\bullet$ {\bf Model-Based Formulation and Analysis.} We propose a multiview nonlinear mixture model that is a natural extension of the mixture model-based linear multiview analysis \cite{ibrahim2019cell,bach2005probabilistic}. 
To be specific, we model each view as a nonlinear mixture of shared and view-specific interference components, where the nonlinear distortions are unknown continuous invertible functions {imposed on each of the received components}.
{This distortion model is reminiscent of the {\it post-nonlinear mixture} (PNL) model in the nonlinear blind source separation literature. Note that the PNL model is a special nonlinear model, yet is considered effective in modeling nonlinearity in many applications---especially applications with nonlinear distortions happening on the sensor end \cite{jutten2004advances,taleb1999source,achard2005identifiability}.}
We propose an identification criterion for extracting the \textit{shared} information across views. We show that solving the formulated problem implicitly removes the unknown nonlinearity up to trivial ambiguities---making the nonlinear multiview analysis problem boil down to a linear CCA problem. This means that our formulation enjoys the same identifiability properties as {those} in the linear case \cite{ibrahim2019cell}, despite of working under a much more challenging scenario.

\noindent
$\bullet$   {\bf Neural Network-Based Algorithm Design.} Based on our formulation, we propose a neural network based implementation. The formulated optimization surrogate is delicately designed to realize the identification criterion in practice. In particular, possible trivial solutions revealed in the analysis are circumvented via a careful construction of the optimization objective and constraints. Based on the formulation, we propose a simple block coordinate descent (BCD) algorithm. The proposed implementation is compatible with existing popular neural network architectures (e.g., \textit{convolutive neural networks} (CNN) and fully connected neural networks \cite{lecun2015deep}) and is scalable for handling big data.

\noindent
$\bullet$ {\bf Extensive Experiments.} We test our method in a number of simulations under different scenarios to validate the identifiability theory and to showcase the effectiveness of the implementation. In addition, a couple of real datasets (i.e, a multiview brain imaging dataset and a multiview handwritten digit dataset) are employed to demonstrate the usefulness of the proposed approach for handling real-world problems.

{Part of the work will be presented at IEEE SAM 2020 \cite{lyu2020multiview}. The journal version additionally contains derivations of the theoretical claims, detailed algorithm design and implementation, more comprehensive simulations, and more real experiments.}

\noindent {\bf Notation} We largely follow the established conventions in signal processing. To be specific, we use $x,\bm x,\bm X$ to represent a scalar, vector, and matrix, respectively. $\|\cdot\|$ denotes the Euclidean norm, i.e., $\|\bm{x}\|_2$ and $\|\bm{X}\|_F$; $\|\cdot\|_0$ denotes the $\ell_0$ {pseudo-norm} which counts the number of nonzero elements of its {argument}; $^\top$ and $^\dagger$ denote the transpose and Moore-Penrose pseudo-inverse operations, respectively; the shorthand notation $f$ is used to denote a function $f(\cdot):\mathbb{R}\rightarrow \mathbb{R}$;
$f'$ and $f''$ denote the first-order and second-order derivatives of the function $f$, respectively; $\tr(\cdot)$ denotes the trace operator; $\bm{1}$ denotes an all-one vector with a proper length; $\bm{I}$ denotes an identity matrix with a proper size; $f\circ g$ denotes the function composition operation; {${\cal X}\times {\cal Y}$ means the Cartesian product of the sets ${\cal X}$ and ${\cal Y}$.}

\section{Background}
In this section, we briefly introduce some preliminaries pertaining to our work.
\subsection{Mixture Models}
Mixture models have proven very useful in signal processing and machine learning. 
The simplest {\it linear mixture model} (LMM) can be written as:
\begin{equation}\label{eq:lmm}
\y_\ell =\A\s_\ell,~\ell=1,2,\ldots,N,
\end{equation}
where $\y_\ell\in\mathbb{R}^M$ denotes the $\ell$th observed signal,
$\A\in\mathbb{R}^{M\times K}$ the mixing system,
and $\s_\ell\in\mathbb{R}^K$ the $K$ latent components (or, the ``sources'') measured at sample $\ell$.
Many applications are concerned with identifying $\A$ and/or $\s_\ell$'s from the observed $\y_\ell$'s; {see \cite{Common2010}}.
If both the mixing system and the sources are unknown, estimating $\A$ and $\s_\ell$ simultaneously poses a very hard problem---which is known as the {\it blind source separation} (BSS) problem; {see, e.g., \cite{Common2010,fu2016robust,fu2015blind,Lathauwer2007,cruces2010bca,zibulevsky2001blind}}. The BSS problem is ill-posed, since in general the model is not identifiable; i.e., even if there is no noise, one could have an infinite number of solutions that satisfy \eqref{eq:lmm}. 
Nonetheless, identifiability can be established via exploiting properties of $\s_\ell$. For example, the seminal work of ICA \cite{Comon1994} shows that the identifiability of $\s_\ell$ can be established (up to scaling and permutation ambiguities) leveraging on statistical independence between the elements of $\s_\ell$.
Later on, identifiability of $\A$ and/or $\s_\ell$ were established through exploiting other properties of the latent components, e.g., convex geometry \cite{fu2016robust,fu2018identifiability}, quasi-stationarity \cite{SOBIUM}, spectral diversity \cite{SOBI1997}, sparsity \cite{zibulevsky2001blind}, and boundedness \cite{cruces2010bca}.

Beyond the LMM, \textit{nonlinear} mixture models (NMMs) have also attracted much attention---since nonlinearity naturally happens in practice for numerous reasons.
For example, starting from the 1990s, a line of work named nonlinear ICA \cite{hyvarinen1999nonlinear,taleb1999source,achard2005identifiability,oja1997nonlinear} considered the following model:
\begin{equation}\label{eq:nmm}
\y_\ell = \bm g(\s_\ell),~~\ell=1,2,\ldots,N,
\end{equation}
where $\bm g(\cdot):\mathbb{R}^{K}\rightarrow \mathbb{R}^M$ is a nonlinear mixing system. While it was shown that \eqref{eq:nmm} is in general not identifiable,
the so-called \textit{post-nonlinear} (PNL) model can be identified, again under the premise that $s_{k,\ell}$ and $s_{j,\ell}$ are statistically independent random processes \cite{taleb1999source,achard2005identifiability,ziehe2003blind}. To be specific, the PNL model is defined as
\begin{equation}\label{eq:pnmm}
\y_\ell = \bm g(\A\s_\ell),~~\ell=1,2,\ldots,N,
\end{equation}
with $\bm g(\cdot)=[g_1(\cdot),\ldots,g_M(\cdot)]^\T$ and $g_m(\cdot):\mathbb{R} \rightarrow \mathbb{R}$ being a univariate nonlinear function.
This model, although less general relative to \eqref{eq:nmm}, is still very meaningful---which finds applications in cases where unknown nonlinear effects happen \textit{individually} at multiple sensors/channels, e.g., in brain signal processing \cite{ziehe2000artifact,oveisi2012nonlinear} and hyperspectral imaging \cite{dobigeon2014nonlinear}.
Under the PNL model, $\s_\ell$ was first shown to be identifiable via exploiting statistical independence \cite{taleb1999source,achard2005identifiability}.
Recently, the work in \cite{yang2019learning} proved that even if the sources are dependent, some other properties, e.g., nonnegativity and a sum-to-one structure (i.e., $\bm 1^\T\s_\ell =1$), can be exploited to identify $\s_\ell$ under the PNL model.

\subsection{Multiview Data and Mixture Learning}
In practice, data representing the same entities are oftentimes acquired in different domains---leading to the so-called multiview data. Multiview data has been frequently connected to mixture models, since it is believed that multiple views of the same data sample have certain shared but latent components. In addition, it is commonly believed that using multiple views may have advantages over using a single view for learning problems, e.g., for combating noise and strong interference---since more information is available.

One way of modeling multiview data is using the following linear model \cite{bach2005probabilistic}:
	\begin{align}\label{eq:bach}
	\y_\ell^{(q)} \approx \A^{(q)}\s_\ell,~q=1,2,\ldots,Q,
	\end{align}
where $q$ is the index of the view and $Q$ is the number of available views. 
In this work, we mainly consider $Q=2$. But the techniques can be readily extended to cover the $Q\geq 3$ case.
From the above, one can see that $\s_\ell$ can be used to model the latent shared components across different views, while $\A^{(q)}$ is the basis of the subspace where the $q$th view is observed. 
In other words, the different appearances of the views are caused by the differences of the observation subspaces, while the latent representations of the views are identical.
{It was shown in \cite{bach2005probabilistic} that CCA is a maximum likelihood estimator (MLE) for ${\rm range}(\A^{(q)})$ under the model $\y_\ell^{(q)} = \A^{(q)}\s_\ell +\bm v_\ell^{(q)}$, where $q=1,2$, if the noise $\bm v_\ell^{(q)}$ is Gaussian---even if the noise is \textit{colored} and {its} covariance is \textit{unknown}. This spells out the benefits of using a CCA-like learning criterion. If one concatenates the views (i.e., to form $\bm y_\ell=[(\y_\ell^{(1)})^\T,(\y_\ell^{(2)})^\T]^\T$ and applies single-view tools like PCA, it is MLE only when the noise follows the i.i.d. zero-mean \textit{white} Gaussian distribution.}
{Similar models have also been studied in the context of joint blind source separation (JBSS) and multiview biological signal analysis \cite{li2009joint,correa2008canonical}. Unlike the model in \cite{bach2005probabilistic}, in \cite{li2009joint}, the latent components in each view are not exactly the same, but closely correlated across views.}
In the recent work \cite{ibrahim2019cell}, the above model in \eqref{eq:bach} is further developed to incorporate view-specific components, where we have:
\begin{align}\label{eq:salah}
	\bm{y}_\ell^{(q)} =\bm{A}^{(q)}[\bm{s}_\ell^\T,(\bm{c}^{(q)}_\ell)^\T]^\T,~\ell=1,\ldots,N
\end{align}
where $\A^{(q)}\in\mathbb{R}^{M_q\times (K+R_q)}$, $ \s_\ell\in\mathbb{R}^K$ collects the \textit{shared components} and $\c^{(q)}_\ell\in\mathbb{R}^{R_q}$ {are} the view-specific components.
This model is plausible, since it gives more flexibility for modeling the cross-view disparities. 
More importantly, it explains the enhanced robustness of multiview analysis against the single view ones, e.g., PCA, for extracting essential shared information ({i.e., $\bm s_\ell$ up to certain ambiguities}) about the data. 
More specifically, consider the following linear CCA formulation \cite{hardoon2004canonical,fu2017scalable}:
	\begin{align}
	\maximize_{\B^{(q)}\in \mathbb{R}^{K\times M_q}}&~ {\rm Tr}\left( \frac{1}{N}\sum_{\ell=1}^N \B^{(1)}\y_\ell^{(1)}\left( \B^{(2)}\y_\ell^{(2)}\right)^\T\right)\\
	{\rm subject~to}&~ {\frac{1}{N}\sum_{\ell=1}^N \B^{(q)}\y_\ell^{(q)}(\y_\ell^{(q)})^\T(\B^{(q)})^\T =\bm I},~q=1,2 \nonumber
	\end{align}
where $\bm B^{(q)}\in\mathbb{R}^{K\times M_q}$\footnote{Note that in the literature, the size of $\B^{(q)}$ can be $R\times M_q$, where $R\in\{1,\ldots,K\}$. For example, setting $R=1$ means that one aims at extracting one shared component, which is similar to the convention in PCA. Nevertheless, for notational simplicity, we set $R=K$ in this paper. The proofs and results can be trivially extended to cover the general case $\bm B^{(q)}\in\mathbb{R}^{R\times M_q}$ where $1\leq R\leq K$.}
and the constraints are employed to avoid degenerate solutions.
CCA aims to find the most correlated linear projections of both views.
Note that CCA was not developed under the particular multiview mixtures in \eqref{eq:salah}. However, it was shown in \cite{ibrahim2019cell} that, under the model in Eq.~\eqref{eq:salah}, solving the above can identify $\widehat{\x}_\ell = \bm \Theta\bm s_\ell$, where $\bm \Theta\in\mathbb{R}^{K\times K}$ is a nonsingular matrix---no matter how strong the view-specific components $\bm c^{(q)}_\ell$ are---{the proof in \cite{ibrahim2019cell} is restated in Appendix~\ref{app:linearcca} using notations of this paper, for ease of exposition}.
On the contrary, PCA always extracts the energy-wise strongest components.

To understand the result in \cite{ibrahim2019cell}, it might be {the} best to view the CCA cost function in its equivalent form as follows \cite{hardoon2004canonical}:
\begin{align}\label{eq:cost_cca}
&\minimize_{\B^{(q)}\in \mathbb{R}^{R\times M_q}} \sum_{\ell=1}^N \left\|\B^{(1)}\y_\ell^{(1)}   - \B^{(2)}\y_\ell^{(2)}\right \|_2^2\\
&	{\rm subject~to}~ {\frac{1}{N}\sum_{\ell=1}^N \B^{(q)}\y_\ell^{(q)}(\y_\ell^{(q)})^\T(\B^{(q)})^\T =\bm I},~q=1,2. \nonumber
\end{align}
One can check that 
$ \B^{(q)} = [\bm \Theta,\bm 0] (\A^{(q)})^\dagger \in\mathbb{R}^{K\times M_q}$
makes the cost in \eqref{eq:cost_cca} zero under the model in \eqref{eq:salah}. The work in \cite{ibrahim2019cell} further shows that this type of solution is unique up to an arbitrary non-singular $\bm \Theta$.
Since $\bm B^{(q)}\bm y_\ell^{(q)}= \bm \Theta\bm s_\ell$, this means that the range space spanned by the shared components $\S^\T$ (where $\S=[\s_1,\ldots,\s_N]$) can be extracted.
Note that the resulting projected views, i.e., $$\widehat{\x}_\ell =\bm B^{(q)}\bm y_\ell^{(q)}= \bm \Theta\bm s_\ell,~\ell=1,\ldots,N,$$
are again LMMs, and many techniques introduced in the previous section can be used to identify $\s_\ell$---with the interference components $\c_\ell^{(q)}$ having been eliminated by CCA.

\subsection{Nonlinear Multiview Learning}
As in the single-view case, nonlinearity is natural to be considered in the multiview case.
For example, nonlinear learning tools such as the kernel method and deep learning were combined with CCA, where interesting results were observed \cite{bach2002kernel,lopez2014randomized,wang2015deep,andrew13deep}.
For example, the idea of deep CCA is to employ deep neural networks, instead of the linear operators $\bm B^{(q)}$'s as in the classic CCA, to perform data transformation. The deep CCA formulation in \cite{andrew13deep} is based on maximizing the correlation between
${\bm f}^{(q)}(\bm y^{(q)}_\ell)$'s
across views, where $\bm f^{(q)}(\cdot)$ is a deep neural network-represented nonlinear transformation.
Nevertheless, these works mostly focus on practical implementations, rather than theoretical aspects. It is unclear if the interesting identifiability results as in the classic multiview mixture model in \cite{ibrahim2019cell} still hold when unknown nonlinearity is imposed onto the views. 
This is the starting point of our work.

\section{Proposed Approach}

\subsection{A Nonlinear Multiview Model}
In this section, we propose a nonlinear multiview analysis method that aims at learning shared information from the following model:
\begin{align}\label{eq:generative}
\y_\ell^{(q)} =\bm g^{(q)}(\A^{(q)}\bm z_\ell^{(q)}),~q=1,2,
\end{align}
where $\bm z_\ell^{(q)}=[\bm{s}_\ell^\T,(\bm{c}^{(q)}_\ell)^\T]^\T$, the shared components are uncorrelated zero-mean {stationary} random processes, i.e., $$\mathbb{E}[s_{i,\ell}]=0,~\mathbb{E}\left[ \s_\ell\s_\ell^\T \right]=\bm \Sigma_s:~{\rm diagonal},$$ the function
$\bm g^{(q)}(\cdot)=[g_1^{(q)}(\cdot),\ldots,g_{M_q}^{(q)}(\cdot)]^\T$ and $g_m^{(q)}(\cdot):\mathbb{R} \rightarrow \mathbb{R}$ represents the view specific nonlinear distortion at channel $m$. 
Note that each {$g_m^{(q)}(\cdot)$} is assumed to be \textit{invertible}---otherwise there is no hope to recover the underlying model.
Under the nonlinear setting, we also assume that
the latent variables are defined over continuous open sets ${\cal S}$ and ${\cal C}_q$, i.e.,
\begin{align}\label{eq:SC}
\s_\ell\in{\cal S},\quad \c^{(q)}_\ell\in{\cal C}_q,~q=1,2.
\end{align}
Note that this model is a natural extension of the multiview linear mixture models as in \cite{bach2005probabilistic,ibrahim2019cell} (especially the one in \cite{ibrahim2019cell}) and the PNL model in single-view nonlinear mixture analysis \cite{taleb1999source,achard2005identifiability,ziehe2003blind}.

As we have discussed, one major motivation behind the model in \eqref{eq:generative} is that in signal sensing and data acquisition, nonlinear distortions oftentimes happen at the sensor end. 
Hence, a PNL model is considered appropriate for such cases. Nonlinear effects are perhaps particularly severe for biosensors, since biological signals are hard to predict or calibrate. On the other hand, multiple views often exist in biology, e.g., electroencephalogram (EEG), magnetoencephalography (MEG), and functional magnetic resonance imaging (fMRI) measurements for the same stimuli \cite{rustandi2009integrating,penny2011statistical,fu2017brainzoom}. 
{Note that in practice, EEG, MEG, and fMRI images may contain closely correlated latent components other than sharing exactly the same latent components \cite{correa2008canonical}. Nonetheless, the proposed model may serve as a reasonable approximation.}
The proposed model in \eqref{eq:generative} can be leveraged on to handle such biosensor-acquired multiview data.

We should mention that the model in \eqref{eq:generative} could also benefit more general data analytics applications such as image/text classification. 
The model in \eqref{eq:generative} is evolved from the classic linear mixture model that has been widely used in data science for pre-processing, in particular, dimensionality reduction via computational tools such as PCA, ICA, and CCA. Adding a layer of nonlinearity can help capture more dynamics that were not modeled by the classic tools, which likely leads to better performance. This is the basic intuition behind the deep autoencoder \cite{hinton1994autoencoders} and kernel/deep CCA \cite{fukumizu2007statistical,bach2002kernel,andrew13deep}. Nonetheless, unlike deep autoencoder or deep CCA that are purely data-driven, we take a model-based route in this work to reveal the insights behind the effectiveness of nonlinear multiview learning---which may also shed some light on more principled design of learning criteria and algorithms.

\subsection{Proposed Formulation}
{Our goal is to learn
$    \x_\ell = \bm \Theta \s_\ell,~\ell=1,\ldots,N,$
with a certain nonsingular $\bm \Theta$ under the nonlinear model \eqref{eq:generative}. 
This goal is identical to that in \cite{ibrahim2019cell} under the simpler linear case [cf. Eq.~\eqref{eq:salah}]}.
Note that we do not impose strong structural assumptions on $\s_\ell$, e.g., statistical independence between $s_{r,\ell}$ and $s_{j,\ell}$ as in \cite{taleb1999source,achard2005identifiability} or the stochasticity assumption (i.e., $\bm 1^\T\s_\ell=1$,~$\bm s_\ell\geq \bm 0$) as in \cite{yang2019learning}---existing nonlinear ICA and mixture learning techniques cannot be applied to the problem of interest.

Our idea is to seek a nonlinear mapping $\bm f^{(q)}(\cdot): \mathbb{R}^{M_q}\rightarrow \mathbb{R}^{M_q}$ and a linear operator $\bm B^{(q)}$ such that the following criterion is minimized:
\begin{subequations} \label{eq:ncca}
	\begin{align}
	&\minimize_{\B^{(q)}\in \mathbb{R}^{K\times M_q},~{\bm f}^{(q)} }\sum_{\ell=1}^N\left\|\B^{(1)} {\bm f}^{(1)}\left(\y_\ell^{(1)}\right)    - \B^{(2)}{\bm f}^{(2)}\left(\y_\ell^{(2)}\right)\right \|_2^2. \nonumber\\
	&{\rm subject~to}:	\quad\quad\quad {\bm f}^{(q)}:~\text{invertible},~q=1,2,\\
	&~\frac{1}{N}\sum_{\ell=1}^N \left(\B^{(q)}{\bm f}^{(q)}\left(\y_\ell^{(q)}\right){\bm f}^{(q)}\left(\y_\ell^{(q)}\right)^\T(\B^{(q)})^\T\right)=\bm I,~ \label{eq:energy}
	\end{align}
\end{subequations}
The idea is very similar to that of the linear CCA reformulation as in \eqref{eq:cost_cca}, with nonlinearity taken into consideration.
Ideally, we wish to obtain
\begin{subequations}\label{eq:desired}
	\begin{align}
  &\B^{(q)} = [\bm \Theta,\bm 0] (\A^{(q)})^\dagger \in\mathbb{R}^{K\times M_q}, \\
  &\bm f^{(q)}(\cdot)=[f_1^{(q)}(\cdot),\ldots,f_{M_q}^{(q)}(\cdot)]^\T : ~\mathbb{R}^{M_q}\rightarrow \mathbb{R}^{M_q},     \\
  &f_m^{(q)}(\cdot) = (g_m^{(q)}(\cdot))^{-1},~m=1,\ldots,M_q,
\end{align}
\end{subequations}
where $\bm \Theta\in\mathbb{R}^{K\times K}$ is nonsingular.
The above will extract the shared row subspace. Also note that the solution in \eqref{eq:desired} is a feasible solution to Problem~\eqref{eq:ncca}, which always exists.
The key question is: Is the solution in \eqref{eq:desired} the \textit{only} solution? That is, does the formulation in \eqref{eq:ncca} have identifiability for the shared subspace spanned by the rows of $\S =[\s_1,\ldots,\s_N]$?

\subsection{Nonlinearity Removal}
To see how we approach the identifiability problem, {we consider the following problem}:
\begin{mdframed}
	\begin{subequations}\label{eq:ncca_population}
		\begin{align}
		{\rm find}&~{\B^{(1)},\B^{(2)},\bm f^{(1)},\bm f^{(2)}}\\
		{\rm s.t.}&~\B^{(1)} {\bm f}^{(1)}\left(\y_\ell^{(1)}\right) = \B^{(2)}{\bm f}^{(2)}\left(\y_\ell^{(2)}\right),  \label{eq:equal} \\
		&~\quad\quad\quad \forall \y_\ell^{(q)} =\bm g^{(q)}(\A^{(q)}\s^{(q)}_\ell),~ \s_\ell \in {\cal S},~ \c_\ell^{(q)}\in {\cal C}_q, \nonumber \\
		&~{\bm f}^{(q)}:~\text{invertible},~q=1,2,\label{eq:invert}\\
		&~\mathbb{E}\left[\B^{(q)}{\bm f}^{(q)}\left(\y_\ell^{(q)}\right){\bm f}^{(q)}\left(\y_\ell^{(q)}\right)^\T(\B^{(q)})^\top\right]=\bm I.  \label{eq:energy_popu}
		\end{align}
	\end{subequations}
\end{mdframed}
Note that the above is derived from \eqref{eq:ncca} assuming that one has uncountably infinite $\y_\ell^{(q)}$'s such that all possible values of  $\y_\ell^{(q)}$ are exhausted. 
We use an equality constraint in \eqref{eq:equal} since when there is no noise, the optimal value in \eqref{eq:ncca} should be zero under the model in Eq.~\eqref{eq:generative}---attaining zero fitting cost is equivalent to satisfying an equality constraint. Like Problem~\eqref{eq:ncca}, Problem~\eqref{eq:ncca_population} also at least admits one solution as in~\eqref{eq:desired}.

To see the reason why the formulation in \eqref{eq:ncca_population} can remove nonlinearity and identify the desired signal subspace,
let us start with a simple illustrative case where $\A^{(q)}\in \mathbb{R}^{2\times 2}$ and there is only one shared component.
This way, the generative model is simplified as follows:
\begin{align*}
\bm{y}^{(1)}_\ell &= \left[g^{(1)}_1\left(A^{(1)}_{11}s_\ell+A^{(1)}_{12}c_\ell^{(1)}\right),  g_2^{(1)}\left(A^{(1)}_{21}s_\ell+A^{(1)}_{22}c_\ell^{(1)}\right)\right]^\top\\
\bm{y}^{(2)}_\ell &= \left[g^{(2)}_1\left(A^{(2)}_{11}s_\ell+A^{(2)}_{12}c_\ell^{(2)}\right), g^{(2)}_2\left( A^{(2)}_{21}s_\ell+A^{(2)}_{22}c^{(2)}_{\ell}\right)\right]^\top.
\end{align*}
For this simplified case, the solution $\B^{(q)}$ becomes a row vector $(\bm b^{(q)})^\T$.

To proceed, we will be using the following condition:
\begin{Def}[Ubiquitously Unanchored]
Consider a collection of $d$ real-valued random components $\bm{v} =[v_{1},\ldots,v_{d}]^\T\in \mathbb{R}^{d}$, where $v_{i}$ resides in a continuous and open set ${\cal V}_i\subseteq \mathbb{R}$ such that ${\rm volume}({\cal V}_i)>0$. 
Denote $\bar{v}_j$ as any fixed value from ${\cal V}_j$.
Assume that for any $i\in\{1,\ldots,d\}$ and any $\bar{v}_j$ where $j\neq i$, the vectors
$$[\bar{v}_{1},\ldots, \bar{v}_{i-1},v,\bar{v}_{i+1},\ldots, \bar{v}_{d}]^\T,~\quad \forall v\in {\cal V}_i$$ 
{are contained in the domain of $\bm v$.} 
Then, the components in $\bm v$ are called {\it ubiquitously unanchored}.
\end{Def}

\begin{Remark}\label{rmk:geometry}
	Intuitively speaking, the {\it ubiquitously unanchored} condition means that for any fixed $v_{j}=\bar{v}_j$ where $j\in\{1,\ldots,d\}/\{i\}$, $v_{i}$ is still allowed to take all the possible values in ${\cal V}_i$---thereby being ``unanchored'' by the other components.
	Of course, if $v_{1}, \ldots, v_{d}$
	are statistically independent, then $\bm v$ satisfies this condition.
	On the other hand, the ubiquitously unanchored condition is much milder than statistical independence.
	If $\bm v$ is a continuous random vector, the ubiquitously unanchored condition boils down to that the joint probability density function (PDF) of $\bm v$ takes positive values in the domain of ${\cal V}={\cal V}_1\times \ldots \times {\cal V}_d$.
	This is much milder than requiring that the components of $\bm v$ to be statistically independent.
	For example, consider a case where $d=2$ and ${\cal V}_i=(-1,1)$ for $i=1,2$. In this case, one can have  {$v_{1} = v_{2} + n$} where $n$ is a random perturbation. Even if the perturbation admits a small variance and thus is weak (meaning that {$v_{1}$ and $v_{2}$ are highly dependent), if the PDF of $n$ admits positive values over $(-1,1)$, the components in $\bm v$ are still ubiquitously unanchored.}
\end{Remark}
\color{black}

\begin{Theorem}[The $2\times 2$ Case]\label{thm:theorem2by2}
    Assume that the elements of $\A^{(q)}\in\mathbb{R}^{2\times 2}$ are drawn from any jointly continuous distribution for $q=1,2$, and that $\bm g^{(q)}$ and the composition $\bm f^{(q)}\circ\bm g^{(q)}$ are both twice differentiable. 
    {Assume that every view admits one shared component ($s_\ell\in\mathbb{R}$) and one interference term ($c_\ell^{(q)}\in\mathbb{R}$), and 
    that the components in $[{s}_\ell,{c}^{(1)}_\ell,{c}^{(2)}_\ell]^\T$ are ubiquitously unanchored.}
    Suppose that $(\bm b^{(q)},\bm f^{(q)})$ for $q=1,2$ are feasible solutions of Problem~\eqref{eq:ncca_population}. 
	Then, we have the following holds almost surely:
	\[     h_i^{(q)}(x)=  f_i^{(q)}\circ g_i^{(q)}(x)  = \alpha_i^{(q)} x + d_i^{(q)},~\alpha_i^{(q)}\neq 0, ~\forall i.   \]
	where $d_i^{(q)}\in\mathbb{R}$ and $\alpha_i^{(q)}\in\mathbb{R}$ for $i=1,\ldots,M_q$;
	i.e., the composition is an affine function with probability one.
\end{Theorem}

\begin{IEEEproof}
	The proof is relegated to Appendix~\ref{app:thm2by2}.
\end{IEEEproof}

\smallskip

Theorem \ref{thm:theorem2by2} shows that even if we have different view-specific components for multiview data, we can always identify the shared components under mild assumptions. This property is quite appealing since no strong assumptions like statistical independence, non-negativity or simplex structure are required for $\bm s_\ell$---as {is} opposed to existing frameworks, e.g., those in \cite{yang2019learning,taleb1999source,achard2005identifiability,hyvarinen1999nonlinear,hyvarinen2016unsupervised}. In addition, there is no constraint on the energy of each component---i.e., even if the energy of the shared component is significantly smaller compared to that of the view-specific component, the proposed criterion can still recover the shared subspace. This property is the same as that of the classic linear CCA \cite{ibrahim2019cell}, which we restated in {Appendix~\ref{app:linearcca}}. 
{Theorem~\ref{thm:theorem2by2} articulates the theoretical benefits of utilizing multiple views in nonlinear component analysis. This is consistent with previous study on linear multiview analysis \cite{via2011joint,bach2005probabilistic,ibrahim2019cell}.}

In the proof of Theorem~\ref{thm:theorem2by2}, we have implicitly used the fact that there at least exists a solution $\bm b^{(q)}$ to Problem~\eqref{eq:ncca} such that $ \| \bm b^{(q)} \|_0 =2$ (i.e., $\bm b^{(q)}$ is a ``dense'' vector). In fact, this holds for more general case where $\bm B^{(q)}$ is a $K\times M_q$ matrix. We have the following proposition:

\begin{Prop}\label{prop:zeronorm}
	Assume that $\A^{(q)}\in\mathbb{R}^{M_q\times (K+R_q)}$ is drawn from any joint absolutely continuous distribution, where $M_q\geq K+R_q$. Then, there exists a solution {$\B^{(q)}\in\mathbb{R}^{K\times M_q}$} of Problem~\eqref{eq:ncca_population} that satisfies $ \|  \bm B^{(q)} \|_0 ={K}M_q, $
    with probability one. 
\end{Prop}
\begin{IEEEproof} 
	The proof is relegated to {Appendix~\ref{app:prop}}.
\end{IEEEproof}

\smallskip

With the insights gained from the $2\times 2$ case and Proposition~\ref{prop:zeronorm}, we are ready to show the nonlinear distortion identifiability under the general problem setting, i.e., under $\bm A^{(q)}\in\mathbb{R}^{M_q\times (K+R_q)}$.
We have the following theorem:

\begin{Theorem}[Nonlinearity Removal]\label{thm:general}  Consider the nonlinear model in \eqref{eq:generative}. Assume that $M_q \geq K+R_q,~q=1,2,$, that the mixing matrices $\A^{(q)}$ for $q=1,2$ are drawn from any absolutely continuous distributions, {and that the components in $[{\bm s}_\ell^\top, ({\bm c}^{(1)}_\ell)^\top, ({\bm c}^{(2)}_\ell)^\top]^\top$ are ubiquitously unanchored
}. Suppose that $(\bm B^{(q)},\bm f^{(q)})$ for $q=1,2$ are solutions of \eqref{eq:ncca_population} with $\|\bm B^{(q)}\|_0=KM_q$. 	Then, the following holds almost surely:
	\[     h_i^{(q)}(x)=  f_i^{(q)}\circ g_i^{(q)}(x)  = \alpha_i^{(q)} x + d_i^{(q)},~\alpha_i^{(q)}\neq 0,~\forall i.   \]
	where $d_i^{(q)}\in\mathbb{R}$ and $\alpha_i^{(q)}\in\mathbb{R}$ for $i=1,\ldots,M_q$;	i.e., the composition is an affine function with probability one.
\end{Theorem}
\begin{IEEEproof}
	The proof is relegated to {Appendix~\ref{app:thm}}.
\end{IEEEproof}
Note that the for the general case, we have an additional constraint $\|\bm B^{(q)}\|_0=KM_q$, which arises because the general case is more challenging in terms of analysis. 
{The condition $\|\bm B^{(q)}\|_0=KM_q$ means that $\B^{(q)}$ contains no zero entries. Note that this condition was not needed in the special $2\times 2$ case (cf. Theorem~\ref{thm:theorem2by2}). This makes the conditions stated in Theorem~\ref{thm:general} a bit more restrictive---having more latent components gives rise to a more challenging nonlinearity removal problem. }

In Theorems~\ref{thm:theorem2by2}{--}\ref{thm:general}, the shared components $s_{k,\ell}$ and $s_{j,\ell}$ can be heavily dependent---yet they cannot be completely dependent, since they have to satisfy the ubiquitously unanchored condition. However, if the dimensions of the interference terms $\c_\ell^{(q)}$ are large enough so that the views are sufficiently diverse, we show that $\bm s_\ell$ can even be completely dependent, without affecting nonlinearity removal:
\begin{Theorem}\label{thm:dependent}
Consider the nonlinear model in Eq.~\eqref{eq:generative}. Assume that $M_q \geq K+R_q,~q=1,2,$ and that the mixing matrices $\bm A^{(q)}$ for $q=1,2$ are drawn from any absolutely continuous distributions. Assume that the components in $[{s}_{k,\ell}, ({\bm c}^{(1)}_\ell)^\top, ({\bm c}^{(2)}_\ell)^\top]^\top$ are ubiquitously unanchored for any $k$. Further assume that the dimensions of the components satisfy 
\begin{align}
\frac{R_q(R_q+1)}{2}\geq M_q.    
\end{align}
Suppose that $(\bm B^{(q)},\bm f^{(q)})$ for $q=1,2$ are solutions of Eq. \eqref{eq:ncca_population} with $\|\bm B^{(q)}\|_0=KM_q.$ 	Then, the composition $f_i^{(q)}\circ g_i^{(q)}(x)$ for all $i,q$ are affine functions with probability one.
\end{Theorem}

\begin{IEEEproof} 
	The proof is relegated to Appendix~\ref{app:dependent}.
\end{IEEEproof}
\color{black}
{Under Theorem~\ref{thm:dependent}, fully dependent $s_{k,\ell}$ and $s_{j,\ell}$ are allowed to exist; also see our experiments in Sec.~\ref{sec:num}.}
{In terms of shared component extraction}, we have the following corollary:
\begin{Corollary}[Subspace Identifiability]\label{cor:subspace}
	Under the generative model \eqref{eq:generative}, assume that a feasible solution of Problem~\eqref{eq:ncca_population}, denoted by  $\B^{(q)}$ and $\bm f^{(q)}$ for $q=1,2$, can be found, where $\|\bm B^{(q)}\|_0=KM_q$. 
	Assume that $\bm s_\ell\in\mathbb{R}^K$ and $\bm c_\ell^{(q)}\in\mathbb{R}^{M_q}$ for $\ell=1,\ldots,N$ are drawn from certain jointly continuous distribution with $N\geq K+R_1+R_2 $.
	 Also assume that the learned functions satisfy that the components of $\bm B^{(q)}\bm f^{(q)}(\bm y_\ell^{(q)})$ admit zero mean.
	Then, we have
	\[                          \B^{(q)}\bm f^{(q)}(\bm y^{(q)}_\ell)=\bm \Theta\bm s_\ell,~q=1,2,  \]
	for a certain nonsingular $\bm \Theta\in\mathbb{R}^{K\times K}$ almost surely.
\end{Corollary}
\begin{IEEEproof}
The above can be shown by combining Theorem~\ref{thm:general} and the identifiability theorem of linear CCA \cite{ibrahim2019cell}, which we restate in Appendix~\ref{app:linearcca} in the supplementary materials.
\end{IEEEproof} 
We would like to remark that the assumption that $f_i^{(q)}\circ g_i^{(q)}$ has zero mean is natural, since in our model $s_{i,\ell}$ has zero mean.
In addition, from a CCA viewpoint, correlation is measured with centered data.
However, {neither} of the formulations in \eqref{eq:ncca_population} {nor} \eqref{eq:ncca} can enforce this directly---from what we have proved in Theorems~\ref{thm:theorem2by2}{--}\ref{thm:general}, the learned mapping $h_i^{(q)}$ is affine, not linear. That is, there is a constant term $d_i^{(q)}$ existing. Nonetheless, this term can be easily removed via adding a constraint in the formulation---more discussions will be seen in the next section.
{Another remark is that, linear CCA identifies $\bm \Theta\bm s_\ell$ under the model in \eqref{eq:salah}  with finite samples \cite{ibrahim2019cell}. However, our nonlinearity removal result is established under the assumption that {the equality constraint in \eqref{eq:ncca_population} holds over a continuous domain}, making the conditions in our case more stringent. Nevertheless, the proposed approach works well with large but finite $N$ {(i.e., number of samples)}, as will be seen in the experiments. Our result may also shed some light on the reason why nonlinear approaches are often in favor of ``big data''.}

\subsection{Related Works}
We should mention that there are a couple of works that are related to mixture models and nonlinear CCA.
A classic work in \cite{bach2002kernel} utilizes kernel CCA to solve the ICA problem under the linear mixture model.
The method splits $\y_\ell=\A\s_\ell$ to $\bm y_\ell^{(1)}=\A^{(1)} \s_\ell$ and $\y_\ell^{(2)}=\A^{(2)}\s_\ell$, where {$\A=[(\A^{(1)})^\T,(\A^{(2)})^\T]^\T$,} and then applies kernel CCA to match the transformed $\bm y_\ell^{(1)}$ and $\y_\ell^{(2)}$---through which $\s_\ell$-identification can be achieved. The method uses nonlinear CCA to solve the problem of interest, while the model is not nonlinearly distorted. Another work in \cite{ziehe2003blind} considers the model in \eqref{eq:pnmm}, i.e., $\y_\ell=\bm g(\A\s_\ell)$. Algorithms were derived for removing nonlinearity, again, through splitting the received signals into two parts, i.e., $\bm y_\ell^{(1)}$ and $\y_\ell^{(2)}$, and maximizing the correlation between transformed $\bm y_\ell^{(1)}$ and $\y_\ell^{(2)}$. This is effectively multiview matching as we did. Nevertheless, the work does not consider the view-specific interference terms $\bm c_\ell^{(q)}$---which is an important consideration in practice. The work also imposes strong assumptions, e.g., Gaussianity of $\x_\ell$, where $\x_\ell=\A\s_\ell$, for establishing identifiability of $\bm g(\cdot)$.

\section{Practical Implementation}
The identifiability theorems indicate that the nonlinear functions $\bm g^{(q)}(\cdot)$ in the data generation process for $q=1,2$ may be removed up to affine transformations---if one can find a solution to Problem~\eqref{eq:ncca_population}.
Problem~\eqref{eq:ncca_population} addresses the population case that requires practical approximations under finite samples and workable parametrization for $\bm f^{(q)}$ and $\bm g^{(q)}$.
In this section, we cast \eqref{eq:ncca_population} into a numerical optimization-friendly form and propose an algorithm for tackling the reformulated problem.

\subsection{Parametrization for Nonlinear Functions}
To tackle Problem~\eqref{eq:ncca_population}, we first parameterize $\bm f^{(q)}$ and $\bm g^{(q)}$ using two neural networks.
Since we seek for continuous functions, neural networks are good candidates since they are the so-called `universal function approximators'.
Note that one-hidden-layer networks can already represent all continuous functions over bounded domains to $\epsilon$-accuracy with a finite number of neurons \cite{cybenko1989approximation}.
Nevertheless, we keep our parametrization flexible to incorporate multiple layers---which have proven effective in practice \cite{lecun2015deep}.
In addition, many established network configurations and architectures can be incorporated into our framework. This is particularly of interest since multiview data oftentimes are with quite diverse forms, e.g., image and text. Using some established neural network paradigms for different types of data (e.g., CNN for image data \cite{lecun2015deep}) may enhance performance.

In our case, since we consider {$f_m^{(q)}(\cdot):\mathbb{R}\rightarrow \mathbb{R}$ and} $g_m^{(q)}(\cdot):\mathbb{R}\rightarrow \mathbb{R}$ as nonlinear functions, we parameterize them {using neural networks. For example, we use the following to approximate $g_m^{(q)}(x)$ (and the same for $f_m^{(q)}(x )$):}
\begin{align*}
&g_m^{(q)}(x) \approx (\bm w_L^{(q)})^\T \times \\
&\sigma\left( \bm W_{L}^{(q)} \ldots \sigma\left(\bm W_{2}^{(q)}\sigma\left(    \bm w_{1}^{(q)}x  + \bm \gamma_1^{(q)}  \right) +\bm \gamma_2^{(q)} \right) \ldots +\bm \gamma_{L-1}^{(q)} \right),
\end{align*} 
where $\bm w^{(q)}_1\in\mathbb{R}^{N_1}$ and $\bm w^{(q)}_L\in\mathbb{R}^{N_L}$ are the network weights of the input and output layers, respectively, $\bm W^{(q)}_{l}\in\mathbb{R}^{N_l\times N_{l-1}}$ denotes the network weights from layer $l-1$ to layer $l$, $\bm \gamma^{(q)}_l\in\mathbb{R}^{N_l}$ is the bias term of layer $l$, and $\sigma(x)$ is the so-called activation function. One typical activation function is the ${\sf sigmoid}$ function, i.e.,
$   \sigma(x) = \frac{1}{1+e^{-x}}. $
There exist many other choices, e.g., the ${\sf tanh}$ function, and the so-called \textit{rectified linear unit} ${\sf ReLU}$ function \cite{lecun2015deep}. Note that different configurations of $\bm W_l^{(q)}$ and $\sigma(\cdot)$ lead to different types of neural networks; see \cite{lecun2015deep}.

\subsection{Reformulation}
With the neural network-based parametrization {denoted as $\bm f_{\text{NN}}^{(q)}$ and $\bm g_{\text{NN}}^{(q)}$}, one can use \eqref{eq:ncca} as a working surrogate, with proper modifications.
To be specific, we consider the following optimization problem:
	\begin{align}\label{eq:opt}
	\minimize_{\bm{U},{\bm \theta_{\rm NN}},\bm{B}^{(q)}} &\sum_{q=1}^{Q}\sum_{\ell=1}^N\left\| \bm{u}_\ell-\bm{B}^{(q)}{\bm{f}_{\text{NN}}^{(q)}}\left(\bm{y}_\ell^{(q)}\right) \right\|_2^2 \nonumber \\
	&+ \lambda\sum_{q=1}^{Q}\sum_{\ell=1}^N\left\| \bm{y}_\ell^{(q)}-{\bm{g}_{\text{NN}}^{(q)}}\left({\bm{f}_{\text{NN}}^{(q)}}\left(\bm{y}_\ell^{(q)}\right)\right) \right\|_2^2\nonumber\\
	{\rm subject~to}&~ \frac{1}{N}\left[ \sum_{\ell=1}^N \bm u_\ell\bm u_\ell^\top\right]=\bm{I},\quad\frac{1}{N}\sum_{\ell=1}^N\bm{u}_\ell=\bm{0} 
	\end{align}
where $Q=2$ for the two-view case, and we introduce a slack variable $\U=[\u_1,\ldots,\u_N]\in\mathbb{R}^{K\times N}$ that represents the extracted shared components (ideally we wish $\bm u_\ell = \s_\ell$), $\bm{y}_\ell^{(q)}$ is the $\ell$th input data for the $q$th view, ${\bm{f}_{\text{NN}}^{(q)}(\cdot)=[f^{(q)}_{\text{NN},1}(\cdot),\ldots, f^{(q)}_{\text{NN},M_q}(\cdot)]^\T}$ is a neural network (NN)-parametrized element-wise non-linear mapping that we aim at learning for nonlinearity removal, and ${\bm{g}_{\text{NN}}^{(q)}(\cdot)=[g^{(q)}_{\text{NN},1}(\cdot),\ldots, g^{(q)}_{\text{NN},M_q}(\cdot)]^\T}$ is another NN-parametrized nonlinear function for learning the generative function in \eqref{eq:generative}. For conciseness, we use {$\bm \theta_{\rm NN}$} to denote the network parameters in the NNs representing ${\bm{f}_{\text{NN}}^{(q)}}$ and ${\bm{g}_{\text{NN}}^{(q)}}$.

To explain the formulation, the first term in the cost function is by lifting the constraint \eqref{eq:equal} to the cost function via introducing a slack variable $\bm U$.
This is reminiscent of the fitting formulation in \eqref{eq:ncca}. Note that introducing $\bm U$ is important, since it entails us to propose a lightweight algorithm.
The second term {is a regularization that promotes} ${\bm f_{\text{NN}}^{(q)}}$ to be invertible [i.e., to reflect the {invertibility} constraints in \eqref{eq:ncca} and \eqref{eq:ncca_population}], at least for the available data samples $\bm y_\ell^{(q)}$ for $\ell=1,\ldots,N$.
Note that if ${\bm f_{\text{NN}}^{(q)}}$ is invertible, then there exists ${\bm g_{\text{NN}}^{(q)}}$ such that the second term is zero---but the converse is not necessarily true. This is because the second term is only imposed on the available finite samples in practice, not all the possible $\bm y_\ell^{(q)}$'s.
Nonetheless, using such a data reconstruction to prevent ${\bm f_{\text{NN}}^{(q)}}$ from being an irreversible function in general works, especially when the number of samples is large. This idea is known as the \textit{autoencoder} {\cite{hinton1994autoencoders}}, which is considered a nonlinear counterpart of PCA. {Using autoencoder to promote invertibility is a simple heuristic, and can often strike a good balance between effectiveness and computational complexity.}
The term $(1/N)\U\U^\T=\bm I$ corresponds to \eqref{eq:energy}.

We should highlight the constraint 
$ (1/N)\sum_{\ell=1}^N\u_\ell=\bm 0,$
i.e., the extracted matrix $\bm U$ has a zero column mean, {which is enforced for extracting zero-mean latent components (cf. our discussions following Corollary~\ref{cor:subspace})}. 
In addition, adding this constraint is in fact quite vital for avoiding numerical problems. To see this subtle point, recall that according to {Theorems \ref{thm:theorem2by2} -- \ref{thm:general}}, we have
\begin{equation}\label{eq:trivial}
{\bm f_{\text{NN}}^{(q)}}(\bm y_\ell^{(q)})=\bm D^{(q)}\A^{(q)}\s_\ell^{(q)} +\bm d^{(q)}, 
\end{equation}    
where $\bm D^{(q)}={\rm Diag}(\alpha_1^{(q)},\ldots,\alpha_M^{(q)})$. If we directly match ${\bm f_{\text{NN}}^{(1)}}(\bm y_\ell^{(1)})$ with ${\bm f_{\text{NN}}^{(2)}}(\bm y_\ell^{(2)})$, i.e., enforcing
$\bm B^{(1)}{\bm f_{\text{NN}}^{(1)}}(\bm y_\ell^{(1)})={\bm B^{(2)}}{\bm f_{\text{NN}}^{(2)}}(\bm y_\ell^{(2)}) =\bm u_\ell, $
then, one trivial solution is to simply making $\bm D\approx \bm 0$ and $\bm d^{(1)}=\bm d^{(2)}$, with $\B^{(1)}=\B^{(2)}$---i.e., the constant $\bm d^{(q)}$ can easily dominate.
Hence, we enforce $\bm u_\ell$ to be zero mean, which will automatically take out $\bm d^{(q)}$.

\begin{Remark}
	We should mention that we do not constrain $\|\bm B^{(q)}\|_0=M_qK$ in our working formulation since it seems that $\bm B^{(q)}$ with zero entries rarely happens if $\B^{(q)}$ is randomly initialized (it never happened in our extensive experiments). Hence, incorporating such a hard constraint just for `safety' \textit{in theory} may not be worthy---a lot more complex algorithms may be required for handling this constraint. 
\end{Remark}

\subsection{Proposed Algorithm}
We propose a \textit{block coordinate descent} (BCD)-based algorithm to handle Problem~\eqref{eq:opt}.
\subsubsection{The $({{\bm \theta}_{{\rm NN}}},\bm B)$-Subproblem}
We first consider the problem of updating the neural networks when fixing $\bm U$. This is an unconstrained optimization problems and can be handled by gradient descent. Denote the loss function as
$ {\sf L}(\bm \theta,\bm U)=\sum_{\ell=1}^N {\sf L}_\ell(\bm \theta,\bm U),$
where $\bm \theta$ collects {$\bm \theta_{\rm NN}^{(q)}$} and $\bm B^{(q)}$ for $q=1,2$, and ${\sf L}_\ell(\bm \theta,\bm U) = \sum_{q=1}^{Q}\| \bm{u}_\ell-\bm{B}^{(q)}{\bm{f}_{\text{NN}}^{(q)}}(\bm{y}_\ell^{(q)}) \|_2^2  +\lambda\sum_{q=1}^{Q}\| \bm{y}_\ell^{(q)}-{\bm{g}_{\text{NN}}^{(q)}}({\bm{f}_{\text{NN}}^{(q)}}(\bm{y}_\ell^{(q)})) \|_2^2.$
At iteration $t$,  the update rule is simply
\begin{align}\label{eq:thetagrad}
\bm \theta_{t+1} \leftarrow \bm \theta_t  - \gamma_t \nabla_{\bm \theta} {\sf L}(\bm \theta_t,\bm U_t).
\end{align}
where $\gamma_t$ is the step size chosen for the $t$th update.

Note that computing $\nabla_{\bm \theta} {\sf L}(\bm \theta_t,\bm U_t)$ is normally not easy, since computing the gradient of neural networks is a resource-consuming process---the gradient normally requires \textit{backpropagation} (BP) based algorithms to compute in a sample-by-sample manner if multiple layers are involved. Hence, instead of using the full gradient to update $\bm \theta$, one can also use stochastic gradient, i.e.,
\begin{align}\label{eq:sgrad}
\bm \theta_{t+1} \leftarrow \bm \theta_t  - \gamma_t\sum_{\ell\in{\cal I}_t} \nabla_{\bm \theta}  {\sf L}_\ell(\bm \theta_t,\bm U_t),
\end{align}
where ${\cal I}_t$ is an index set randomly sampled at iteration $t$.

\subsubsection{$\bm U$-Subproblem}
To update $\bm U$, we solve the following subproblem:
\begin{align}\label{eq:Uopt}
\bm U_{t+1}\leftarrow \arg\min_{\bm{U}} &\sum_{q=1}^{Q}\sum_{\ell=1}^N\| \bm{u}_\ell-\bm{B}_{t+1}^{(q)}{\bm{f}_{t+1}^{(q)}}(\bm{y}_\ell^{(q)}) \|_2^2 \\
&\text{subject~to }~ (1/N)\bm{U}\bm{U}^\top=\bm{I},\quad \frac{\bm{U}\bm{1}}{N}=\bm{0},\nonumber
\end{align}
{where $\bm{f}_{t+1}^{(q)}$ is $\bm f^{(q)}_{\rm NN}$ parametrized by the corresponding network parameters from $\bm \theta_{t+1}$.}
This problem is seemingly difficult since it has two constraints, in which one is nonconvex.
However, this problem turns out to have a semi-algebraic solution.
{
One can see that the $\bm U$-subproblem can be re-expressed as
\begin{align}\label{eq:Uopt2}
\minimize_{\bm{U}} &\left\| \bm{U}-\frac{1}{2}\sum_{q=1}^{Q}\bm{B}_{t+1}^{(q)}\bm{F}_{t+1}^{(q)} \right\|_F^2 \\
\text{subject~to }&~ (1/N)\bm{U}\bm{U}^\top=\bm{I},\quad \frac{\bm{U}\bm{1}}{N}=\bm{0},\nonumber
\end{align}
where $\bm F^{(q)}_{t+1}=[{{\bm f}_{t+1}^{(q)}}(\bm y^{(q)}_1),\ldots,{\bm f_{t+1}^{(q)}}(\bm y^{(q)}_N)]$.
The above can be shown by expanding \eqref{eq:Uopt}:
\begin{align*}
    \sum_{q=1}^{Q}\sum_{\ell=1}^N&\| \bm{u}_\ell-\bm{B}_{t+1}^{(q)}{\bm{f}_{t+1}^{(q)}}(\bm{y}_\ell^{(q)}) \|_2^2 = \sum_{q=1}^{Q}\| \bm{U}-\bm{B}_{t+1}^{(q)}\bm F_{t+1}^{(q)} \|_F^2\\
    &=\sum_{q=1}^{Q}\left(\|\bm{U}\|_F^2+\|\bm{B}_{t+1}^{(q)}\bm F_{t+1}^{(q)}\|_F^2\right)-2\tr(\bm{U}^\top\bm{B}_{t+1}^{(q)}\bm F_{t+1}^{(q)}).
\end{align*}
Since the first and the last terms are constants, we have the following optimization problem:
\begin{align*}
    \minimize_{\begin{subarray}{c}
    	(1/N)\bm U\bm U^\T =\bm I\\\
    	\frac{\bm{U}\bm{1}}{N}=\bm{0}
    	\end{subarray}} &\sum_{q=1}^{Q} -\tr(\bm{U}^\top\bm{B}_t^{(q)}\bm F_{t}^{(q)}),
\end{align*}
which is equivalent to that in \eqref{eq:Uopt2}.
Problem~\eqref{eq:Uopt2} is seemingly hard, since it has two constraints---and one constraint is nonconvex.
Nonetheless, we show the above can be solved in closed form.
To this end, we show the following Lemma:
\begin{Lemma}\label{lem:UZ}
	Consider the following optimization problem
	\begin{subequations}\label{eq:UZ}
		\begin{align}
		\minimize_{\bm U}&~\|\bm U-\bm Z\|_F^2\\
		{\rm subject~to}&~(1/N)\bm U\bm U^\T =\bm I,~\frac{\bm{U}\bm{1}}{N}=\bm{0}.
		\end{align}
	\end{subequations}
   An optimal solution is
 $\bm U_\star =  \sqrt{N}\bm{PQ}^\top$,
   where $\bm{P}$ and $\bm{Q}$ are left and right singular vectors of $\bm{ZW}$ respectively, with $\bm{W}=\bm{I}-\frac{1}{N}\bm{11}^\top$.
\end{Lemma}
\begin{IEEEproof}
	The proof is relegated to Appendix~\ref{app:lem}.
\end{IEEEproof}

}

\bigskip
The overall algorithm is summarized in Algorithm~\ref{alo:ncca}.
One can see that the algorithm does not have computationally heavy updates.
The most resource-consuming step is the SVD used for updating $\bm U$. However, this step only takes ${\cal O}(NK^2)$ flops, which is still linear in the number of samples. Note that $K$ is the number of shared components sought, which is often small in practice.
A side note is that we also observe that one can update $\bm \theta$ multiple times until switching to the next block---which often improves the speed of convergence.

\begin{algorithm}[ht]\label{alo:ncca}
\footnotesize
 \KwData{Data $\bm{Y}^{(q)}\in \mathbb{R}^{M_q \times N}$, estimated dimension of the latent space and network structures.}
 \KwResult{$\bm{U}$ and $\bm{\theta}$}
 Initialize $\bm{U}_0 \leftarrow \sqrt{N}\bm{PQ}^\top \text{where }  \bm{PDQ}^\top=(\sum_{q=1}^Q \bm{F}_{t+1}^{(q)})\bm W$\;
 $t\leftarrow 1$;
 
 \While{stopping criterion is not reached}{

    $\bm \theta_{s,0}\leftarrow \bm \theta_t$;

   $s\leftarrow 1$;
  
   \While{stopping criterion is not reached}{

        $\bm \theta_{t,s+1} \leftarrow \bm \theta_{t,s}  - \gamma_{t,s} \sum_{\ell\in{\cal I}_{t,s}}\nabla_{\bm \theta}  {\sf L}_\ell(\bm \theta_{t,s},\bm U_t)$;
        
        $s\leftarrow s+1$;
    }

  $\bm \theta_{t+1}\leftarrow \bm \theta_{t,s}$;

  $\bm{U}_{t+1} \leftarrow \sqrt{N}\bm{PQ}^\top \text{where }  \bm{PDQ}^\top=(\sum_{q=1}^Q \bm{F}_{t+1}^{(q)})\bm W$;
  
  $t\leftarrow t+1$;

 }

 \caption{Nonlinear Multiview Component Analysis ($\texttt{NMCA}$) }\label{algo:proposed}
\end{algorithm}

\begin{Remark}
	We should mention that although our technical part was developed under $q=1,2$, the theorems and algorithm naturally hold when one has $Q\geq 3$ views.
	Another remark is that many off-the-shelf tricks for speeding up training neural networks, e.g., adaptive step size and momentum \cite{kingma2014adam}, can also be employed to determine $\gamma_{t,s}$, which can normally accelerate convergence.
\end{Remark}

\begin{Remark}\label{rmk:fully}
  In the proposed approach, we use one neural network to parameterize an ${f_{\text{NN},m}^{(q)}}$ or ${g_{\text{NN},m}^{(q)}}$. 
  Hence, $M_q$ individual networks are used to approximate ${\bm f_{\text{NN}}^{(q)}}$.
  This follows our PNL-based nonlinear signal model. In practice, {if the number of channels ($M_q$) is large, such a parametrization could be costly and adds considerable memory and computational burdens to the implementation}. One workaround is to use a single fully connected, multiple-input-multiple-output (MIMO) neural network to approximate ${{\bm f}_{\text{NN}}^{(q)}}:\mathbb{R}^{M_q}\rightarrow \mathbb{R}^{M_q}$ (and the same applies to ${{\bm g}_{\text{NN}}^{(q)}}$). This effectively means that all the individual mappings ${f_{\text{NN},i}^{(q)}}$ for $i=1,\ldots,M_q$ share neurons in the parameterization (as is opposed to each ${f_{\text{NN},i}^{(q)}}$ using an individual network). 
  Using a single fully connected network can reduce the computational burden substantially. It also sometimes outputs better results, perhaps because the associated optimization problem {is} better solved.
  Using a single MIMO network is a valid approximation for our model in \eqref{eq:ncca_population} since the MIMO neural network can also approximate any MIMO continuous function in principle; see \cite[Proposition 1]{zamzam2019data}, which is a simple extension of the universal approximation theory for multiple-input-single-output (MISO) functions \cite{cybenko1989approximation}.
\end{Remark}

\subsection{Connection to Deep CCA Approaches}
An interesting observation is that, although started from very different perspectives, the proposed formulation in \eqref{eq:opt} and the line of work, namely, deep CCA \cite{andrew13deep,wang2015deep,benton2017DeepGC} end up with similar formulations. In particular, the deep canonically correlated autoencoder (DCCAE) \cite{wang2015deep} formulation is as follows:
\begin{align}\label{eq:dccae}
\minimize_{{\bm \theta_{\rm NN}}, {\bm{B}^{(q)}}} &\sum_{\ell=1}^N\left\|\bm{B}^{(1)}{\bm{f}_{\text{NN}}^{(1)}}\left(\bm{y}_\ell^{(1)}\right)-\bm{B}^{(2)}{\bm{f}_{\text{NN}}^{(2)}}\left(\bm{y}_\ell^{(2)}\right) \right\|_2^2 \nonumber \\
&+ \lambda\sum_{q=1}^2\sum_{\ell=1}^N\left\| \bm{y}_\ell^{(q)}-{\bm{g}_{\text{NN}}^{(q)}}\left(\bm{B}^{(q)}{\bm{f}_{\text{NN}}^{(q)}}\left(\bm{y}_\ell^{(q)}\right)\right) \right\|_2^2\nonumber\\
{\rm subject~to}&~\frac{1}{N}\sum_{\ell=1}^N \left(\B^{(q)}{{\bm f}_{\text{NN}}^{(q)}}\left(\y_\ell^{(q)}\right){{\bm f}_{\text{NN}}^{(q)}}\left(\y_\ell^{(q)}\right)^\T(\B^{(q)})^\T\right)\nonumber\\ 
&\quad\quad\quad\quad\quad\quad\quad\quad=\bm I,~ q=1,2. 
\end{align}

If one changes the second term in \eqref{eq:opt} from using ${\bm{g}_{\text{NN}}^{(q)}}({\bm{f}_{\text{NN}}^{(q)}}(\bm{y}_\ell^{(q)}))$ to ${\bm{g}_{\text{NN}}^{(q)}}(\bm{B}^{(q)}{\bm{f}_{\text{NN}}^{(q)}}(\bm{y}_\ell^{(q)}))$ for reconstruction, and removing the zero-mean constraint in \eqref{eq:opt}, the two formulations are conceptually equivalent.
The seemingly subtle differences, however, are quite vital---which make DCCAE inappropriate for the model in \eqref{eq:generative}. First, the reconstruction term in DCCAE's cost function (i.e., the second term) is based on reconstructing $\bm{y}^{(q)}$ from $\B^{(q)}{{\bm f}_{\text{NN}}^{(q)}}$. This could be problematic under the model of interest in \eqref{eq:generative}, since the $\bm c_\ell^{(q)}$ part can be nullified by $\bm B^{(q)}{\bm f_{\text{NN}}^{(q)}}$---which means that such reconstruction can never be attained, making the training objective awkward. In addition, perhaps more importantly, the learned signals $\B^{(q)}{{\bm f}_{\text{NN}}^{(q)}}$ in DCCAE are not enforced to have zero mean. As we have seen in our analysis, nonlinear removal can only be done up to an affine transformation. Without the zero-mean constraint, trivial or numerically pathological solutions could easily happen [cf. the discussion in Eq.~\eqref{eq:trivial}]. We will also show this effect in the simulation.

Similarly, if one uses $Q\geq 3$ and removes the second term and the zero-mean constraints in our formulation, the formulation becomes the same as the generalized GCCA criterion in \cite{benton2017DeepGC}. 
In particular, the GCCA work in \cite{benton2017DeepGC} also employs a slack variable $\bm U$ to represent the shared latent components as in our work.
However, as we have seen in the analysis, both the reconstruction (for invertibility) and the zero-mean (for taking away constants) parts are quite critical for avoiding trivial solutions.

\section{Numerical Results}\label{sec:num}
In this section, we apply the proposed algorithm to a number of synthetic-data and real-data experiments to showcase the effectiveness of our method and to validate our analysis.

\subsection{Setup}
\noindent{\bf Baselines} We use a number of baselines throughout this section:

\noindent
$\bullet$ \textbf{Principal Component Analysis (PCA)} {\cite{wold1987principal}}: PCA is widely used for feature extraction and dimensionality reduction. The first principal component has the largest variance. 

\noindent
$\bullet$ \textbf{Canonical Correlation Analysis (CCA)} {\cite{hotelling1936relations}}: CCA finds \textit{linear} transformations of the two views that exhibit maximum correlation with each other. 

\noindent
$\bullet$  \textbf{Kernel CCA (KCCA)} \cite{lopez2014randomized}: KCCA matches the two views in a high dimensional feature space via certain nonlinear kernel functions. The baseline from \cite{lopez2014randomized} is a scalable version of KCCA.

\noindent
$\bullet$  \textbf{Deep CCA (DCCA)} \cite{andrew13deep}: DCCA employs deep neural networks to represent the nonlinear transformations of the two views.

\noindent
$\bullet$   \textbf{Deep Canonically Correlated AutoEncoders (DCCAE)} \cite{wang2015deep}: DCCAE utilizes autoencoder on top of DCCA, to avoid trivial solutions.  

\noindent
{\bf Performance Metric}: Per Corollary~\ref{cor:subspace}, the nonlinear multiview analysis recovers the range space of $\bm S^\T$, where $\bm S=[\s_1,\ldots,\s_N]$. Therefore, we employ the following {\it subspace distance} measure ${\sf dist}({\cal S},{\cal U})=\|\bm{P}_s^{\perp}\bm{Q}_{\bm{u}}^\T\|_2$ as the performance metric, where ${\cal S}={\rm range}(\S^\T)$ and ${\cal U}={\rm range}(\U^\T)$, $\bm{P}_s^{\perp}$ is defined as
$\bm{P}^{\perp}_s=\bm{I}-\bm{S}^\top(\bm{S}\bm{S}^\top)^{-1}\bm{S}$,
and $\bm{Q}_{\bm{u}}$ is the orthogonal basis of $\bm{U}$.
Note that $\|\X\|_2$ here denotes the matrix 2-norm, i.e., the largest singular value of $\X$.
The metric is bounded within $[0,1]$. 
If $\bm{U}\approx\bm \Theta\bm S$, then $\|\bm{P}_s^{\perp}\bm{Q}_{\bm{u}}^\T\|_2$ should be close to zero.

Algorithm~\ref{algo:proposed} is implemented as follows. The $\bm \theta$ update is implemented with multiple updates in each iteration (i.e., 100 times for each fixed $\U_t$). The batch size $|{\cal I}_t|$ is 1,000.
The \texttt{Adam} algorithm \cite{kingma2014adam} is employed for updating $\bm \theta$. The initial step size is set to be $10^{-3}$ in all the experiments. The algorithm is stopped after 10,000 epochs in total. Our algorithm is implemented using \texttt{PyTorch}.

\subsection{Synthetic-Data Experiments}
\begin{figure}[t]
\centering
    \includegraphics[width=0.6\linewidth]{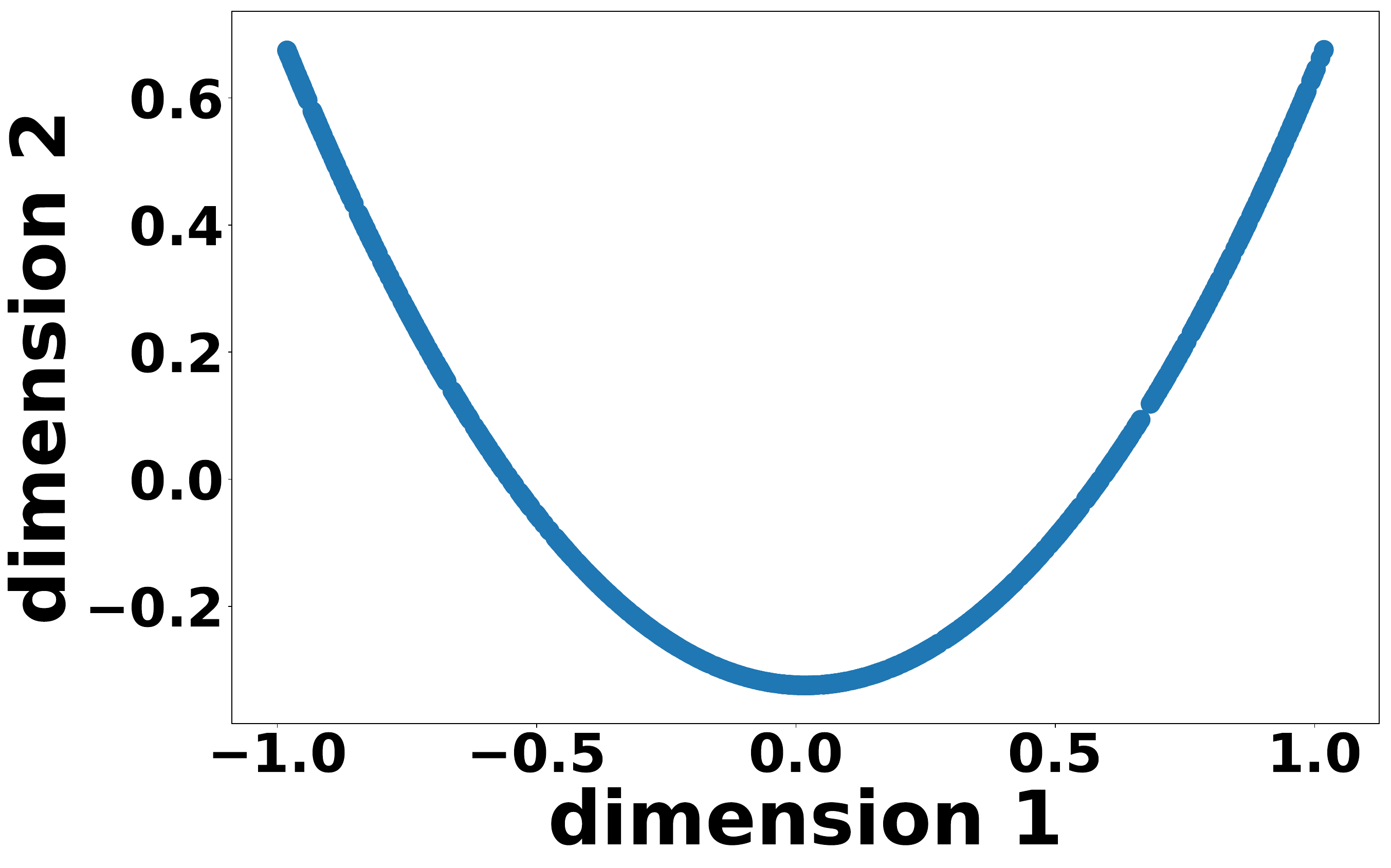}
\caption{The shared 2-dimensional components.}
\label{fig:source}
\vspace{-.5cm}
\end{figure}

\begin{figure}
\centering
    \subfigure{\includegraphics[width=0.44\linewidth]{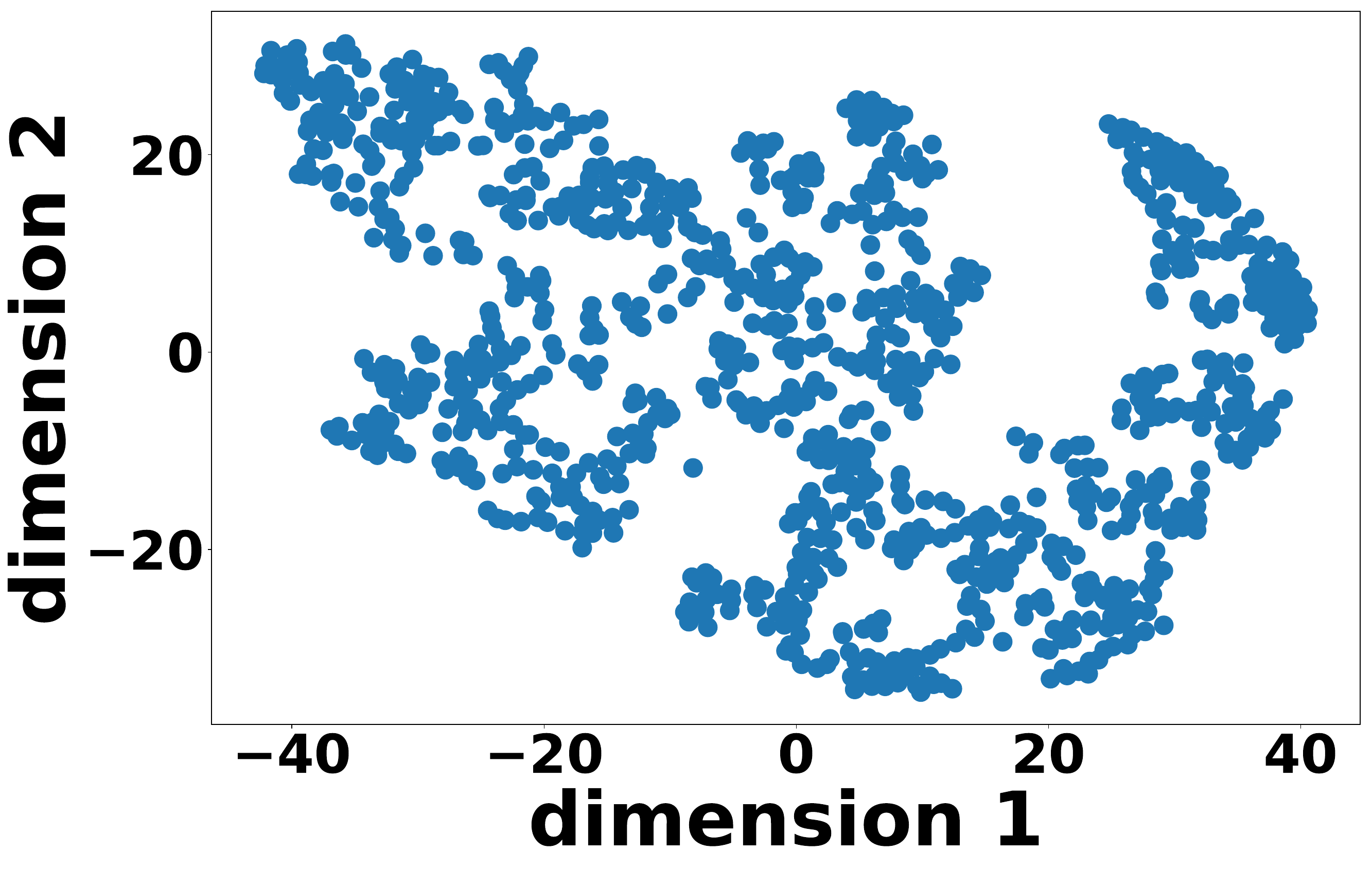}}
    \subfigure{\includegraphics[width=0.44\linewidth]{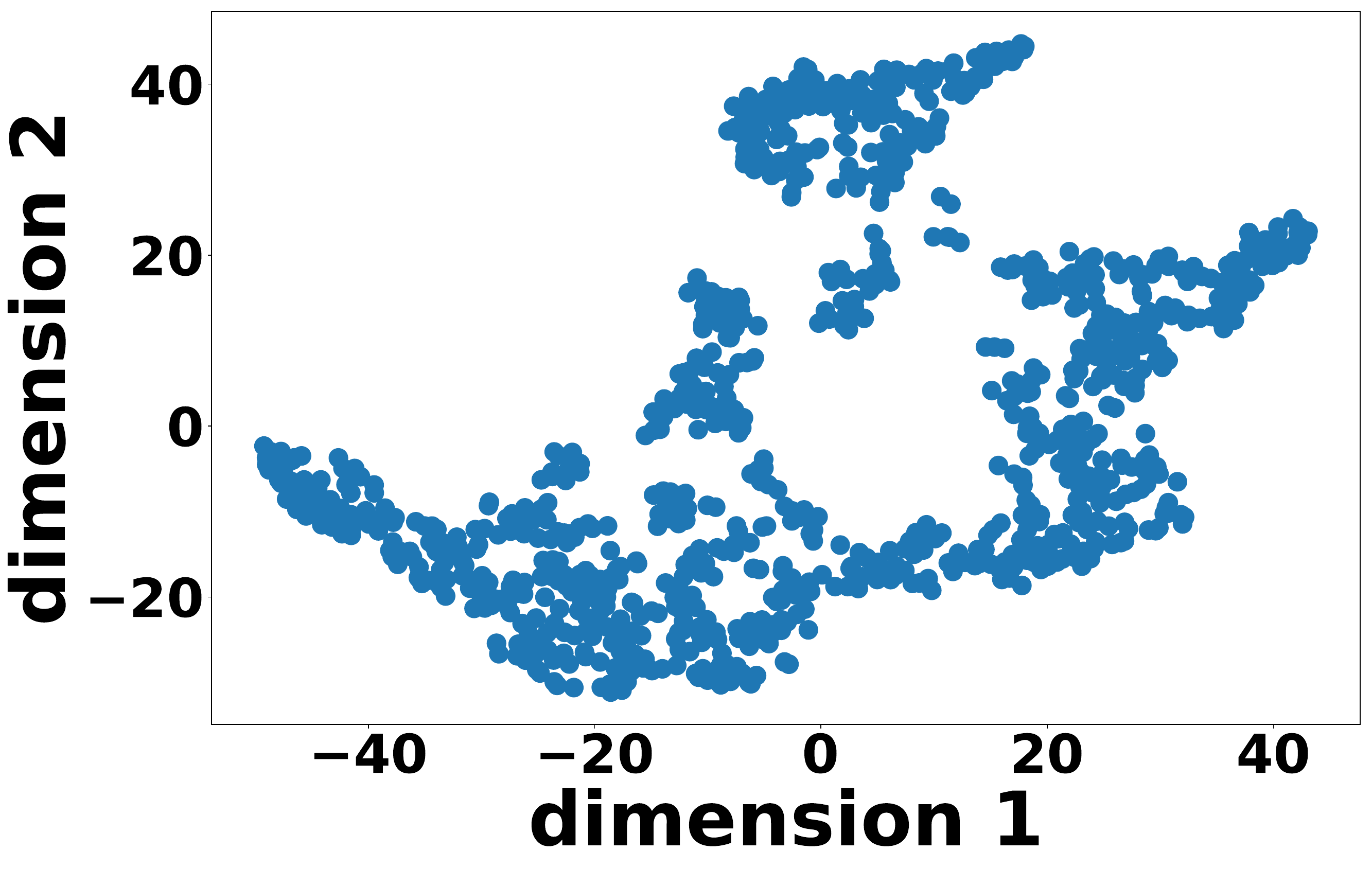}}
\caption{{Visualization of view 1 (left) and view 2 (right).}}
\label{fig:view}
\vspace{-.5cm}
\end{figure}

In the first experiment, we construct the views as follows.
{The shared components $\bm S=[\bm s_1,\ldots,\bm s_N] \in \mathbb{R}^{2\times N}$ are sampled from a parabola [i.e., $ s_{2,\ell}=(s_{1,\ell})^2$, where $s_{1,\ell}\in(-1,1)$]}. 
{In this case, the shared components are completely dependent.}
Besides, the sample mean is subtracted.
The shared components $\s_\ell\in\mathbb{R}^2$ for $\ell=1,\ldots,N$ are shown in Fig. \ref{fig:source}.
View-specific components $\bm{c}^{(q)}_\ell\in\mathbb{R}^{3}$ for $q=1,2$ are set to be i.i.d. Gaussian components with means {$-0.5$} and $0.8$, respectively, and variances $1.0^2$ and $1.5^2$, respectively. {Hence, we have $M_1=5$  and $M_2=5$. Note that under this setting, the conditions in Theorem~\ref{thm:dependent} are satisfied}. The sample size for each view is $N=1,000$. 
The elements of the mixing matrices $\bm{A}^{(1)}\in\mathbb{R}^{5\times 5},\bm{A}^{(2)}\in\mathbb{R}^{5\times 5}$ follow zero-mean unit-variance i.i.d. Gaussian distribution.
The nonlinear functions employed are as follows:
$g^{(1)}_1(x)=3\text{sigmoid}(x)+0.1x$,
$g^{(1)}_2(x)=5\text{sigmoid}(x)+0.2x$,
$g^{(1)}_3(x)=0.2\exp(x)$,
$g^{(1)}_4(x)=-4\text{sigmoid}(x)-0.3x$,
$g^{(1)}_5(x)=-3\text{sigmoid}(x)-0.2x$;
$g^{(2)}_1(x)=5\text{tanh}(x)+0.2x$,
$g^{(2)}_2(x)=2\text{tanh}(x)+0.1x$,
$g^{(2)}_3(x)=0.1x^3+x$,
$g^{(2)}_4(x)=-5\text{tanh}(x)-0.4x$,
$g^{(2)}_5(x)=-6\text{tanh}(x)-0.3x$.
Fig.~\ref{fig:view} visualizes the observed views (via t-SNE\cite{maaten2008visualizing}). One can see the views are severely distorted. In the experiment, our method uses $5$ individual neural networks in order to learn the inverse of the 5 nonlinear {mappings} from $g^{(q)}_1$ to $g^{(q)}_5$. Every network has one-hidden-layer with 256 neurons. We adopt the {\sf ReLU} activation function in the networks.

\begin{figure}
\centering
    \subfigure{\includegraphics[width=0.48\linewidth]{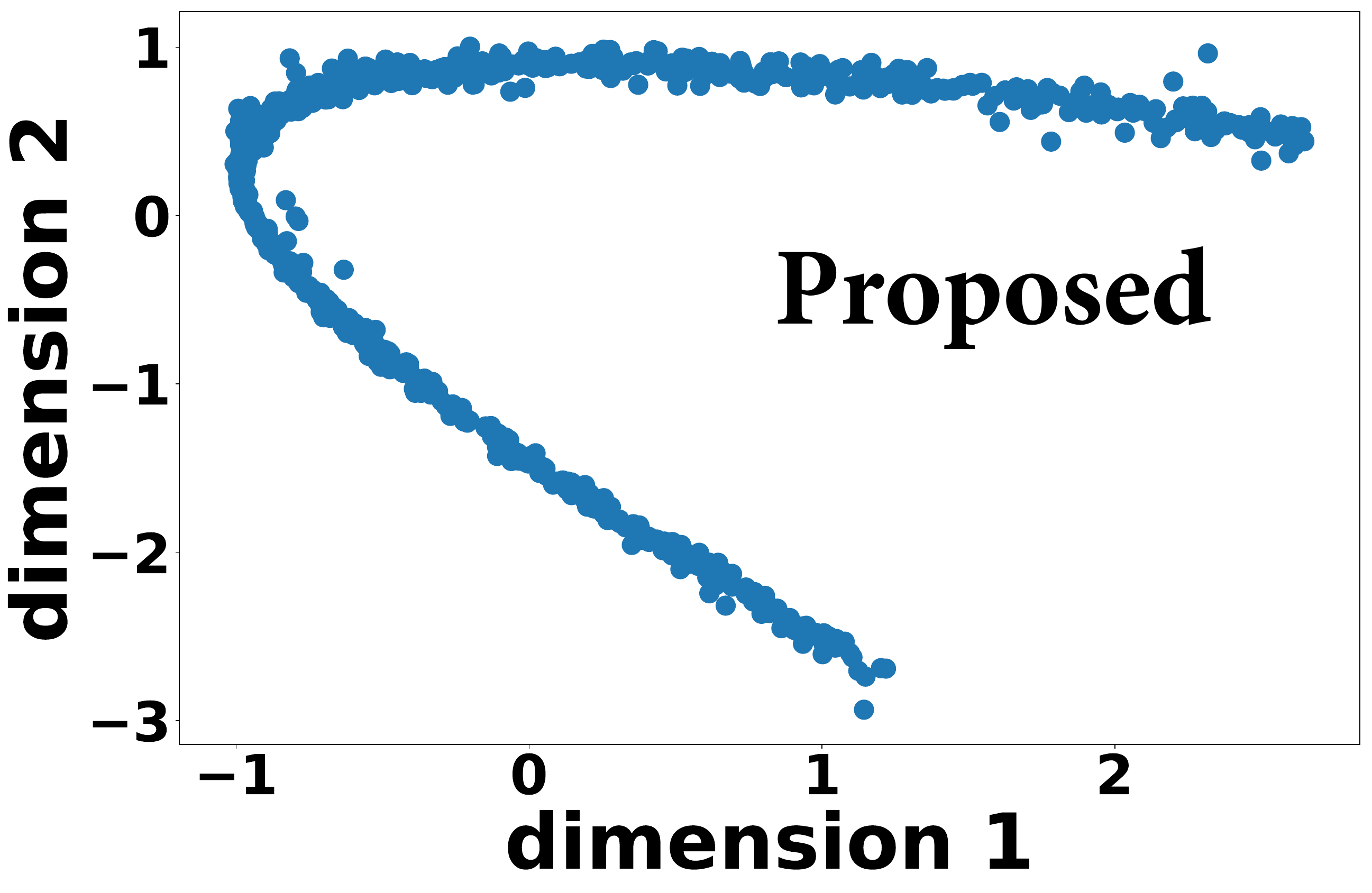}}
    \subfigure{\includegraphics[width=0.48\linewidth]{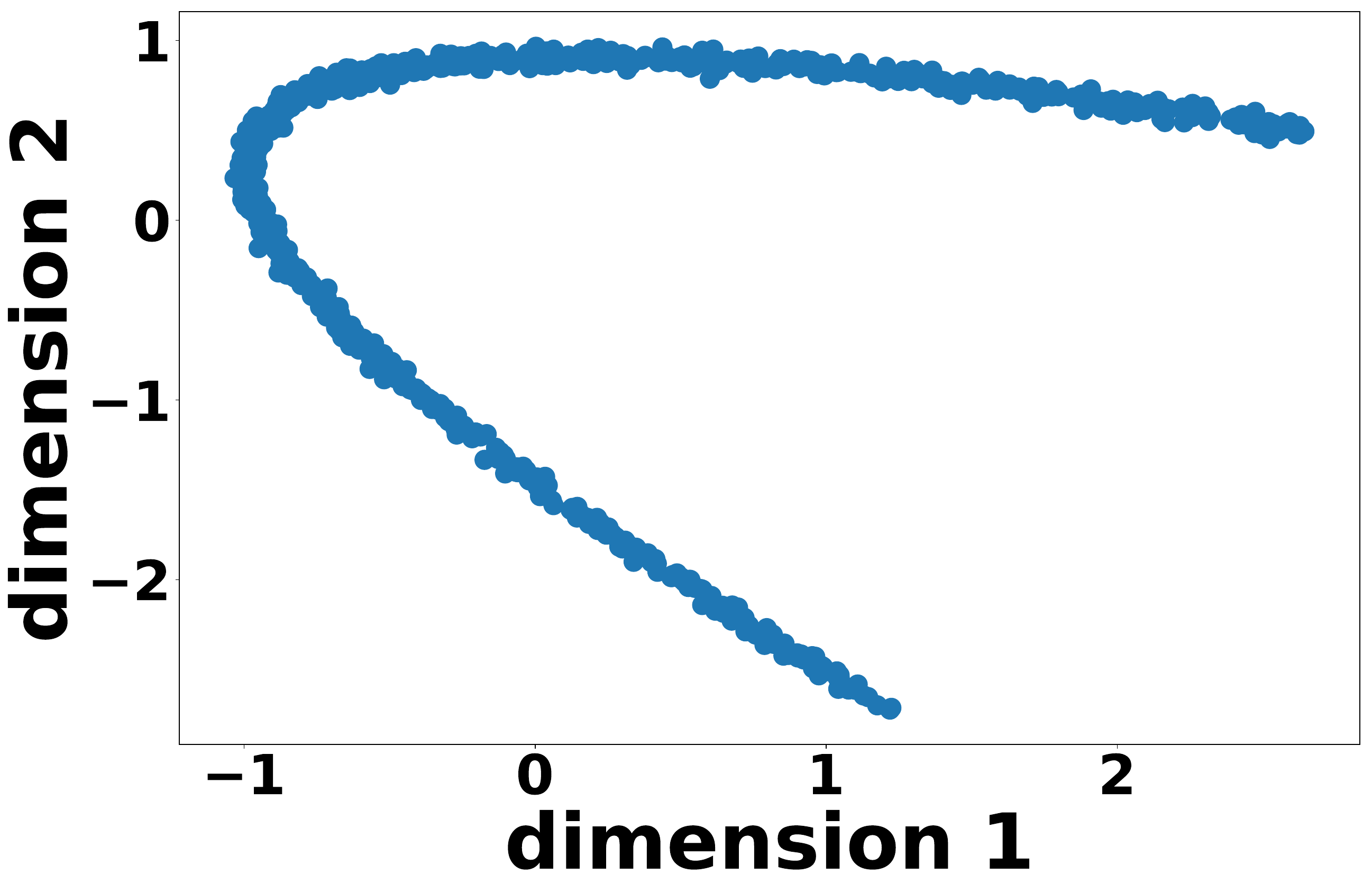}}

    \subfigure{\includegraphics[width=0.48\linewidth]{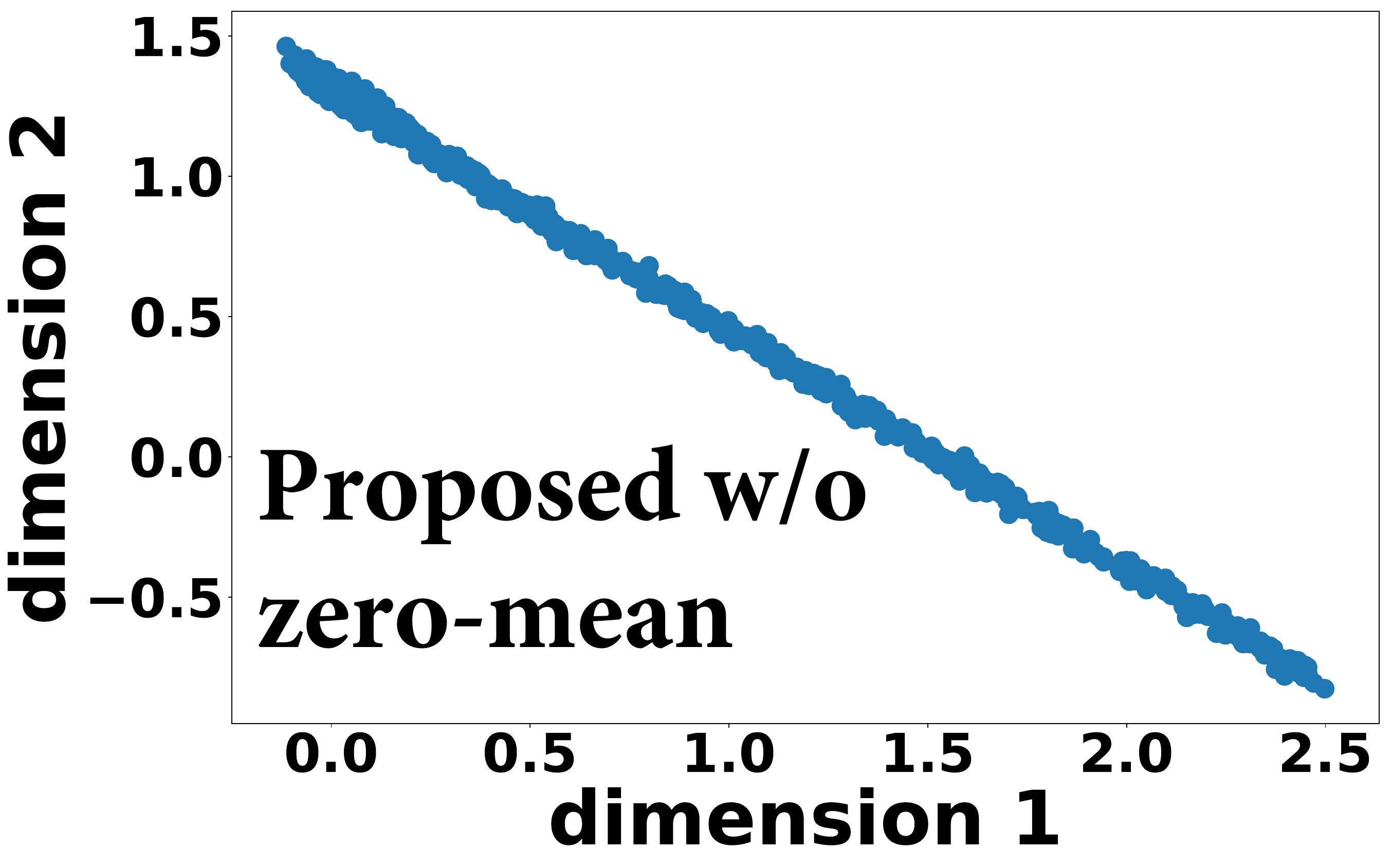}}
    \subfigure{\includegraphics[width=0.48\linewidth]{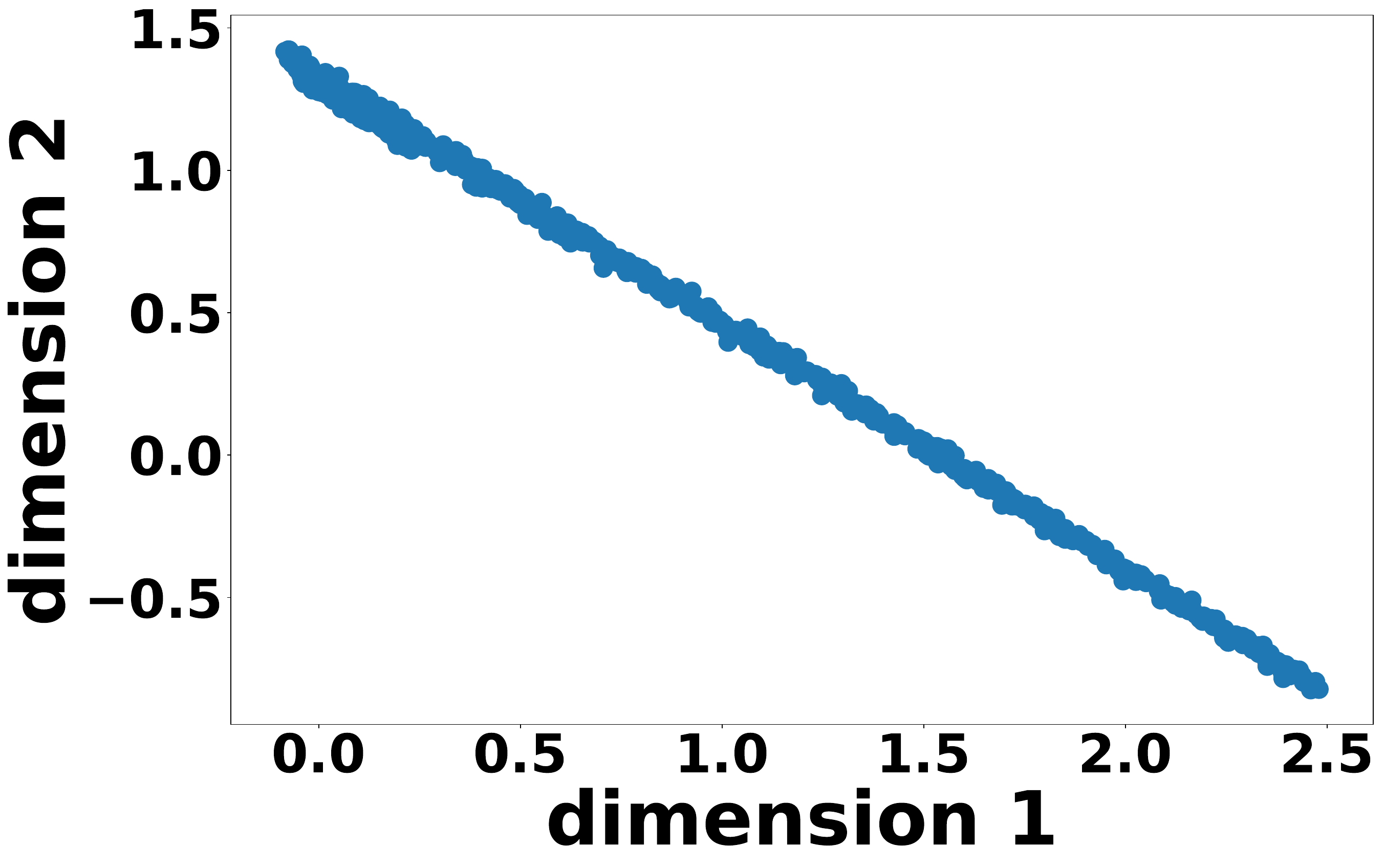}}
    
    \subfigure{\includegraphics[width=0.48\linewidth]{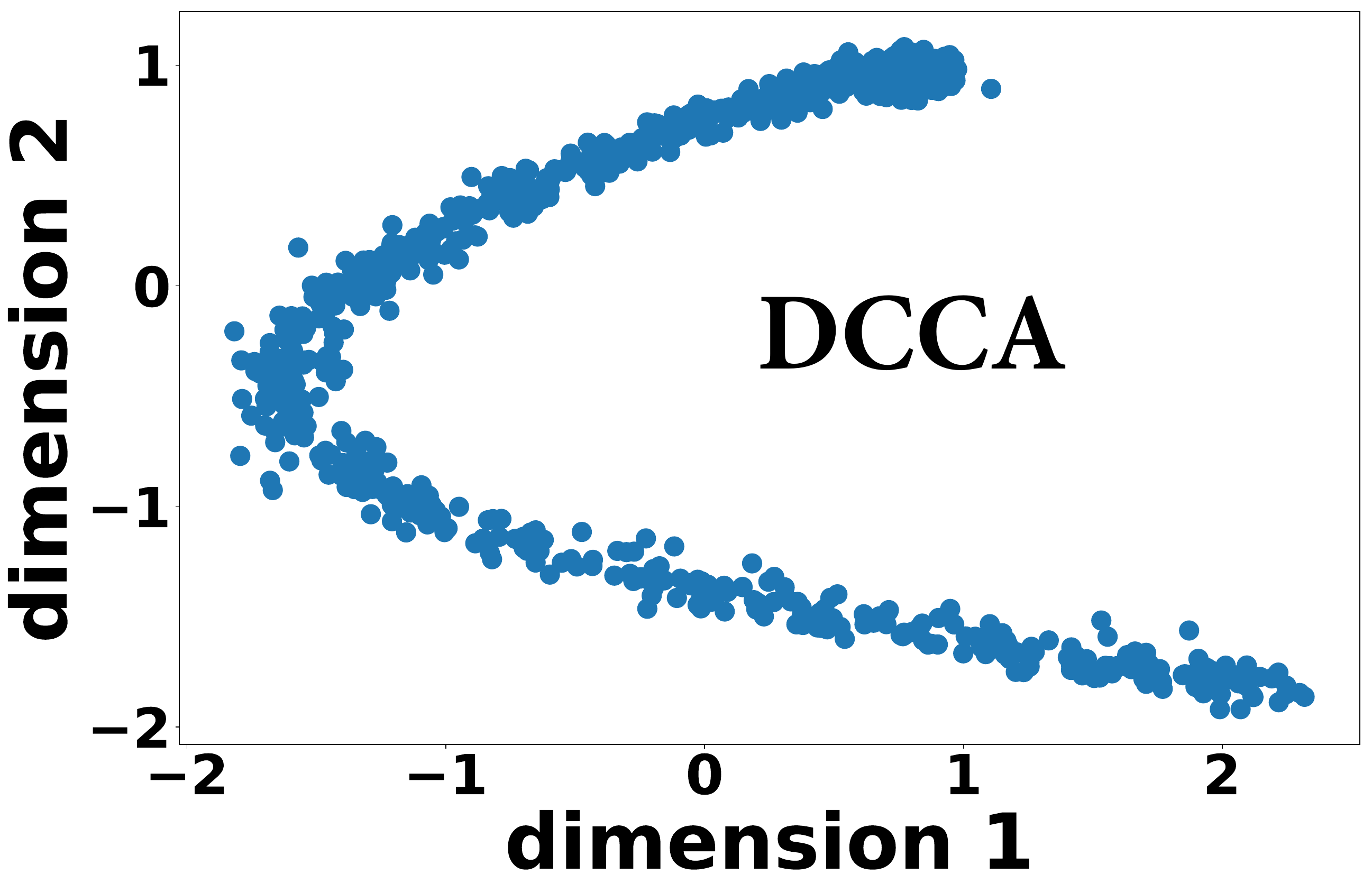}}
    \subfigure{\includegraphics[width=0.48\linewidth]{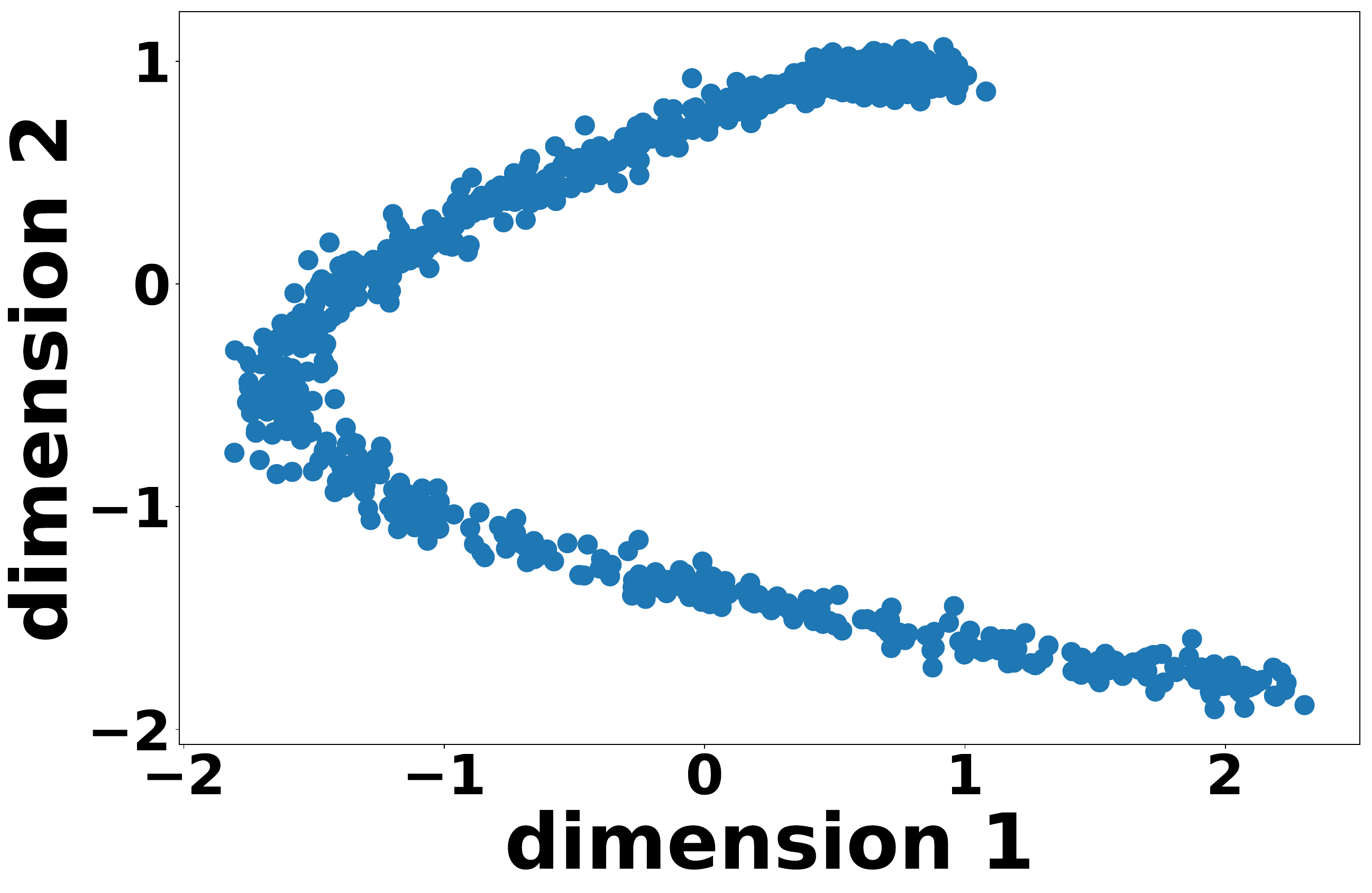}}
    
    \subfigure{\includegraphics[width=0.48\linewidth]{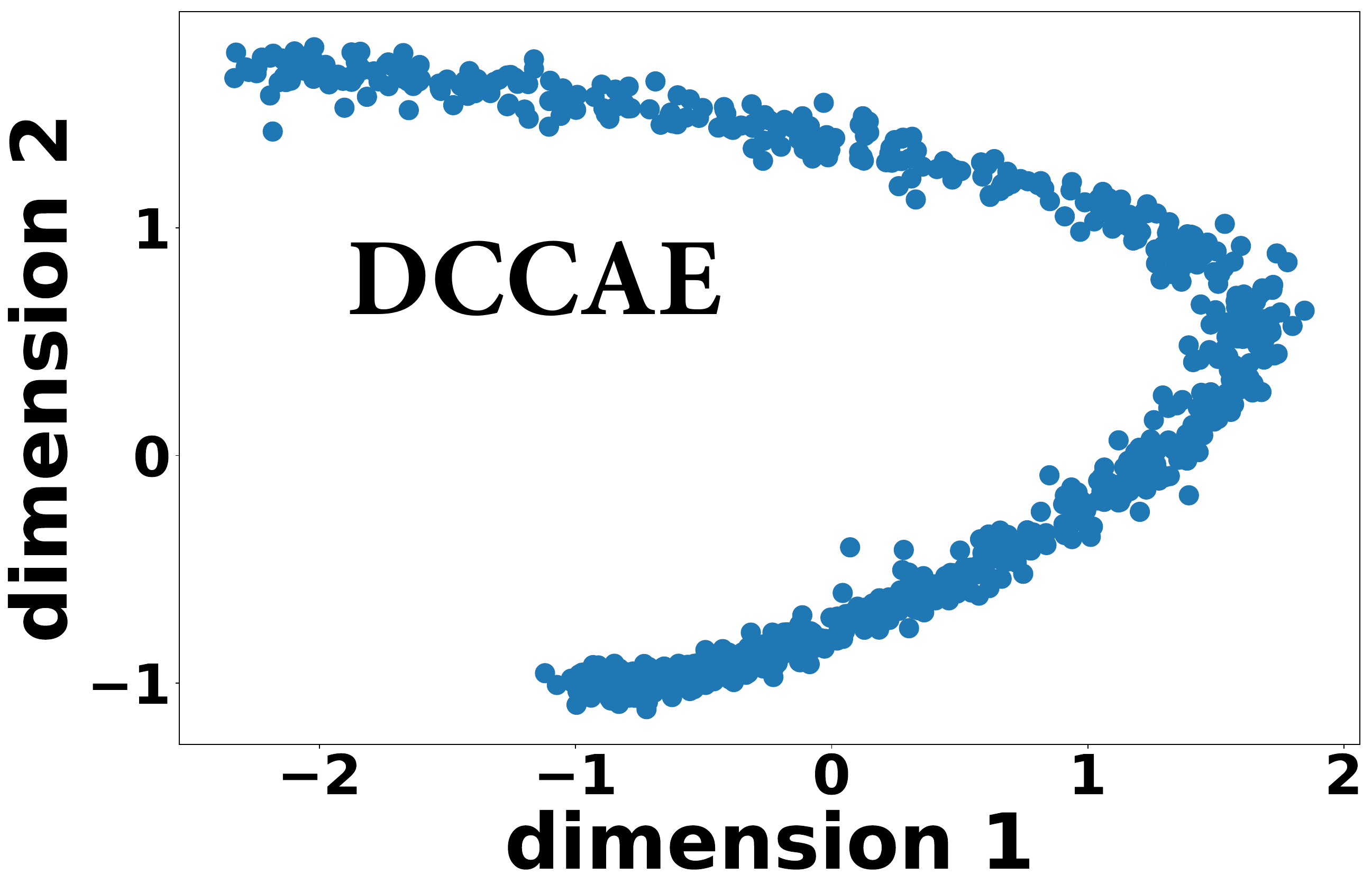}}
    \subfigure{\includegraphics[width=0.48\linewidth]{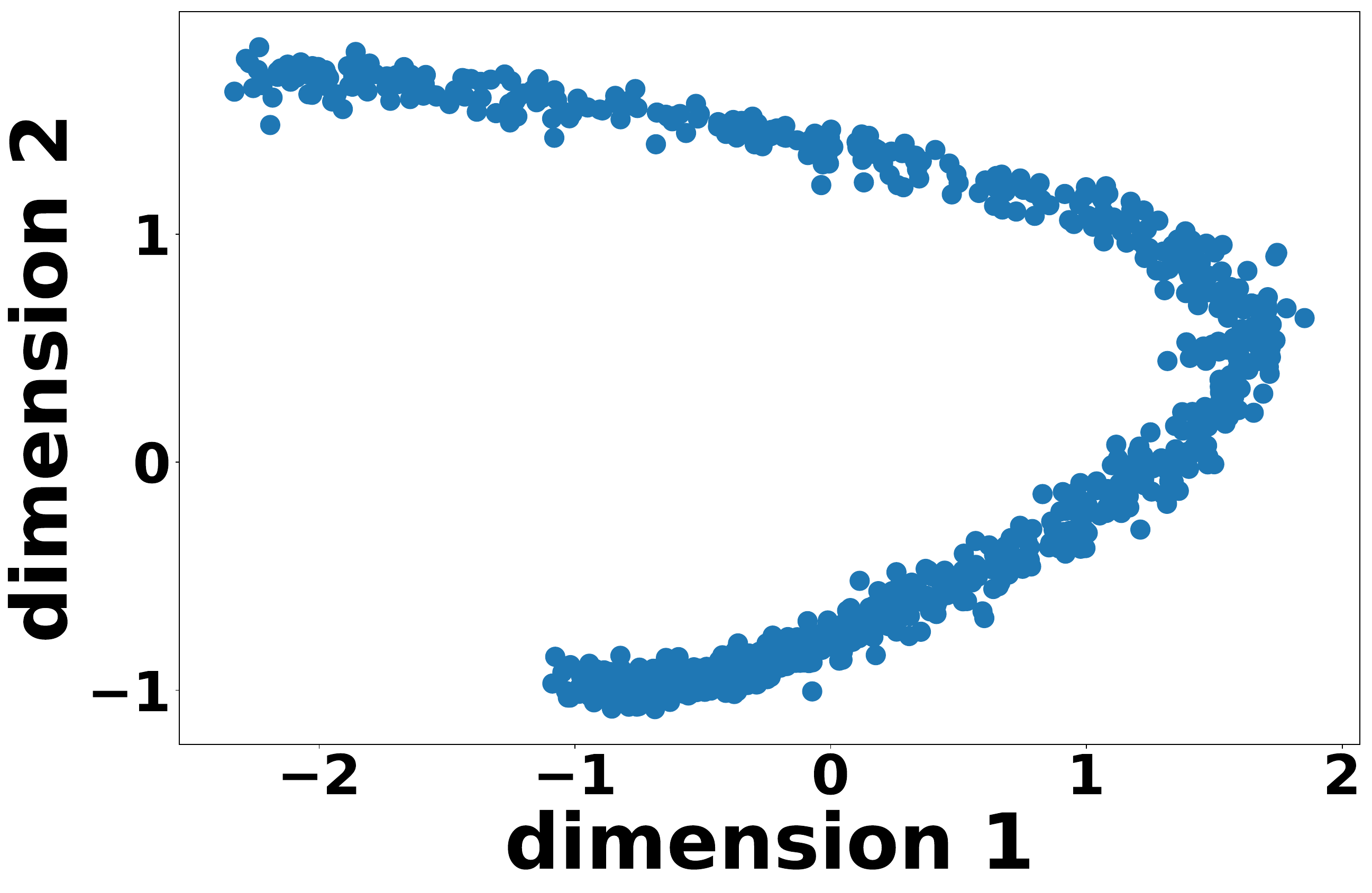}}
    
    \subfigure{\includegraphics[width=0.48\linewidth]{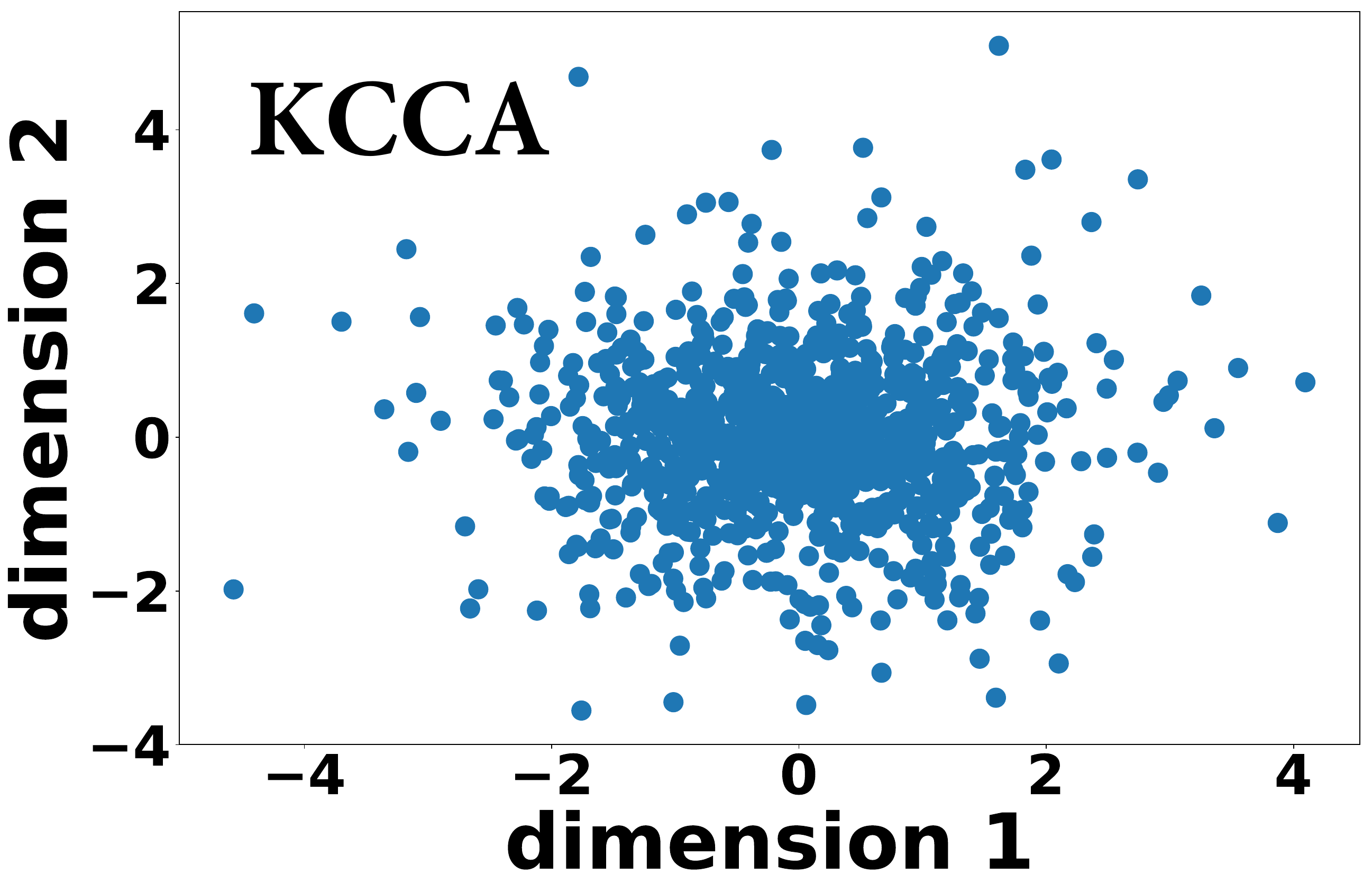}}
    \subfigure{\includegraphics[width=0.48\linewidth]{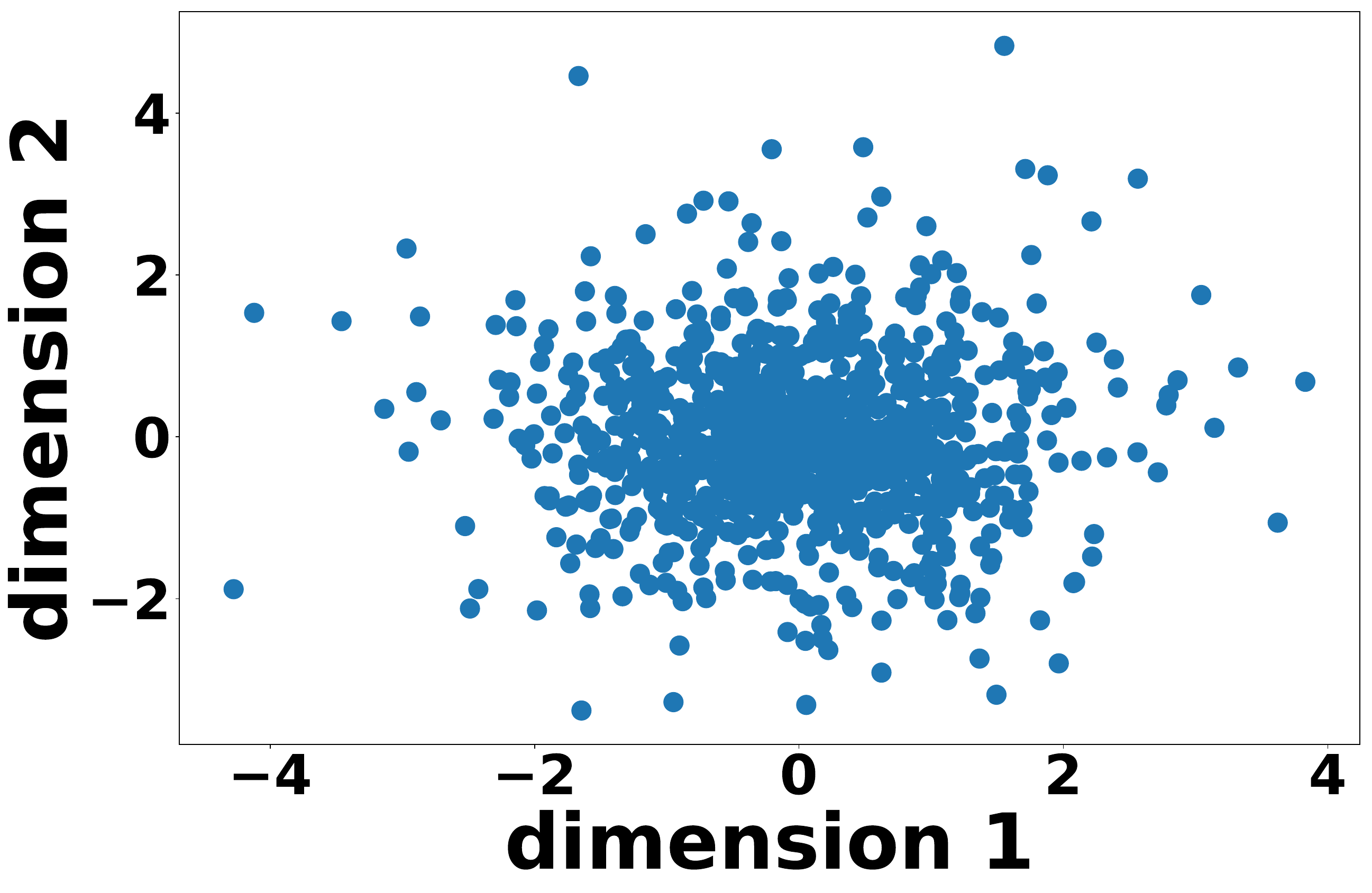}}
    
    \subfigure{\includegraphics[width=0.48\linewidth]{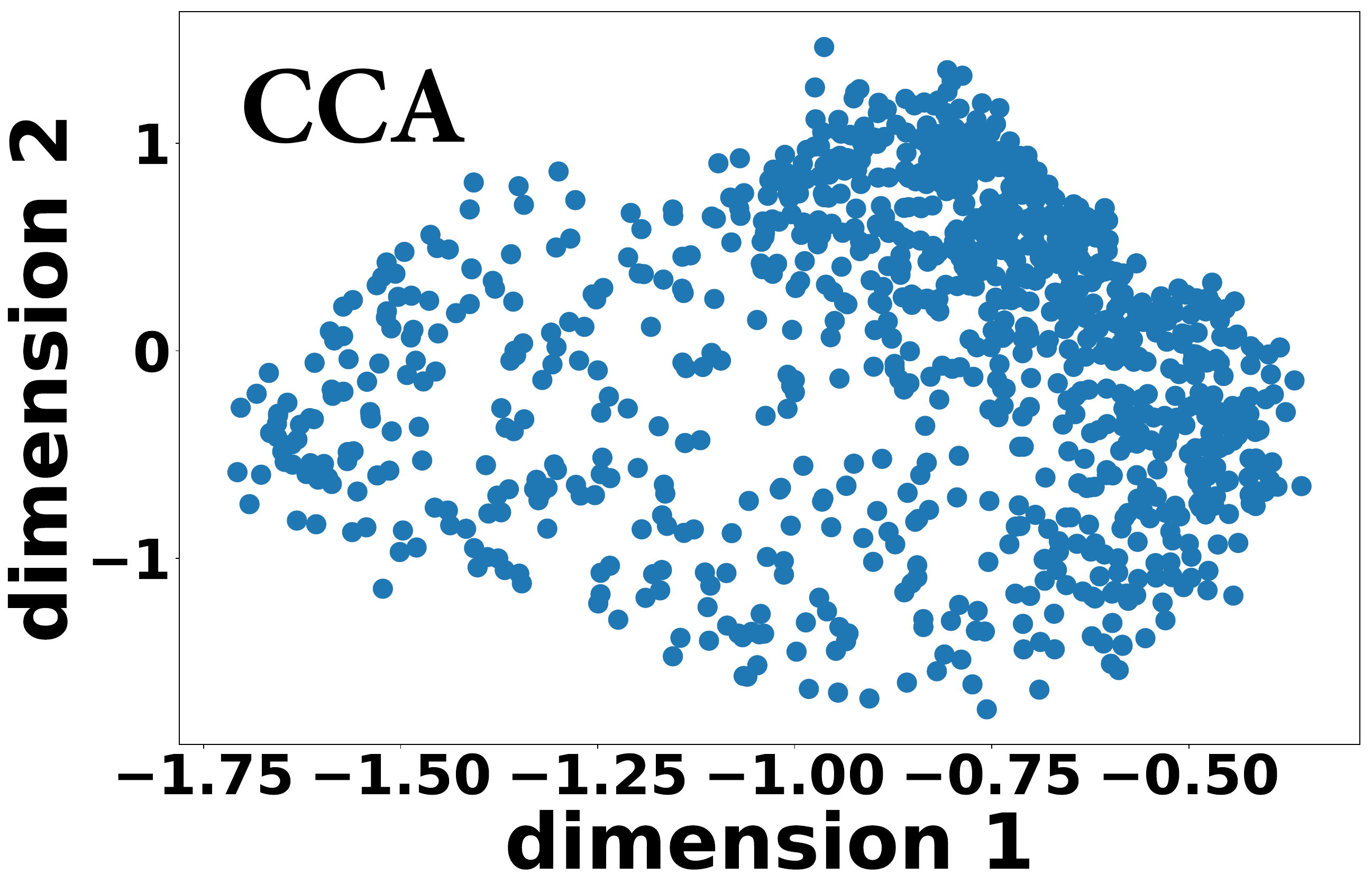}}
    \subfigure{\includegraphics[width=0.48\linewidth]{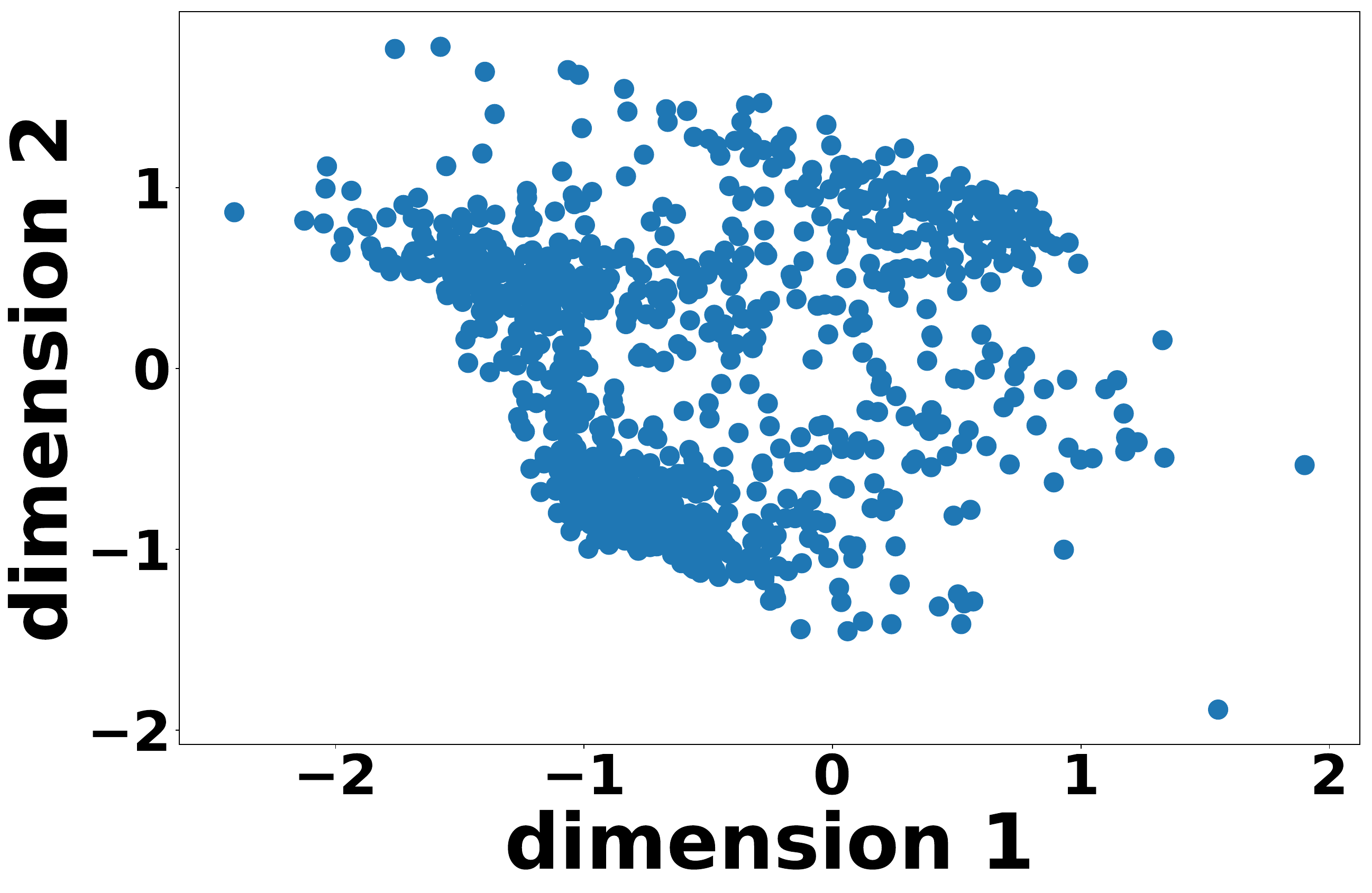}}
    
    \subfigure{\includegraphics[width=0.48\linewidth]{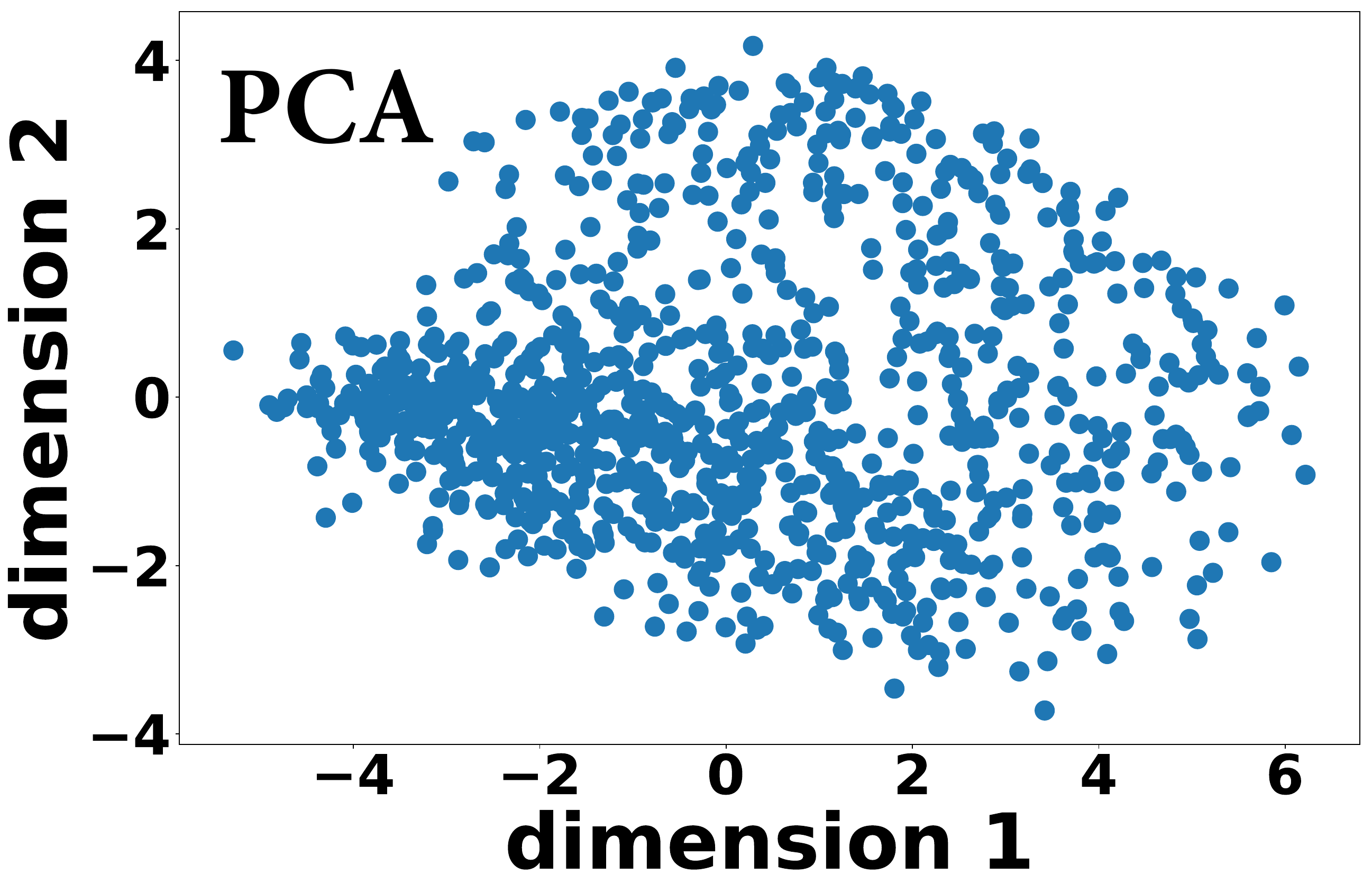}}
    \subfigure{\includegraphics[width=0.48\linewidth]{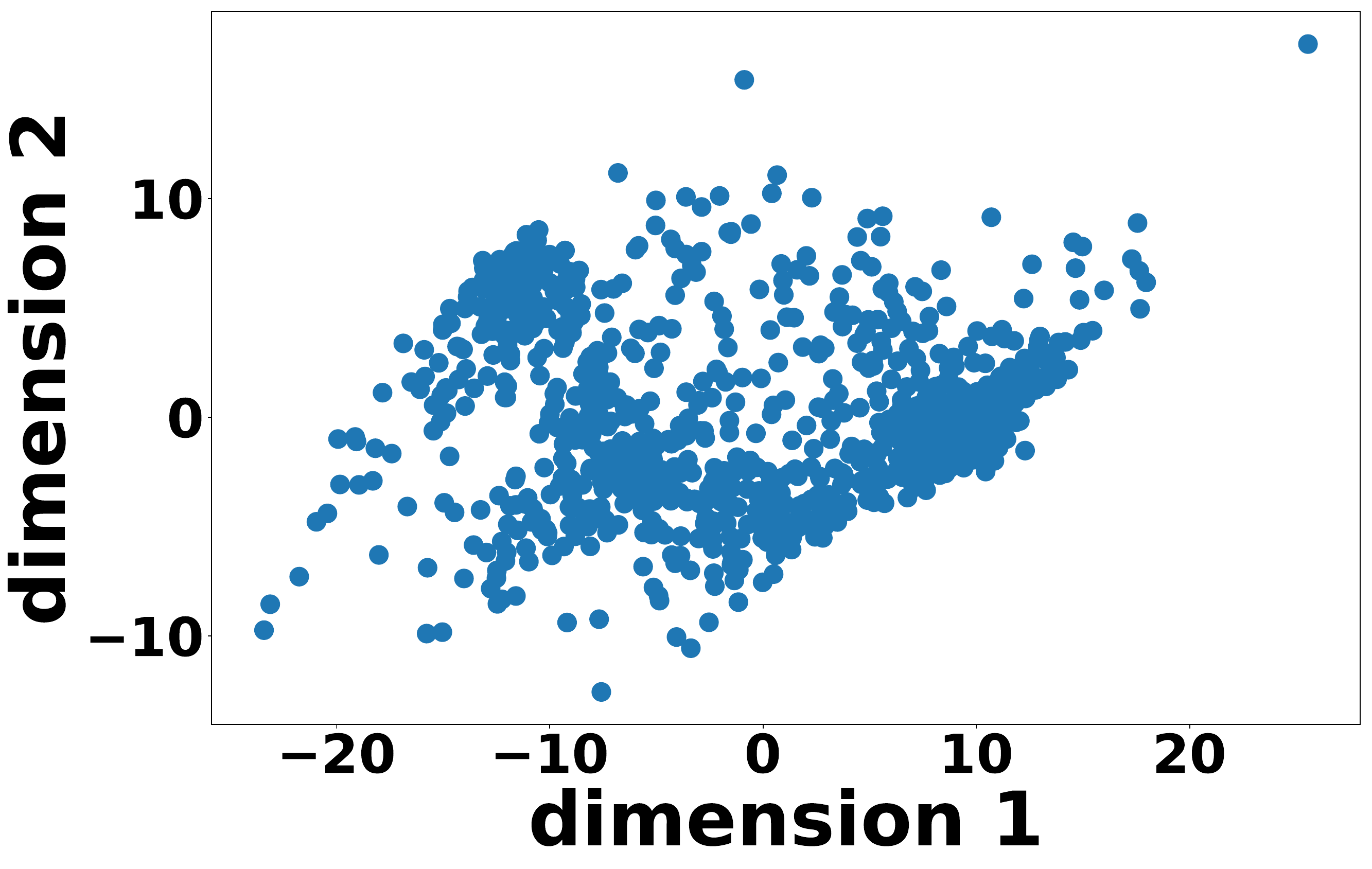}}
\caption{{Example of the recovered shared component. Top to bottom: Proposed, Proposed w/o zero-mean constraint, DCCA, DCCAE, KCCA, CCA and PCA.}}
\label{fig:recovered_source}
\vspace{-.5cm}
\end{figure}

The learned $\widehat{\bm x}_\ell=\B^{(q)}{\bm f_{\text{NN}}^{(q)}}(\y_\ell^{(q)})$ by the proposed approach and the outputs of the baselines are visualized in Fig.~\ref{fig:recovered_source}. One can see that the proposed approach yields a scaled and permuted version of the parabola in Fig.~\ref{fig:source}---which is as expected. The baselines are not as promising: DCCA and DCCAE also work reasonably---but the scatter plots are much noisier. The linear mapping based methods PCA and CCA clearly fail.
One particular observation is that if we do not impose the zero-mean constraint in \eqref{eq:opt}, the result is much worse compared to the one with this constraint. This supports our claim that trivial or noninteresting solutions may easily happen if the zero-mean constraint is not imposed.

\begin{figure}
\centering
    \subfigure{\includegraphics[width=0.49\linewidth]{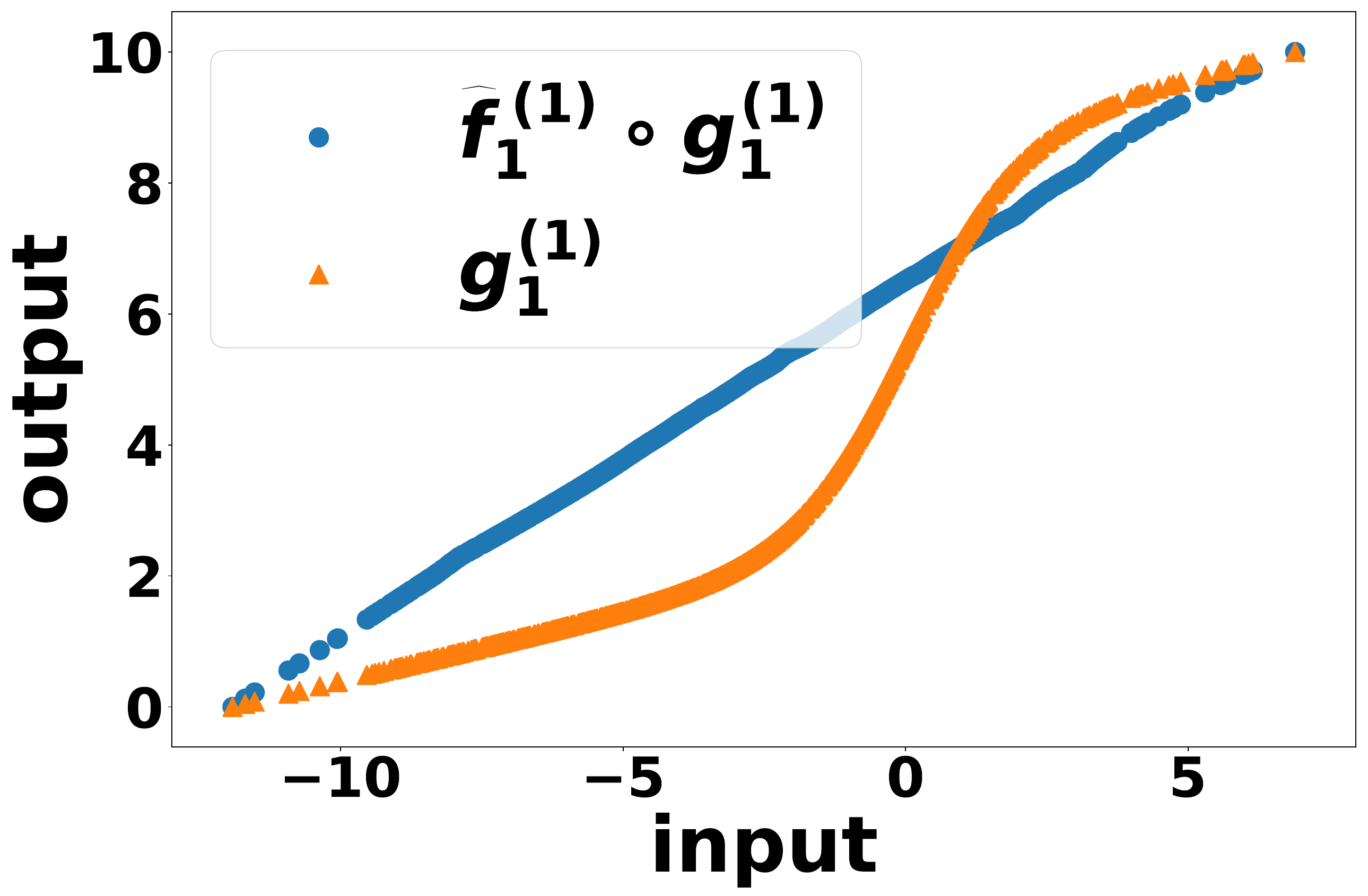}}
    \subfigure{\includegraphics[width=0.49\linewidth]{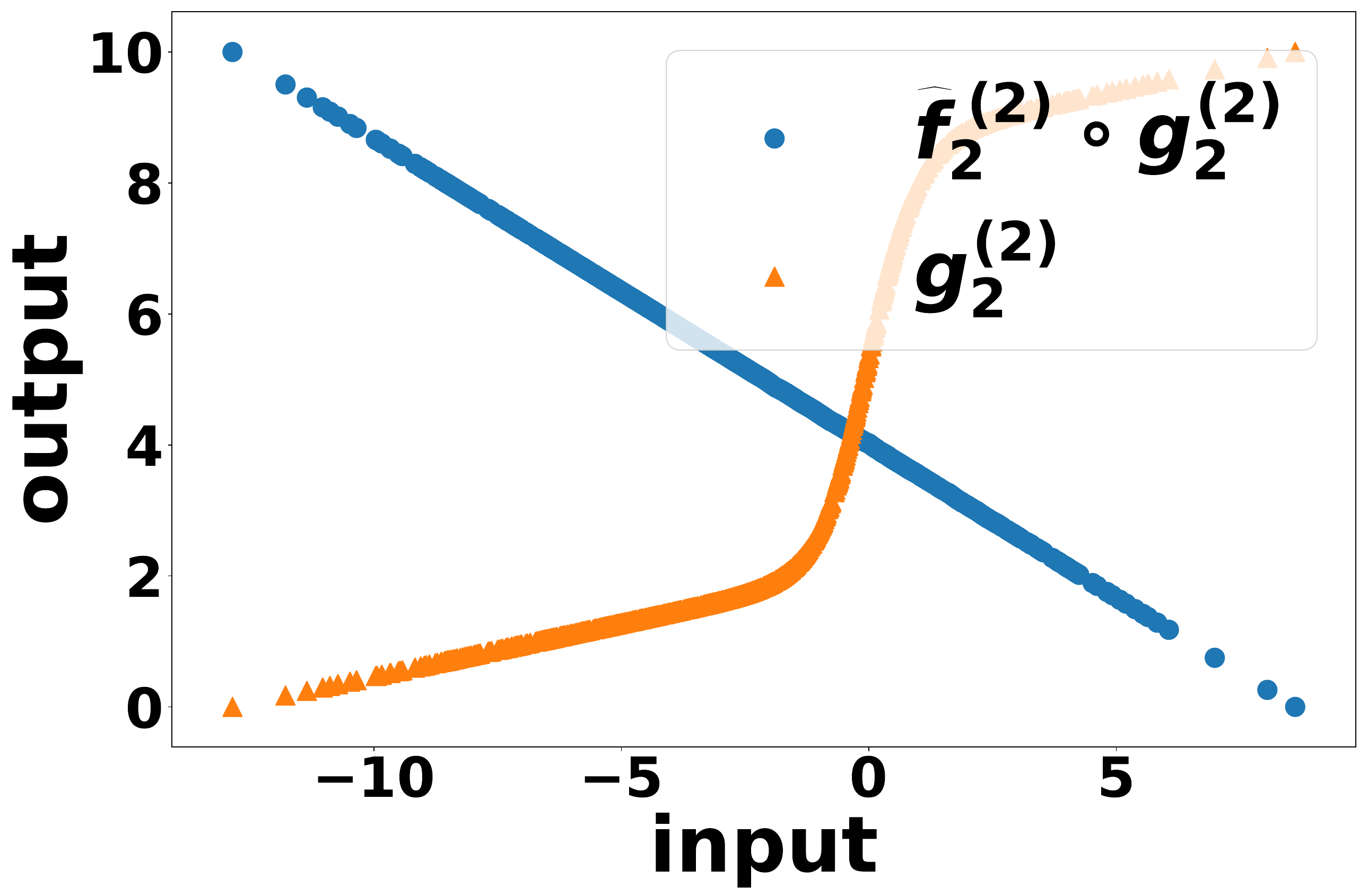}}
    
    \subfigure{\includegraphics[width=0.49\linewidth]{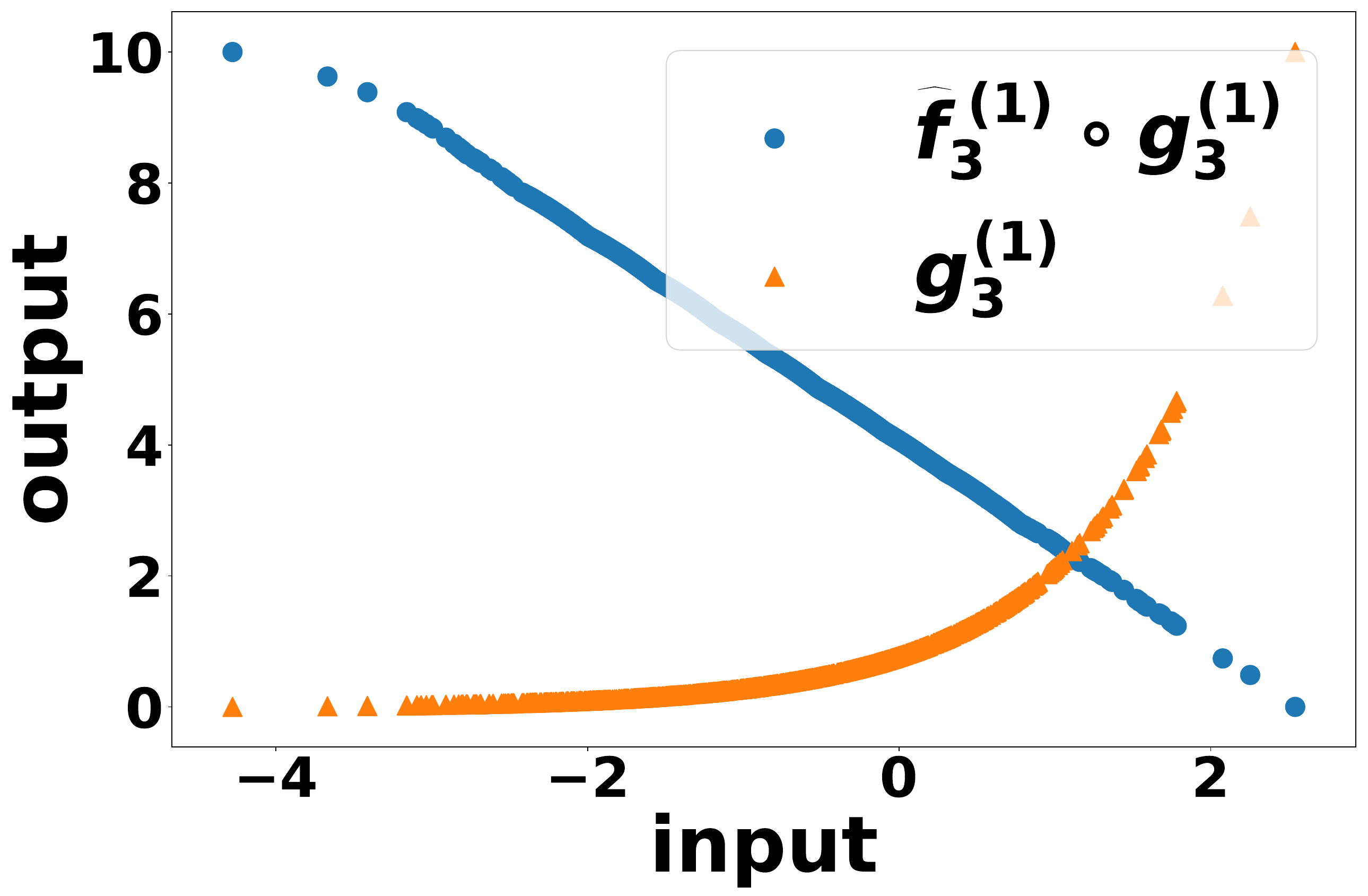}}
    \subfigure{\includegraphics[width=0.49\linewidth]{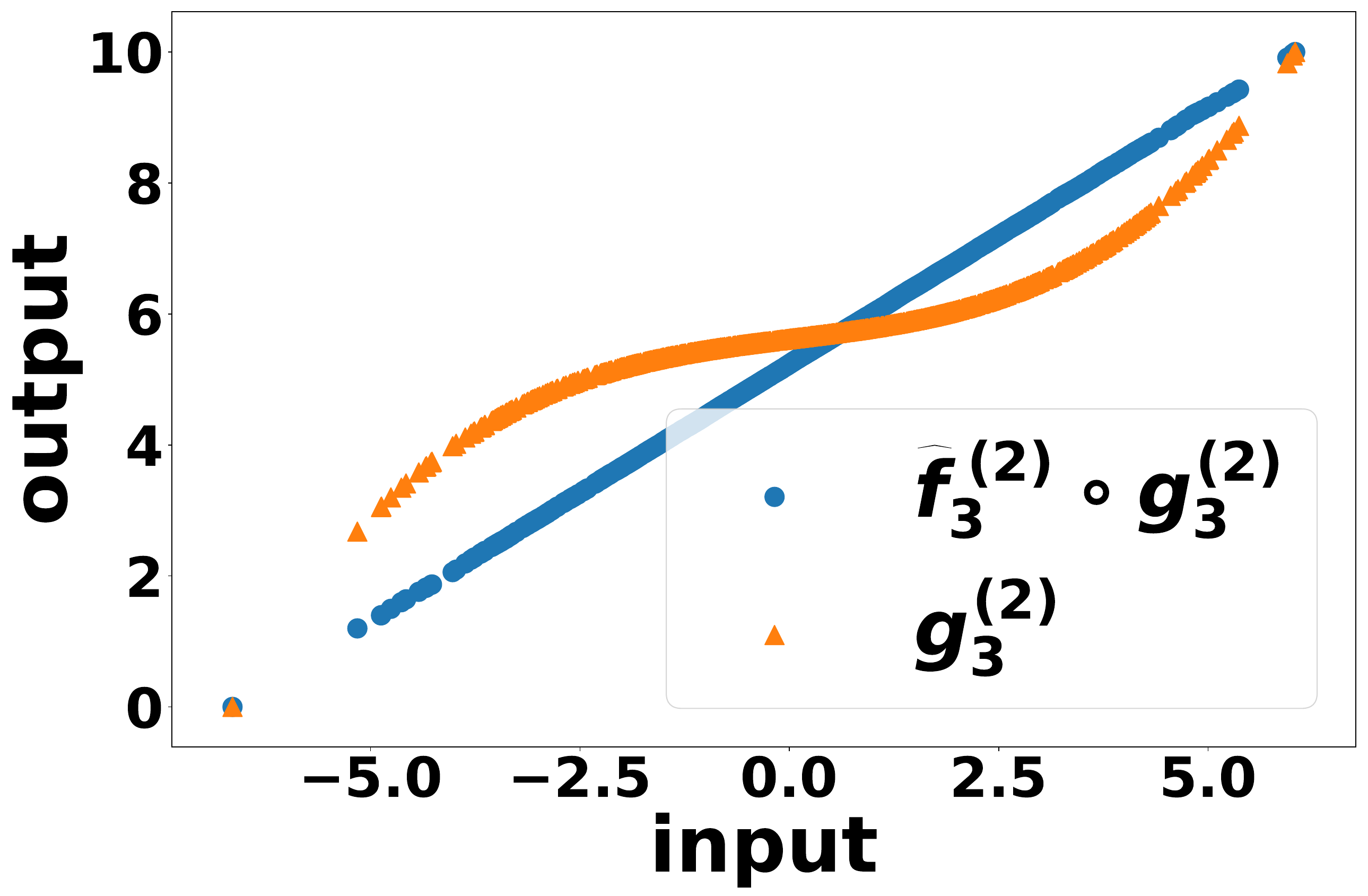}}
    
    \subfigure{\includegraphics[width=0.49\linewidth]{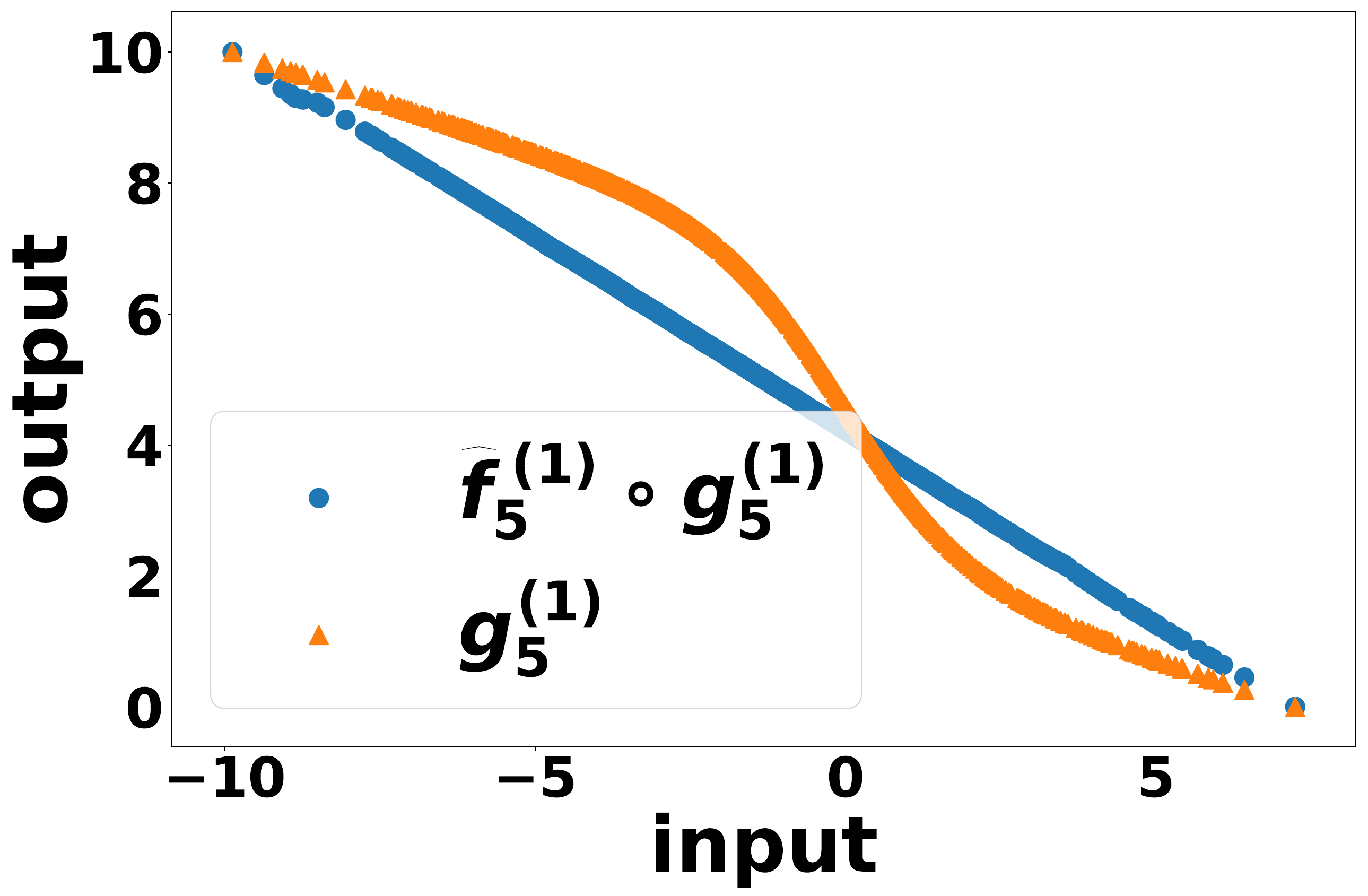}}
    \subfigure{\includegraphics[width=0.49\linewidth]{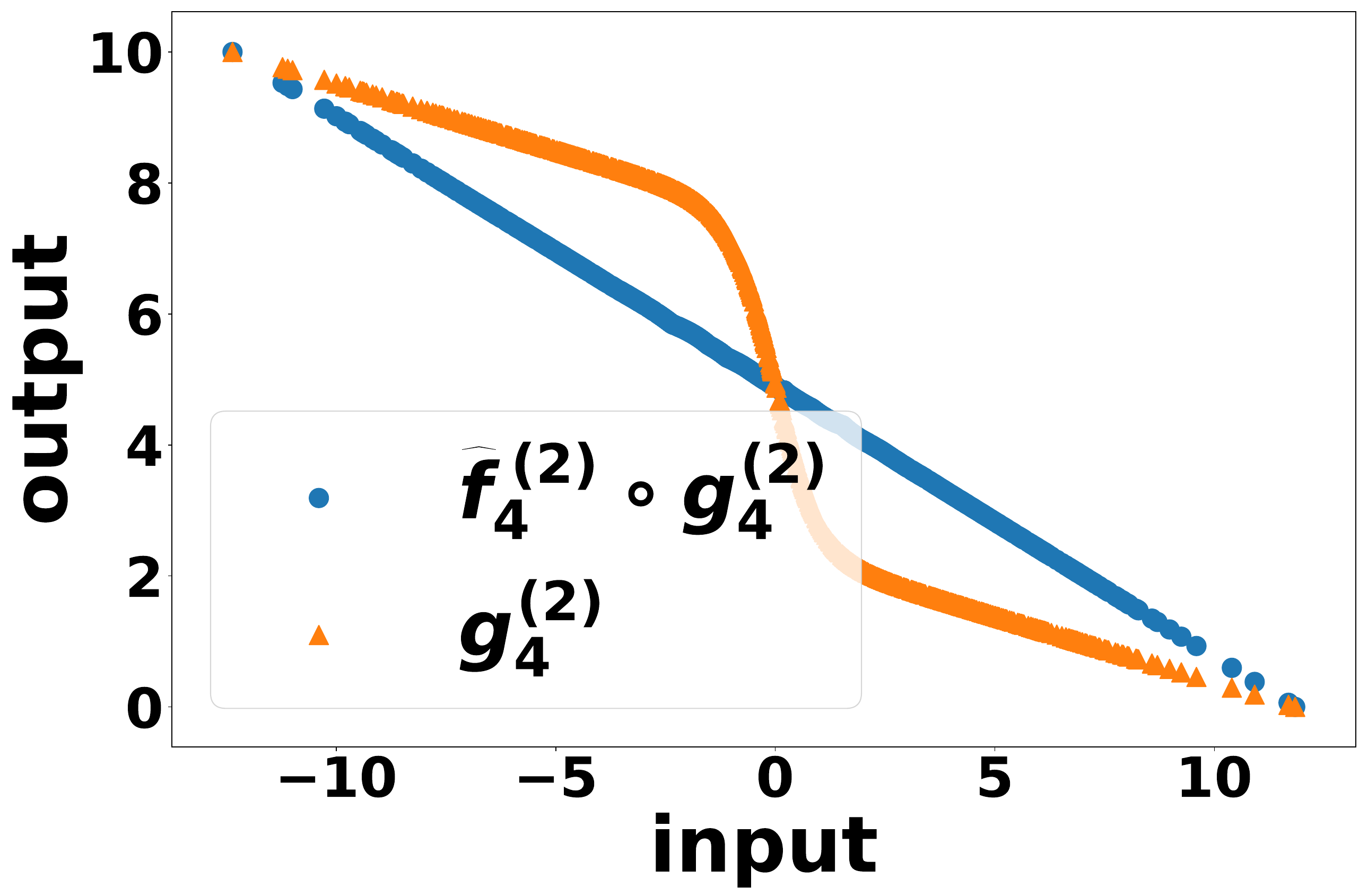}}
\caption{{The nonlinear distortion and the learned composite functions of two views. Left: view 1; Right: view 2. From top to bottom: randomly selected 3 dimensions.}}
\label{fig:composite}
\vspace{-.5cm}
\end{figure}

Fig.~\ref{fig:composite} shows the learned $\widehat{f}^{(q)}_m\circ g^{(q)}_m$ for $q=1,2$ with 3 randomly selected dimensions. One can see that in this simulation, all the composition functions are visually affine.

\begin{table}[t!]
	\centering
	\caption{{The {\sf dist} value [mean (standard deviation)] of the approaches under test.}}\label{tab:sim1}
	\resizebox{.9\linewidth}{!}{
		\begin{tabular}{c|ccccc}
			method  & Proposed & Proposed w/o zero-mean & DCCA  \\ \hline \hline
			{\sf dist} &  $0.035(8.95e^{-3})$   &   $0.998(7.35e^{-4})$   &  $0.523(1.51e^{-1})$    \\
			method  & DCCAE & KCCA & CCA \\ \hline \hline
			{\sf dist} & $0.530(1.63e^{-1})$ &  $0.998(4.57e^{-3})$ &   $0.993(8.56e^{-3})$   \\
			method & PCA  \\\hline \hline
			{\sf dist} &   $0.999(1.52e^{-3})$
		\end{tabular}}
		\vspace{-.5cm}
\end{table}

Table~\ref{tab:sim1} shows the averaged {\sf dist} from 10 random trials under the same settings as those for in Figs.~\ref{fig:recovered_source}-\ref{fig:composite}. One can see that the proposed approach admits a {\sf dist} value that is almost zero, validating our subspace identifiability claim. It is much lower than those of the baselines, perhaps because our method exploits model information. Again, subtracting the mean in our formulation [cf. the constraint $(1/N)\sum_{\ell=1}^N\u_\ell=\bm 0$] is indeed crucial, since the result without this constraint is much worse.

\begin{table}[t!]
\centering
\caption{{The {\sf dist} value [mean (standard deviation)] of the approaches against number of samples.}}\label{tab:samples}
\resizebox{\linewidth}{!}{\Huge
\begin{tabular}{c|ccccc}
\# samples  & 500 & 1000 & 2000 & 5000 & 10000 \\ \hline \hline
Proposed &  $0.101(1.65e^{-2})$   &  $0.035(8.95e^{-3})$    &   $0.030(3.61e^{-3})$   &   $0.017(2.00e^{-3})$   &  $0.009(1.60e^{-3})$    \\
DCCA & $0.799(3.31e^{-2})$ & $0.523(1.51e^{-1})$ & $0.316(3.85e^{-2})$ & $0.165(9.82e^{-3})$ & $0.129(1.02e^{-2})$ \\
DCCAE & $0.817(3.31e^{-2})$ & $0.530(1.63e^{-1})$ & $0.310(3.39e^{-2})$ &  $0.183(1.21e^{-2})$ & $0.118(1.11e^{-2})$ \\
KCCA & $0.998(1.70e^{-3})$ & $0.998(4.57e^{-3})$ & $0.999(5.01e^{-3})$ & $0.989(5.10e^{-3})$ & $0.986(1.50e^{-2})$ \\
CCA & $0.987(5.32e^{-3})$& $0.993(8.56e^{-3})$ & $0.988(3.67e^{-3})$ & $0.998(7.23e^{-3})$ & $0.978(2.36e^{-2})$ \\
PCA & $0.987(4.12e^{-2})$ & $0.999(1.52e^{-3})$ & $0.993(3.37e^{-2})$& $0.998(7.33e^{-3})$ & $0.992(1.27e^{-2})$ 
\end{tabular}}
\vspace{-.5cm}
\end{table}

Table~\ref{tab:samples} shows the performance of the algorithms when the number of samples $N$ changes. 
 {Again, we have $\s_\ell\in\mathbb{R}^2$ and each view has three interference components in this simulation.} 
Note that the nonlinearity removal theorems are derived under the population case, i.e., the functional equations in \eqref{eq:equal} hold for all $\bm s_\ell\in{\cal S}$ and $\bm c^{(q)}_\ell\in{\cal C}_q$. This is not attainable in practice, where we only have a finite number of samples. Hence, it is of interest to observe how the performance scales with $N$.
From Table~\ref{tab:samples} one can see that the {\sf dist} value of the proposed approach clearly decreases when the number of samples increases from 500 to 10,000. On the other hand, using 500 samples, the proposed approach already exhibits good performance.

To observe the impact of the view-specific interference, we define the {\it Shared Component to Interference Ratio} (SCIR)
$${\sf SCIR}=10\log_{10}\left(\frac{\|\S\|_F^2/K}{\frac{1}{Q}\sum_{q=1}^Q\|\C^{(q)}\|_F^2/R_q}\right)~{\rm dB}.$$ 
Table~\ref{tab:scir} shows the result under different {\sf SCIR}s. {Other settings are the same as those in the previous simulation.}
One can see that even if the ratio is -20 dB, the performance of the proposed approach is still very good---the average subspace distance metric is 0.064,
whereas all the other baselines fail to work.
This observation is consistent with our analysis, as well as the interference-robust property of linear CCA \cite{ibrahim2019cell}.

\begin{table}[t!]
	\centering
\caption{{The {\sf dist} value [mean (standard deviation)] of the approaches  {\sf SCIR}.}}
\label{tab:scir}
\begin{tabular}{c|ccc}
SCIR  & {$-10$} dB & {$-20$} dB \\ \hline \hline
Proposed & $0.035(8.95e^{-3})$ & $0.064(3.40e^{-2})$   \\
DCCA & $0.523(1.51e^{-1})$ & $0.999(2.16e^{-3})$  \\
DCCAE & $0.530(1.63e^{-1})$ & $0.998(1.71e^{-3})$  \\
KCCA & $0.998(4.57e^{-3})$ & $0.999(2.32e^{-3})$   \\
CCA & $0.993(8.56e^{-3})$ & $0.998(1.09e^{-3})$ \\
PCA & $0.999(1.52e^{-3})$ & $0.997(2.13e^{-3})$
\end{tabular}
\vspace{-.5cm}
\end{table}

\begin{table}[t!]
\centering
\caption{{The {\sf dist} value [mean (standard deviation)] under Different structures of neural networks.}}
\label{tab:struct}
\begin{tabular}{cc|c}
\begin{tabular}[c]{@{}c@{}}\# params\\ of each dim\end{tabular} & structure      & {\sf dist} \\ \hline \hline 
\multirow{2}{*}{$\approx$ 512}                                            & {[}1,256,1{]}      &    $0.035(8.95e^{-3})$     \\ 
                                                                & {[}1,16,16,16,1{]} &   $0.058(9.32e^{-3})$       \\ \hline
\multirow{2}{*}{$\approx$ 2048}                                           & {[}1,1024,1{]}     &    $0.039(1.05e^{-2})$      \\ 
                                                                & {[}1,32,32,32,1{]} & $0.048(7.68e^{-3})$   \\ \hline
\multirow{2}{*}{$\approx$ 8192}                                           & {[}1,4096,1{]}     &   $0.053(2.13e^{-2})$       \\ 
                                                                & {[}1,64,64,64,1{]} &   $0.058(2.25e^{-2})$
\end{tabular}
\vspace{-.5cm}
\end{table}

\begin{table}[t]
	\centering
	\caption{Different dimensions of source components}\label{tab:dimension}
	\resizebox{.95\linewidth}{!}{\huge
		\begin{tabular}{c|cccc}
			source dim & 1           & 2       & 3     & 4     \\ \hline \hline
			{\sf dist}     & $0.032(1.00e^{-3})$ & $0.016(2.56e^{-3})$ & $0.015(2.62e^{-3})$ & $0.015(3.34e^{-3})$
	\end{tabular}}
	\vspace{-.5cm}
\end{table}

Table~\ref{tab:struct} shows the {\sf dist} value against different network structures. 
Observing this is of interest, since in machine learning tuning the structure of networks to fit a task can sometimes be tedious. The proposed framework is unsupervised, which means that there may not even be training samples for us to tune the network structure.
Fortunately, the observation here is that the result is not heavily affected by the choice of neural network structures---although using more parameters and more layers could slightly improve performance. 

\begin{figure*}
\centering
    \subfigure{\includegraphics[width=0.19\linewidth]{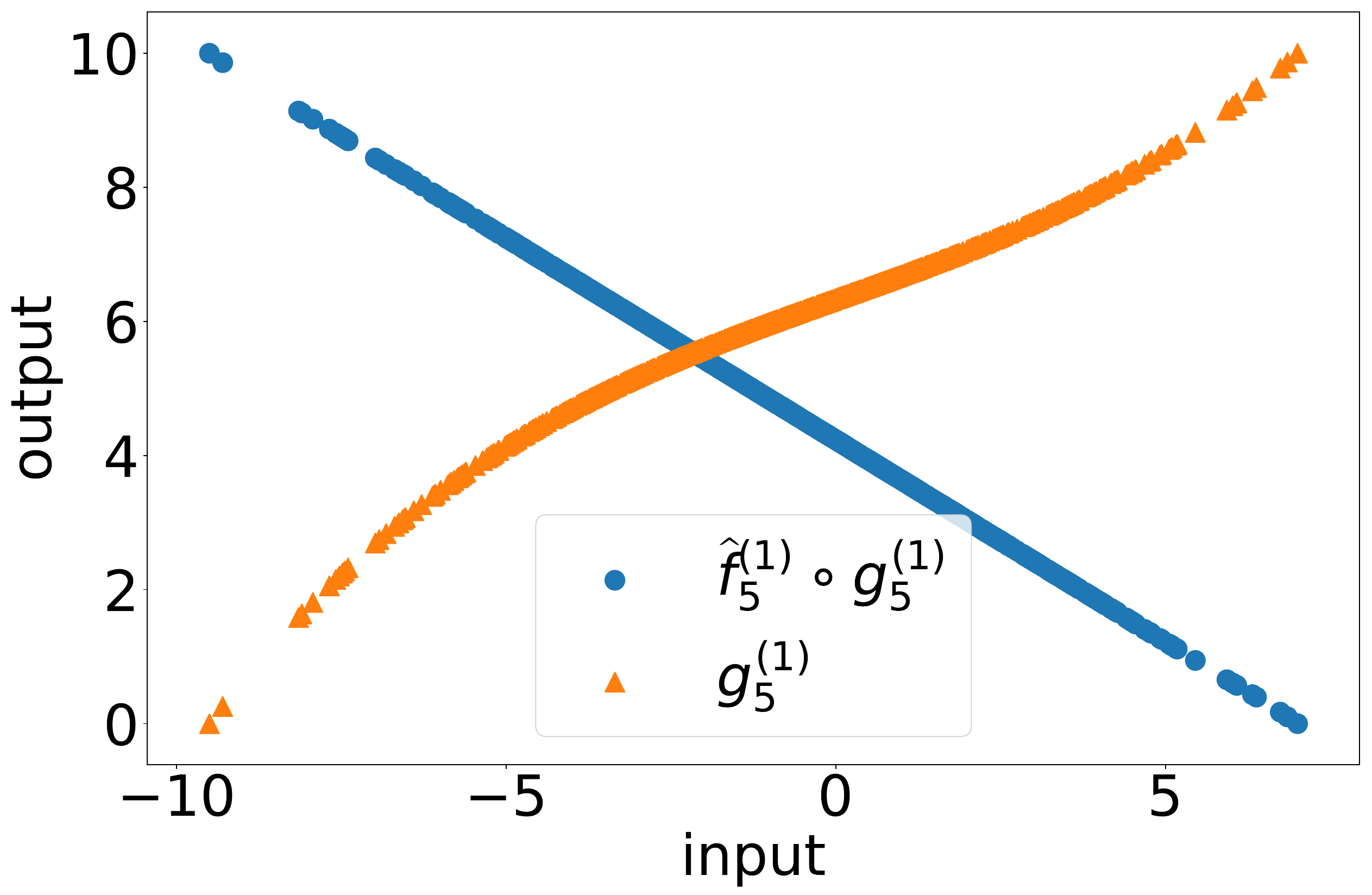}}
    \subfigure{\includegraphics[width=0.19\linewidth]{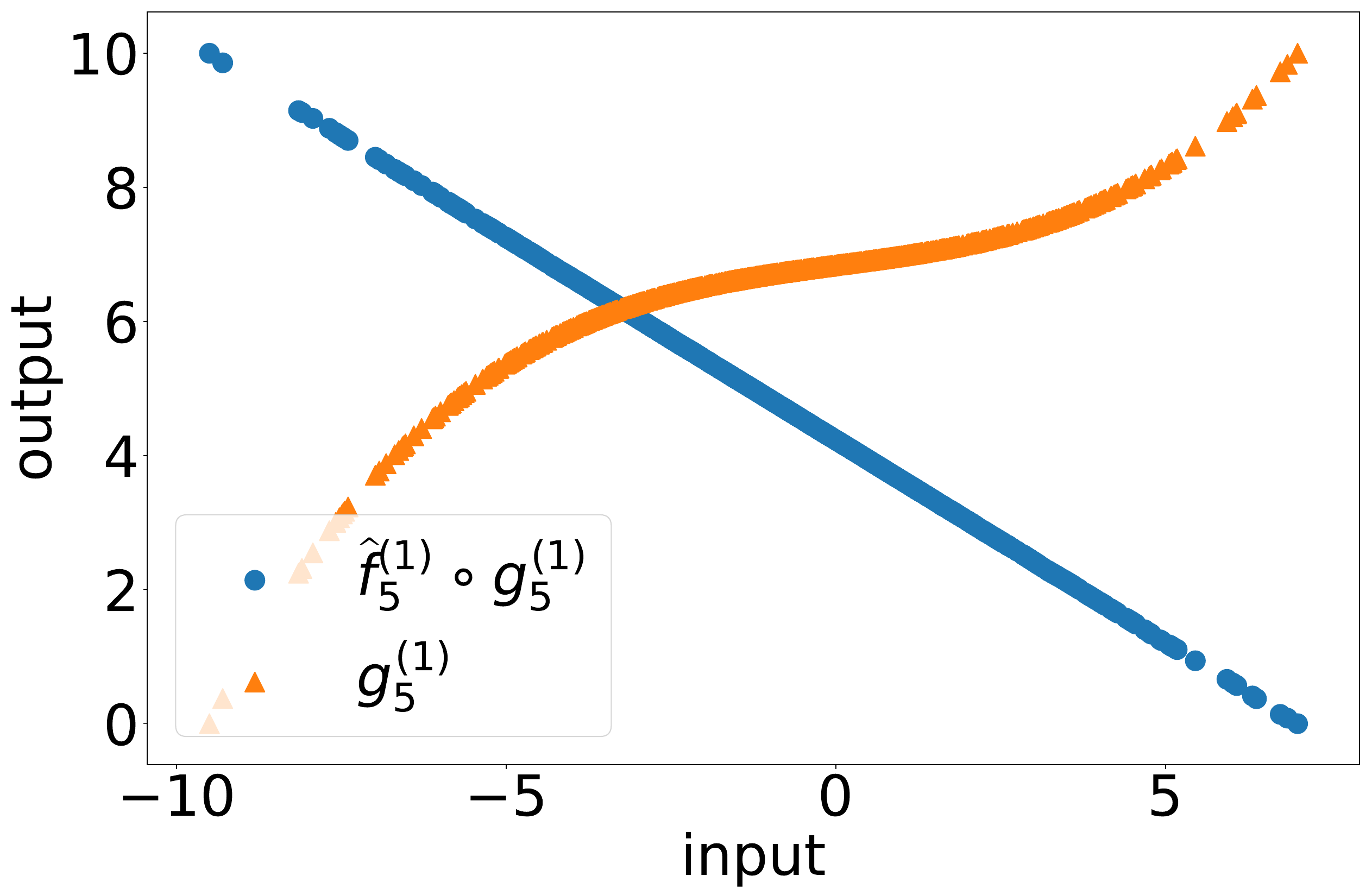}}
    \subfigure{\includegraphics[width=0.19\linewidth]{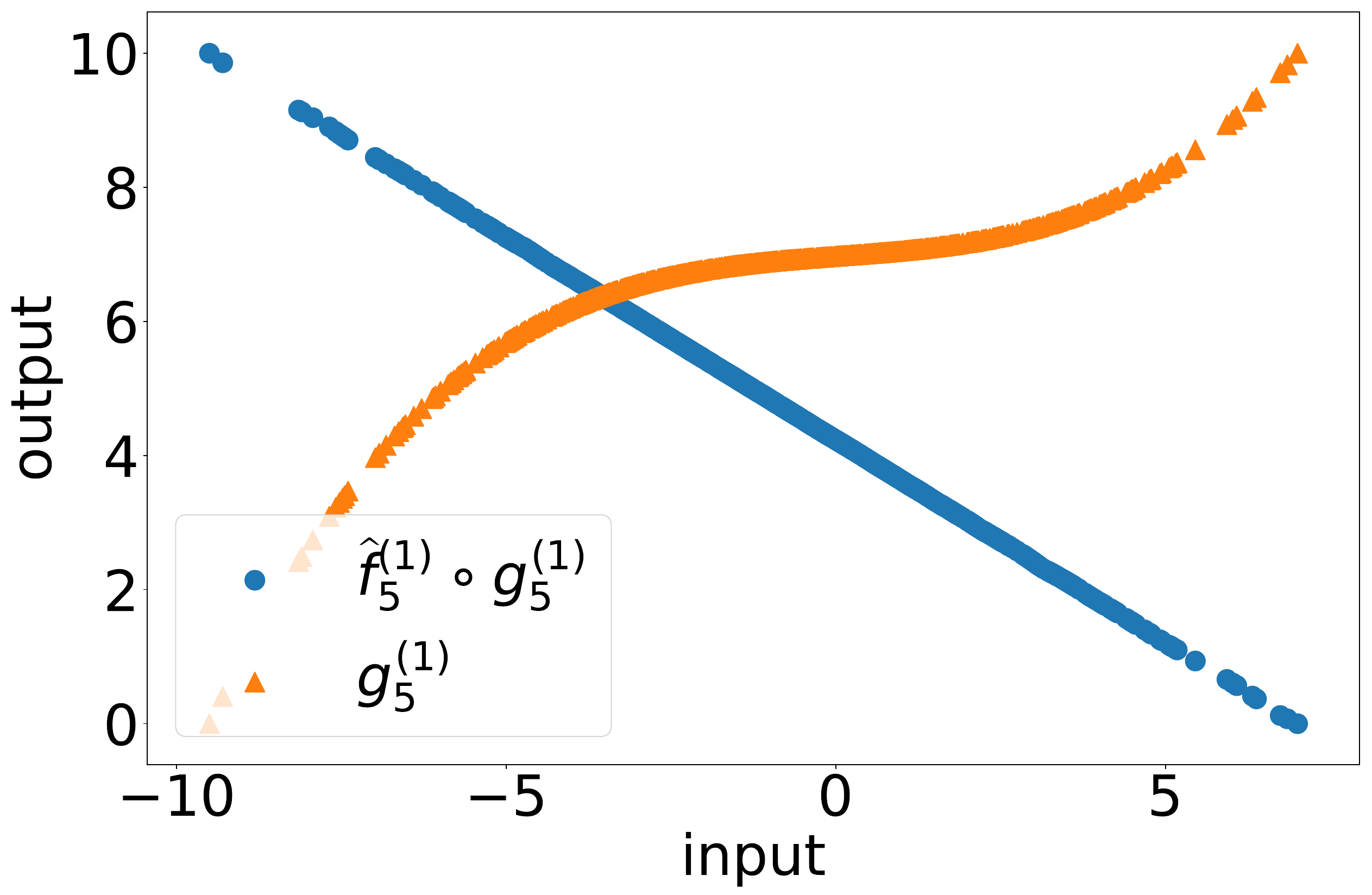}}
    \subfigure{\includegraphics[width=0.19\linewidth]{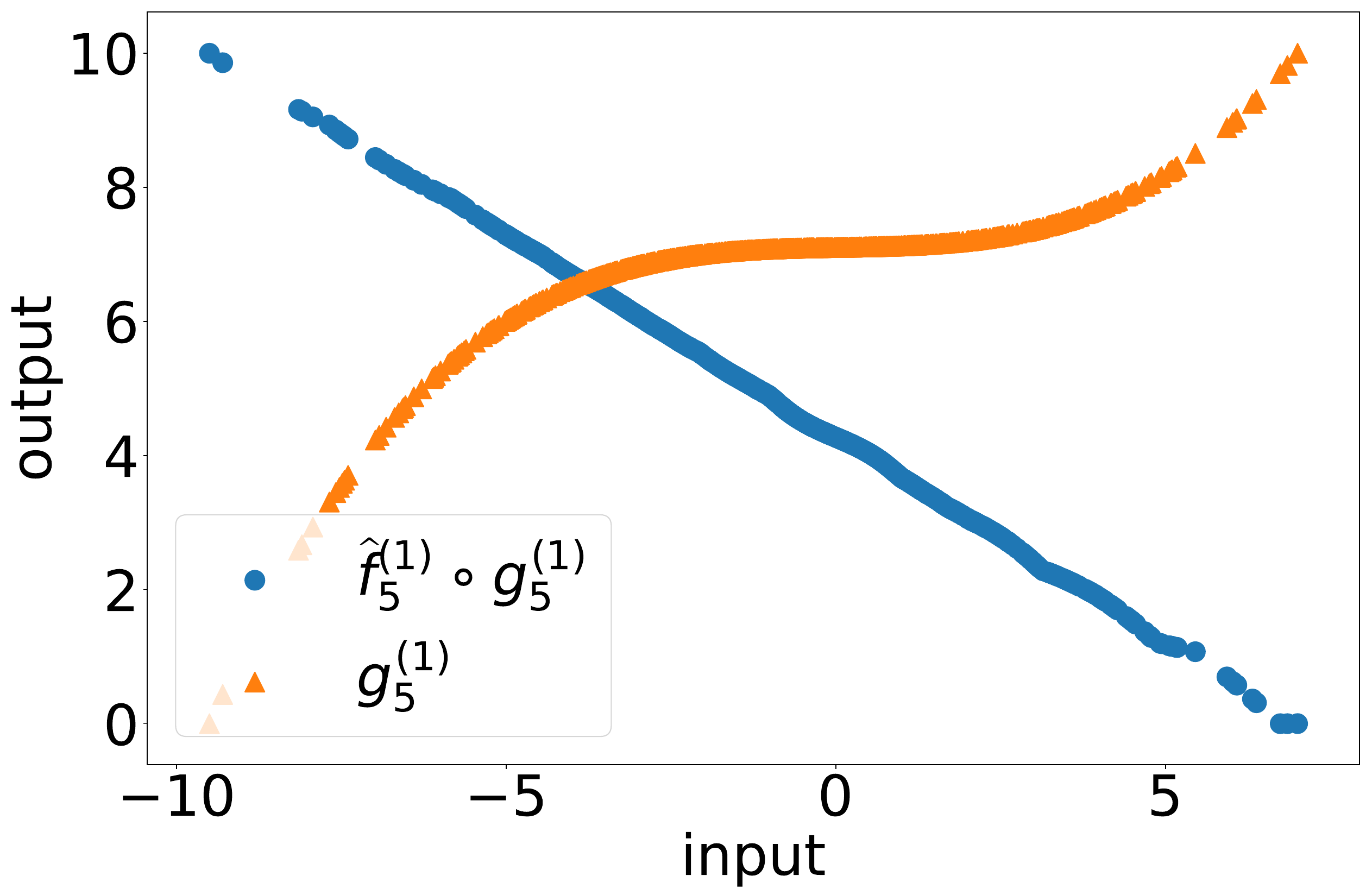}}
    \subfigure{\includegraphics[width=0.19\linewidth]{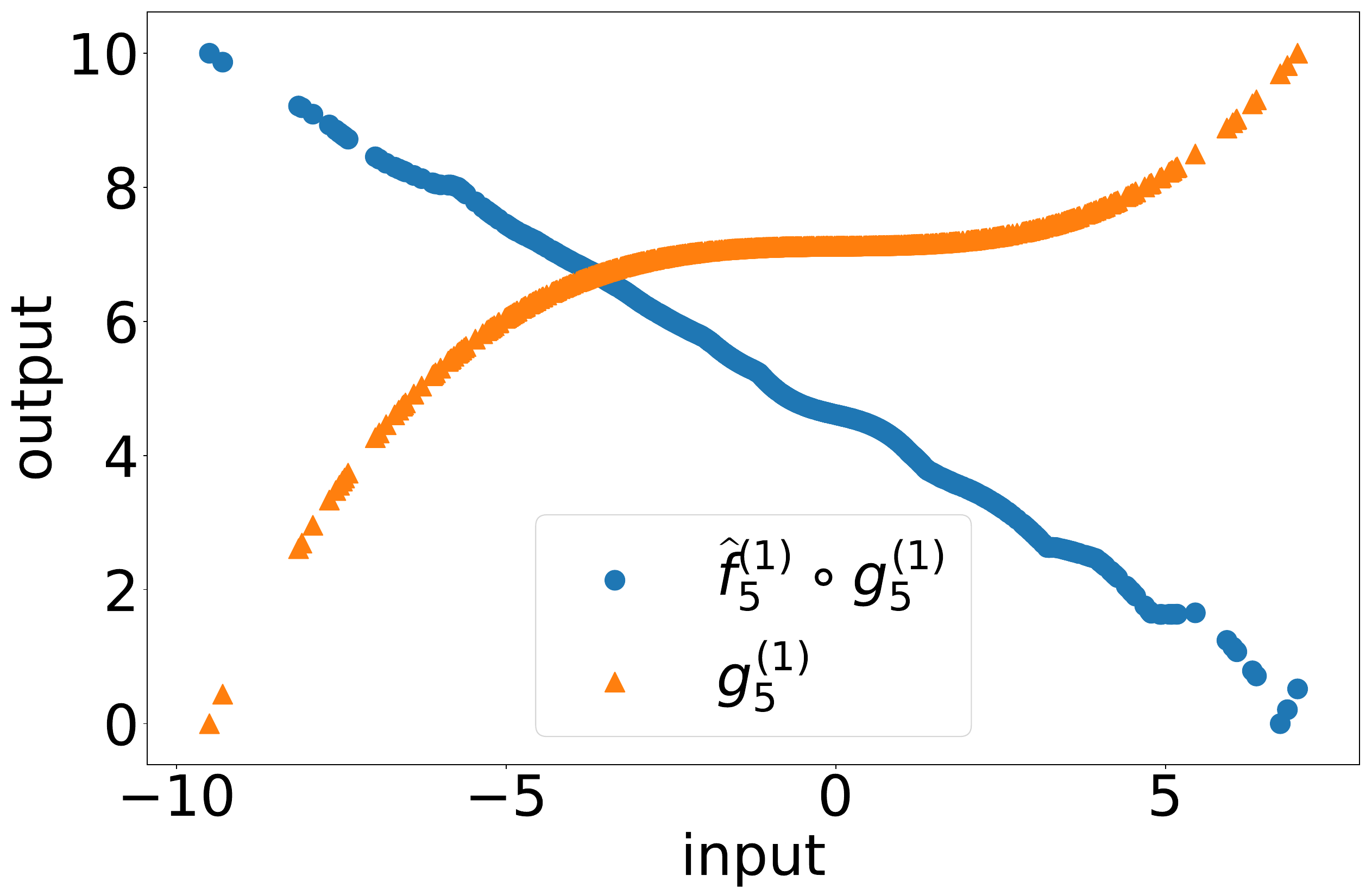}}
\caption{The nonlinear distortion and the learned composite functions of one example dimension. From left to right: $\alpha$ is 0.01, 0.05, 0.1, 0.5, 1 and the corresponding recovered subspace metric is 0.011, 0.014, 0.019, 0.158, 0.345}
\label{fig:diffalpha}
\vspace{-.5cm}
\end{figure*}

In Fig.~\ref{fig:diffalpha}, we test to what extent can the nonlinearity be solved by the proposed model. Specifically, we set the nonlinear function $g^{(q)}_i(x)=\alpha x^3+x$, then we try different values of $\alpha$, i.e., $\alpha=0.01, 0.05, 0.1, 0.5,$ and $ 1$. In this simulation, the shared components are zero-mean i.i.d. Gaussian variables and $K=3$, and the view-specific components for both views also follow this distribution with $R_1=R_2=2$. 
One can see that when $\alpha$ increases, the performance of the proposed approach indeed deteriorates, given the same computational resources, as expected. Nonetheless, even when $\alpha=0.5$, one can see that the nonlinearity imposed on the views is largely removed.

In Table~\ref{tab:dimension}, we present the results of the algorithms under different $K$, i.e., the dimension of the shared components.
We fix the observation dimension to be 5 and the $K$ ranges from 1 to 4. One can see that different $K$ settings essentially do not affect the results---which is consistent with CCA identifiability analyses in \cite{ibrahim2019cell} and this work.

\subsection{Real Data Experiments}
In this section, we test the proposed algorithm on two real datasets, namely, an EEG/MEG dataset and a multiview handwritten digits dataset.

\subsubsection{EEG/MEG Dataset}
This dataset is called Multi-modal Face Dataset\footnote{https://www.fil.ion.ucl.ac.uk/spm/data/mmfaces/}. It contains EEG and MEG measurements data of the same subjects when they are shown with real faces and scrambled faces. The task is to train a classifier using the EEG and MEG measurements to predict if the human subject sees a real face or a scrambled face.

The feature sizes of an EEG sample and an MEG sample are $M_1=$130 and $M_2=$151, respectively. There are 172 events during the experiment, among which 86 are real faces and {the rest are} scrambled faces. For each event, 161 snapshots are recorded for both EEG and MEG, respectively, resulting in $172\times 161=27,692$ samples for each view.
One could use the raw data to train classifiers, but the raw EEG/MEG signals are noisy, and are considered mixtures of components \cite{ziehe2000artifact,oveisi2012nonlinear}.
Also, nonlinear distortions have likely happened in the sensing process. This makes the dataset suitable for evaluating the proposed approach.

We train two linear classifiers, namely, the linear support vector machines (SVM) and the logistic regression classifier, based on the learned low-dimensional representations $\u_\ell$ for $\ell=1,\ldots,N$. The whole dataset is split into a training set, a validation set and a testing set. The validation set is employed to stop the learning procedures for algorithms. 
It is used for training the linear classifiers as well. 

Since the dimension of the input samples are large, we adopt fully connected networks to parameterize the nonlinear mappings, as we discussed in Remark~\ref{rmk:fully}. We use a three-layer network for ${\bm{f}_{\text{NN}}^{(q)}}$, each having 256 neurons. We use a single-layer network for ${\bm{g}_{\text{NN}}^{(q)}}$, which also has 256 neurons in the hidden layer. We set $\lambda=10^{-6}$. Dropout \cite{srivastava2014dropout} is also used in optimization. In this experiment, we use $K=5$ for our approach. This parameter is selected over \{2, 5, 10, 20, 50\} by tuning on the validation set. For PCA, the number of components is automatically selected so that 90\% energy is retrained. For DCCA, the network structure is the same as ours for ${\bm{f}_{\text{NN}}^{(q)}}$. DCCAE shares the same network structure of ours. All the results are averaged over 5 different random splits.

The results can be found in Table \ref{tab:eeg}.
One can see that using nonlinear dimensionality reduction techniques indeed improves upon the performance of classic techniques, which proves the necessity for taking nonlinear distortion into consideration.
In addition, the proposed approach is the most promising one among the baselines.
This is perhaps because the biological signal sensing problem fits our model well.

\begin{table}[]
\caption{Classification error rate (\%) on EEG/MEG data.}
\label{tab:eeg}
\centering
\resizebox{.85\linewidth}{!}{\Huge
\begin{tabular}{cc|ccccccc}
train/validation/test                  &          & Raw   & PCA   & CCA   & KCCA  & DCCA  & DCCAE & Proposed  \\ \hline \hline
\multirow{2}{*}{90\%/5\%/5\%}  & SVM      & 45.79 & 47.07 & 47.59 & 43.65 & 40.55 & 41.40 & 28.91 \\ 
                         & Logistic & 46.58 & 46.48 & 46.84 & 43.46 & 38.61 & 38.58 & 28.44 \\ \hline
\multirow{2}{*}{80\%/5\%/15\%} & SVM      & 43.70 & 41.47 & 40.79 & 39.99 & 39.43 & 40.06 & 32.20 \\ 
                         & Logistic & 41.73 & 42.29 & 41.30 & 40.67 & 40.47 & 41.01 & 31.19 \\ 
\end{tabular}}
\vspace{-.5cm}
\end{table}

\subsubsection{Multiview Digit Dataset}
This dataset from \cite{van1998handwritten} consists of features of handwritten digits (0 to 9) extracted from a collection of Dutch utility maps. There are 200 patterns per class (2,000 patterns in total) that have been digitized in binary images. These digits are represented in different views.

We employ three views for this experiment. 
The three views consist of 76 Fourier coefficients of the character shapes, 64 {Karhunen-Lo\`{e}ve} coefficients, and 47 Zernike moments, respectively.
The training set contains 1,200 samples, and the validation and testing sets both have 400 samples. To validate the performance of the learned representation, we use spectral clustering \cite{ng2002spectral} as the downstream task tool. After the models are learned, the testing set is fed to it and then clustering is performed on the output of the model.

All the settings are identical to those in the previous experiment, The latent dimension is selected over \{2, 5, 10, 20, 50\} by tuning on the validation set. Similarly, $\lambda$ is chosen over \{$1e^{-3}$, $1e^{-4}$, $1e^{-5}$, $1e^{-6}$\}.
The results are averaged over 5 different random splits. 

The clustering accuracy results on the testing set are shown in Table. \ref{tab:mfeat}. 
In the first three columns, we test the algorithms using two of the three views. In the last column, we test {our algorithm (since our approach can naturally handle more than two views) and GCCA} on all of the three views.
{Here, we employ the SUMCOR GCCA formulation and the algorithm in \cite{fu2016efficient}}.
Since KCCA, DCCA and DCCAE can only deal with two views, they do not have results in the last column. 
The observation is a bit different to what we have seen for the EEG/MEG experiment. 
The classic methods sometimes outperform the nonlinear baselines. This may be due to the fact that purely data-driven models may require more data to train.
The proposed approach again shows promising results. Note that although our approach has a similar implementation as DCCA and DCCAE, the subtle differences exist because we utilize the information of the underlying signal model. Hence, our method can be understood as a mixed model and data-driven approach. It appears that using model information could help in practice, especially when the amount of data is not very large.

\begin{table}[]
\caption{Spectral clustering ACC (\%) of different algorithms.}
\label{tab:mfeat}
\centering
\resizebox{.8\linewidth}{!}{\Huge
\begin{tabular}{c|cc|cc|cc|ccc}
      & View1            & View2           & View1            & View3           & View2            & View3           & View1         & View2        & View3        \\ \hline \hline
Raw  &    51.55     &   64.15    &    51.55    &   50.05     &   64.15      &  50.05      &    51.55  &   64.15  &  50.05   \\
PCA   &    52.10     &    65.75   & 52.10   &    51.05    &    65.75     &  51.05      & 52.10  &   65.75  &   51.05  \\ 
{(G)CCA}   &    69.45     &  65.55  &    72.70   &  68.05  &    68.60     &  64.15   & {62.85}         & {71.50}        & {66.20}        \\ 
KCCA  &     71.80    &  71.65  & 73.55 &    72.80    &   63.65      &    59.45    & -         & -        & -        \\ 
DCCA  &  72.90    & 72.10   &   69.70      &    72.00 &   60.55      &    62.15    & -         & -        & -        \\ 
DCCAE & 69.80   &   71.30  &    67.35    & 68.80       &  62.15       &   59.20     & -         & -        & -        \\ 
Proposed  & 73.40 & 72.90 &    71.15      &     71.05    &  70.35   &   66.15     &  74.20   &  77.05   &  73.60     
\end{tabular}
}
\vspace{-.5cm}
\end{table}

\section{Conclusion}
In this work, we revisited the multiview analysis problem under a nonlinear setting.
We showed that even under a class of unknown nonlinear distortions, the appealing identifiability guarantees for the classic linear multiview analysis still hold---if a proper identification criterion is used.
We also proposed a neural network-based implementation for the proposed criterion, as well as a BCD-based optimization algorithm.
A suite of numerical results were presented to corroborate our analysis. 
The proposed work also offers theoretical explanations to popular deep learning based canonical correlation analysis schemes, e.g., DCCAE and deep GCCA, from a model-based and identifiability-driven point of view.

Several future directions are in order: First, it is of great interest to study a model where different views do not share {exactly} the same latent components, while having similar ones. Such a model may be closer to real-world data. 
{Second, offering sample complexity for nonlinear multiview learning is also very meaningful.}
{Third}, it is well-motivated to study the shared component identifiability under more general nonlinear distortion, since PNL is a special class of nonlinear model. In some case, e.g., image classification problems as in our last experiment, the PNL model {may not be the most accurate}. Using a more general nonlinear model may lead to better performance.

\appendices

\section{Proof of Theorem 1}\label{app:thm2by2}

Let us denote $\bm h^{(q)}=\bm f^{(q)}\circ\bm g^{(q)}: \mathbb{R}^{2}\rightarrow \mathbb{R}^2$ for $q=1,2$ as the composition of the learned $\bm f^{(q)}$ and $\bm g^{(q)}$ in the generative model.
Hence, from Eq.~\eqref{eq:ncca}, we have
\begin{align*}
&b^{(1)}_1 h^{(1)}_1\left(A^{(1)}_{11}s+A^{(1)}_{12}c^{(1)}\right) + b^{(1)}_2 h^{(1)}_2\left(A^{(1)}_{21}s+A^{(1)}_{22}c^{(1)}\right)\\
&=b^{(2)}_1 h^{(2)}_1\left(A^{(2)}_{11}s+A^{(2)}_{12}c^{(2)}\right) + b^{(2)}_2 h^{(2)}_2\left(A^{(2)}_{21}s+A^{(2)}_{22}c^{(2)}\right)
\end{align*}
where we have omitted the subscript ``$\ell$'' for notational simplicity. Note that due to \eqref{eq:energy_popu}, the composition $h_i^{(q)}\neq 0$---and thus such trivial solutions cannot happen.

Denote $\phi(c^{(1)},c^{(2)},s)$ as
\begin{align}
&\phi(c^{(1)},c^{(2)},s)= \sum_{i=1}^2 b^{(1)}_i h^{(1)}_i\left(A^{(1)}_{i1}s+A^{(1)}_{i2}c^{(1)}\right)\nonumber\\
&\quad\quad\quad\quad-\sum_{j=1}^2 b^{(2)}_j h^{(2)}_j\left(A^{(2)}_{j1}s+A^{(2)}_{j2}c^{(2)}\right){=0.} \label{eq:phi_Eq}
\end{align}
To proceed, taking derivative w.r.t. $c^{(1)}$, we have $\frac{\partial\phi(c^{(1)},c^{(2)},s)}{\partial c^{(1)}}=0$, which means
$$
b^{(1)}_1 A^{(1)}_{12} (h^{(1)}_1)' + b^{(1)}_2 A^{(1)}_{22} (h^{(1)}_2)'=0.$$
Note that we have used the fact that $\partial c^{(2)}/\partial c^{(1)}=0$ and $\partial s/\partial c^{(1)}=0$.
To see this, first notice that {the equality $\phi(c^{(1)},c^{(2)},s)=0$} holds for all possible values of $[{s},{c}^{(1)},{c}^{(2)}]^\top$.
By the ubiquitously unanchored condition on $[{s},{c}^{(1)},{c}^{(2)}]^\top$, it implies that for any $\bar{c}^{(2)},\bar{s}$, all possible $c^{(1)}$'s satisfy the above equality---i.e., the {\it changing} of $c^{(1)}$ will not affect any $\bar{c}^{(2)},\bar{s}$ satisfying the equality. Hence, even if $c^{(2)}$ and $s$ are functions of $c^{(1)}$ (since they could be dependent),
we still always have $\partial c^{(2)}/\partial c^{(1)}=0$ and $\partial s/\partial c^{(1)}=0$ ({see more details in the supplementary material}).

By further taking the second-order derivatives {and invoking the ubiquitously unanchored assumption again}, we have $\frac{\partial^2 \phi(c^{(1)},c^{(2)},s)}{\partial (c^{(1)})^2}=0$ and $\frac{\partial^2\phi(c^{(1)},c^{(2)},s)}{\partial c^{(1)} \partial s}=0$, implying
\begin{align*}
&b^{(1)}_1 (A^{(1)}_{12})^2 (h^{(1)}_1)'' + b^{(1)}_2 (A^{(1)}_{22})^2 (h^{(1)}_2)''=0,\\
& b^{(1)}_1 A^{(1)}_{11}A^{(1)}_{12} (h^{(1)}_1)'' + b^{(1)}_2 A^{(1)}_{21}A^{(1)}_{22} (h^{(1)}_2)''=0.
\end{align*}

From the above, one can establish the following system of linear equations:
\begin{align*}
\underbrace{\begin{bmatrix} 
	b^{(1)}_1 A^{(1)}_{11}A^{(1)}_{12} & b^{(1)}_2 A^{(1)}_{21}A^{(1)}_{22}\\
	b^{(1)}_1 (A^{(1)}_{12})^2 & b^{(1)}_2 (A^{(1)}_{22})^2 
	\end{bmatrix}}_{\bm H}
\begin{bmatrix}
(h^{(1)}_1)''\\
(h^{(1)}_2)''
\end{bmatrix}=\bm{0}
\end{align*}

In order to prove that $h^{(1)}_i$'s are affine functions, we need to show that $(h^{(1)}_i)''=0$. In fact, it is sufficient to show that the left matrix $\bm H$ has full column rank. 
Since
we can re-express $\bm H$ in the following form
\begin{align}\label{eq:H_matrix}
\bm H=\begin{bmatrix} 
A^{(1)}_{11} & A^{(1)}_{12}\\
A^{(1)}_{21} & A^{(1)}_{22} 
\end{bmatrix}^\top
\begin{bmatrix} 
b^{(1)}_1 A^{(1)}_{12} & 0\\
0 & b^{(1)}_2 A^{(1)}_{22}
\end{bmatrix}{,}
\end{align}
{it} is readily seen that ${\rm rank}(\bm H)=2$. Indeed, for the left matrix, it is full rank almost surely since $\A^{(1)}$ is drawn from an absolutely continuous distribution. By the same reason, neither $A^{(1)}_{12}$ nor $A^{(1)}_{22}$ is zero with probability one. Hence, if $\|\bm b^{(1)}\|_0=2$, $\bm H$ has to be full rank.

To show $\|\bm b^{(1)}\|_0=2$, suppose that, on the contrary, {$b^{(1)}_1=0$}. One can see that on the left hand side of \eqref{eq:equal} we have
{$  {\sf LHS}=b^{(1)}_2 h^{(1)}_2\left(A^{(1)}_{21}s+A^{(1)}_{22}c^{(1)}\right)$}
while the right-hand side is a function of $s$ and $c^{(2)}$. Note that {$h^{(1)}_2=f^{(1)}_2\circ g^{(1)}_2$} is invertible. Hence, ${\sf LHS}$ has to be a function of $s$ and $c^{(1)}$. Otherwise, $b_2^{(1)}h_2^{(1)}$ cannot eliminate the contribution from $c^{(1)}$, which is a contradiction to the fact that \eqref{eq:equal} holds.
Therefore, the product of these two matrices {on the right hand side of Eq.~\eqref{eq:H_matrix}} is still full rank which leads to the fact that $(\bm{h}^{(1)})''=\bm{0}$. 
By symmetry, one can readily show that $(h^{(2)}_i)''=0$. 

As a final step, we also need to show that $\alpha_i^{(q)}\neq 0$. This is trivial since \eqref{eq:invert} holds. That is,  $h_i^{(q)}(x)=  f_i^{(q)}\circ g_i^{(q)}(x) $ is a composition of two invertible function---thereby being invertible itself. This excludes the possibility of $\alpha_i^{(q)}$ being zero---which would have made $h_i^{(q)}(\cdot)$  non-invertible.

	\section{Proof of Proposition~\ref{prop:zeronorm}}\label{app:prop}
	To show this proposition, it suffices to construct a solution such that $\B^{(q)}$ has no zero elements. Let $\bm f^{(q)}\circ\bm g^{(q)}=\bm I$. Then, one solution of $\bm B^{(q)}$ is
	$\B^{(q)} = [\bm \Theta,\bm 0](\A^{(q)})^\dagger.$
	The above solution is nothing but selecting out some columns of $(\A^{(q)})^\dagger$.
	Recall that we have assume that $\A^{(q)}$ is drawn from a certain joint absolutely continuous distribution, which means that $(\A^{(q)})^\dag$ also follows a certain absolutely continuous distribution \cite{olkin1998density}---thereby the elements of $\bm B^{(q)}$.
	Hence, we have
	$     {\sf Prob}\left(\bm B^{(q)}(i,j)=0\right)=0$.

\section{Proof of Theorem 2} \label{app:thm}
We show the case that $M_1=K+R_1$ and $M_2=K+R_2$, i.e., the mixing matrices $\A^{(1)}$ and $\A^{(2)}$ are square nonsingular matrices. The case where $M_q>K+R_q$ can then be shown via a straightforward approach: The $M_q$ observations output by $\bm g^{(q)}(\cdot)$ can be divided to (possibly overlapped) groups, each of which has $K+R_q$ channels.

It suffices to show that the nonlinearity can be removed by considering any row of $(\B^{(q)})^\T$, denoted as $(\bm b^{(q)})^\T$.
In this case, we consider only extracting one shared component. However, since we further restricted that the extracted components are uncorrelated [cf. the constraint in Eq.~\eqref{eq:energy_popu}], the $K$ extracted components by $\B^{(q)}$ will not have repeated ones.

From Eq.~\eqref{eq:ncca}, we have the following equation
\begin{align*}
&\sum_{m=1}^{M_1} b^{(1)}_m h^{(1)}_m\left(\sum_{k=1}^K A^{(1)}_{m,k}s_k+ \sum_{r=1}^{R_1} A^{(1)}_{m,K+r}c^{(1)}_r\right)\\
&=\sum_{j=1}^{M_2} b^{(2)}_j h^{(2)}_j\left(\sum_{l=1}^K A^{(2)}_{j,l}s_l+ \sum_{s=1}^{R_2} A^{(2)}_{j,K+s}c^{(2)}_s\right).
\end{align*}

Denote $\phi(\bm{c}^{(1)},\bm{c}^{(2)},\bm{s})$ as
\begin{align*}
&\phi(\bm{c}^{(1)},\bm{c}^{(2)},\bm{s})=\\
&\quad\sum_{m=1}^{M_1} b^{(1)}_m h^{(1)}_m\left(\sum_{k=1}^K A^{(1)}_{m,k}s_k+ \sum_{k=1}^{R_1} A^{(1)}_{m,K+k}c^{(1)}_k\right)\\
&\quad -\sum_{j=1}^{M_2} b^{(2)}_j h^{(2)}_j\left(\sum_{l=1}^K A^{(2)}_{j,l}s_l+ \sum_{l=1}^{R_2} A^{(2)}_{j,K+l}c^{(2)}_l\right){=0.}
\end{align*}

Let us first consider $\bm{h}^{(1)}$. Taking derivatives {on both sides of the above, we have} $\frac{\partial^2\phi(\bm{c}^{(1)},\bm{c}^{(2)},\bm{s})}{\partial c^{(1)}_1 \partial s_k}=0,~k=1,\ldots,K$---{which} leads to $K$ linear equations w.r.t. $(h_m^{(1)})''$ for $m=1,\ldots,M_1$. In addition $\frac{\partial^2\phi(\bm{c}^{(1)},\bm{c}^{(2)},\bm{s})}{\partial c^{(1)}_1 \partial c^{(1)}_r}=0,~r=1,\ldots,R_1,$ yields another $R_1$ linear equations. In total, we have collected $K+R_1$ linear equations as follows:
\begin{align*}
\sum_{m=1}^{M_1} &b^{(1)}_m A^{(1)}_{m,k} A^{(1)}_{m,K+1} (h^{(1)}_m)''=0,\ \text{for }k=1,...,K+R_1.
\end{align*}
{In the above derivation, we have used the assumption that $[{\bm s}_\ell^\top, ({\bm c}^{(1)}_\ell)^\top, ({\bm c}^{(2)}_\ell)^\top]^\top$ is ubiquitously unanchored, and thus all the ``cross-derivatives'' are zeros.}

It can be rewritten as the following linear system:
\begin{align*}
(\bm{A}^{(1)})^\T&\begin{bmatrix} 
b^{(1)}_1 A^{(1)}_{1,K+1} & &\\
&\ddots&\\
&& b^{(1)}_{M_1} A^{(1)}_{M_1,K+1} 
\end{bmatrix}
\begin{bmatrix}
(h^{(1)}_1)''\\
\vdots\\
(h^{(1)}_{M_1})''
\end{bmatrix}=\bm{0},
\end{align*}
which holds for all $\bm s_\ell \in{\cal S}$ and $\bm c_\ell^{(q)}\in{\cal C}_q$,
where $\bm{A}^{(1)}\in \mathbb{R}^{M_1\times M_1}$ is the mixing matrix, which is full rank since it is drawn from an absolutely continuous distribution.

Now what remains is to show that the diagonal matrix in the middle is full rank which is equivalently that all the diagonal elements are non-zero. Since $\bm{A}^{(1)}$ is assumed to be continuously random variable, the probability that any element of it being zero is zero. 

Therefore, we only need to show that $b^{(1)}_1,\cdots,b^{(1)}_{M_1}$ are not zeros. By the assumption that $\|\bm B^{(q)}\|_0=KM_q$, none of $\bm{b}^{(1)}$ will be zero. As a result, the matrix in the middle is full rank. Consequently, $(\bm{h}^{(1)})''$ has to be a all-zero vector.
In addition, Proposition~\ref{prop:zeronorm} guarantees that there exists such a solution $\bm b^{(1)}$.
By applying the same proof to the second view, one can show that $\bm h^{(2)}$ is also an affine mapping.

\section{Proof of Theorem~\ref{thm:dependent}}\label{app:dependent}
{For the same $\phi(\bm{c}^{(1)},\bm{c}^{(2)},\bm{s})$ defined in previous theorems, by taking second order derivative $\frac{\partial^2 \phi(\bm{c}^{(1)},\bm{c}^{(2)},\bm{s})}{\partial c^{(1)}_i c^{(1)}_j}$, we have $\frac{R_1(R_1+1)}{2}$ linear equations:
\begin{align}\label{eq:neweq}
\sum_{m=1}^{M_1} &b^{(1)}_m A^{(1)}_{m,K+i} A^{(1)}_{m,K+j} (h^{(1)}_m)''=0,
\end{align}
for $i,j=1,\cdots,R_1$ and $i\leq j$. Note that since $s_k$ and $s_j$ could be fully dependent, we do not take cross-derivatives involving $s_k$ or $s_j$. The reason is that $\frac{\partial s_k}{\partial s_j}\neq 0$ could happen if $s_k$ and $s_j$ are completely dependent (or, if $[s_k,s_j]^\T$ is not ubiquitously unanchored).
Eq.~\eqref{eq:neweq} can be re-expressed as:
\begin{align*}
\underbrace{\begin{bmatrix}
    \left(\bm{a}_{K+1}^{(1)} \circledast \bm{a}_{K+1}^{(1)} \right)^\top\\
    \vdots\\
    \left(\bm{a}_{K+R_1}^{(1)} \circledast \bm{a}_{K+R_1}^{(1)}  \right)^\top\\
    \left(\bm{a}^{(1)}_{K+1}\circledast \bm{a}^{(1)}_{K+2}\right)^\top\\
    \vdots\\
    \left(\bm{a}^{(1)}_{K+R_1-1}\circledast\bm{a}^{(1)}_{K+R_1}\right)^\top\\
\end{bmatrix}}_{\widetilde{\A}}
\begin{bmatrix} 
b^{(1)}_1 & &\\
&\ddots&\\
&& b^{(1)}_{M_1}  
\end{bmatrix}
\begin{bmatrix}
(h^{(1)}_1)''\\
\vdots\\
(h^{(1)}_{M_1})''
\end{bmatrix}=\bm{0},
\end{align*}
where $\circledast$ denotes Hadamard product. Let $\bm{a}^{(1)}_i$ denote the $i$th column of $\bm{A}^{(1)}$.
We need to show that $\widetilde{\A}$ has full column rank---i.e., there are $M_1$ rows that are linearly independent. 

{Since $\widetilde{\A}$ is a random matrix, it suffices to show that one of the $M_1\times M_1$ submatrices admits nonzero determinant somewhere---which implies that the determinant is nonzero everywhere, almost surely \cite{caron2005zero}. To this end,} consider a special case where $\A^{(1)}$ is a Vandermonde matrix, i.e.,
$\bm{a}^{(1)}_i = [1, x_i, x_i^2, \cdots, x_i^{M_1-1}]^\top,$
and $x_i\neq x_j$.
{Let us consider any} $R$ columns of the $R_1$ columns from $\bm{a}_{K+1}^{(1)},\ldots,\a_{K+R_1}^{(1)}$.
Note that we require $R\leq R_1$ to satisfy $M_1=\lfloor\frac{R(R+1)}{2}\rfloor$ (for notational simplicity, we assume $M_1=\frac{R(R+1)}{2}$ in the sequel). Hence, the corresponding rows in the matrix of interest {exhibit} the following form:
\begin{align*}
    \begin{bmatrix}
        1 & x_{1}^2  & \cdots & x_{1}^{2(M_1-1)}\\
        \cdots&\cdots&&\cdots \\
        1 & x_{R}^2 & \cdots & x_{R}^{2(M_1-1)}\\
        1 & x_{1}x_{2} & \cdots & (x_{1}x_{2})^{M_1-1}\\
        \cdots&\cdots&&\cdots\\
        1 & x_{R-1}x_{R} & \cdots & (x_{R-1}x_{R})^{M_1-1}
    \end{bmatrix}
\end{align*}
Note that one can always construct such a sequence---e.g., $x_{1}=1, x_{2}=1.1, x_{3}=1.11, \cdots$---to make sure that the matrix above is full rank. 
This means {that the determinant of the above matrix is nonzero somewhere}. Hence, the matrix $\widetilde{\A}$ has full column rank almost surely \cite{caron2005zero}. This also means that $(h^{(1)}_i)''=0$ for all $i=1,\ldots,M_1$. The same proof applies to $h^{(2)}_i$ for $i=1,\ldots,M_2$. }

\section{Proof of Lemma~\ref{lem:UZ}}\label{app:lem}
First, notice that the rows of $\bm U$ live in the null space of $\bm 1$. Hence, we can re-express $\bm U$ by 
$\bm{U}=\bm{VW},$ where $\bm{W}=\bm{I}-\frac{1}{N}\bm{11}^\top$. Hence, Problem~\eqref{eq:UZ} can be written as the following equivalent problem:
\begin{align*}
\minimize_{\bm V}&~\|\bm V \bm W-\bm Z\|_F^2\\
{\rm subject~to}&~(1/N)\bm V \bm W \bm{W}^\top\bm V^\T =\bm I.
\end{align*}

By expanding the above, we have:
\begin{align*}
\minimize_{\bm{V}}&~\| \bm{VW}\|_F^2-2\tr (\bm{VWZ}^\top)+\|\bm{Z} \|_F^2\\
{\rm subject~to}&~(1/N)\bm V \bm W \bm{W}^\top\bm V^\T =\bm I.
\end{align*}

The last term $\|\bm{Z}\|_F^2$ is a constant, it can be replaced with $\|\bm{ZW}\|_F^2$. The middle term also can be written as $-2\tr (\bm{VWW}^\top\bm{Z}^\top)$ since $\bm{W}$ is an orthogonal complement projector with $\bm{W}=\bm{W}^\top$ and $\bm{W}=\bm{W}^2$. Thus, we have the reformulated problem:
\begin{align*}
\minimize_{\bm{V}}&~\| \bm{VW}\|_F^2-2\tr (\bm{VWW}^\top\bm{Z}^\top)+\|\bm{ZW} \|_F^2\\
{\rm subject~to}&~(1/N)\bm V \bm W \bm{W}^\top\bm V^\T =\bm I.
\end{align*}

Plugging $\bm{U}=\bm V\bm W$ back, we have:
\begin{align*}
\minimize_{\bm U}&~\|\bm U-\bm Z \bm W\|_F^2\\
{\rm subject~to}&~(1/N)\bm U \bm U^\T =\bm I.
\end{align*}

An optimal solution can be obtained via invoking the \textit{Procrustes projection}; i.e., a solution is simply taking SVD of $\bm{ZW}=\bm{PDQ}^\top$ and let $\bm{U}=\sqrt{N}\bm{PQ}^\top$.


\bibliographystyle{IEEEtran}

\clearpage

\begin{center}
	{\bf Supplementary Materials for ``Nonlinear Multiview Component Analysis: Identifiability and Neural Network-Assisted Implementation''}
\end{center}

\begin{center}
Qi Lyu and Xiao Fu
\end{center}

\section{Proof of linear CCA identifiability}\label{app:linearcca}
Since the theorem in \cite{ibrahim2019cell} is an important component to our proof,
we restate the theorem and its proof in \cite{ibrahim2019cell} using the notations in this paper, for ease of exposition.
\begin{Theorem}[Linear CCA Identifiability] Assume that the model in \eqref{eq:salah} holds. Denote $\S=[\s_1,\ldots,\s_N]$ and $\C^{(q)}=[\c^{(q)}_1,\ldots,\c^{(q)}_N]$ for $q=1,2$.
    Also assume that matrix $\bm{T}=[\bm{S}^\top,(\bm{C}^{(1)})^\top,(\bm{C}^{(2)})^\top]\in \mathbb{R}^{N\times (K+R_1+R2)}$ has full column rank, and that $\A^{(q)}\in \mathbb{R}^{M_q\times (K+R_q)}$ has full column rank. Then the range space spanned {by the} shared components, i.e., ${\rm range}(\bm{S}^\top)$, can be identified via solving the linear CCA problem in \eqref{eq:cost_cca}.
\end{Theorem}

\begin{IEEEproof}
According to our generative model, in the linear case,
\begin{align}\label{eq:linear_generative}
\y_\ell^{(q)} &=\A^{(q)}[\s_\ell^\T,(\c^{(q)}_\ell)^\T]^\T,~\ell=1,\ldots,N,
\end{align}
{Note that equation above is exactly the same as model Eq.~\eqref{eq:salah} in \cite{ibrahim2019cell}}.

First consider the vector case where $s_\ell \in \mathbb{R}$. In this case, $\bm{B}^{(q)}$ becomes a vector $\bm{b}^{(q)}$. By solving the problem in \eqref{eq:cost_cca}, we have the following when the cost reaches zero:
\begin{align*}
    (\bm{Y}^{(1)})^\top\bm{b}^{(1)}=(\bm{Y}^{(2)})^\top\bm{b}^{(2)}.
\end{align*}
Without loss of generality, we can set $\bm{b}^{(q)}=\A^{(q)}\left((\A^{(q)})^\top\A^{(q)}\right)^{-1}\bm{r}^{(q)}$ where $\bm{r}^{(q)}$ is a vector that is in $\mathbb{R}^{1+R_q}$. This is because that the $\bm{b}^{(q)}$ vector can always be decomposed into the orthogonal projection on $\A^{(q)}$ and on its complement projection as well. The complement one will always be canceled since there is a $(\bm{A}^{(q)})^\top$ term in $(\bm{Y}^{(q)})^\top$. Consequently, we have
\begin{align*}
    [\bm{s}^\top, (\bm{C}^{(1)})^\top]\bm{r}^{(1)}=[\bm{s}^\top, (\bm{C}^{(2)})^\top]\bm{r}^{(2)},
\end{align*}
which can be rewritten as:
\begin{align*}
    &[\bm{s}^\top, (\bm{C}^{(1)})^\top, (\bm{C}^{(2)})^\top]
    \begin{bmatrix}
        \bm{r}^{(1)}\\
        \bm{0}_{R_2}
    \end{bmatrix}\\
    &\quad \quad=[\bm{s}^\top, (\bm{C}^{(1)})^\top, (\bm{C}^{(2)})^\top]
    \begin{bmatrix}
        \bm{r}^{(2)}(1)\\
        \bm{0}_{R_1}\\
        \bm{r}^{(2)}(2:\text{end})
    \end{bmatrix}
\end{align*}
where $\bm{0}_{R_1}$ is all-zero vector with size $R_1$ and $\bm{0}_{R_2}$ is that of size $R_2$.
Rearranging terms, we have:
\begin{align*}
    \bm{Tr}=\bm{0}
\end{align*}
where $\bm{T}=[\bm{s}^\top, (\bm{C}^{(1)})^\top, (\bm{C}^{(2)})^\top]\in \mathbb{R}^{N\times (1+R_1+R2)}$ and {$\bm{r}=[(\bm{r}^{(1)}(1)-\bm{r}^{(2)}(1))^\top, \bm{r}^{(1)}(2:\text{end})^\top, -\bm{r}^{(2)}(2:\text{end})^\top]^\top$} where $\bm{r}^{(1)}(2:\text{end})$ containing all except the first entries in $\bm{r}^{(1)}$.

It is obvious that if $\bm{T}$ has full column rank, then $\bm{r}$ has to be an all-zero vector which means that $\bm{r}^{(1)}=c\bm{e}_1$ and $\bm{r}^{(2)}=c\bm{e}'_1$ where $\bm{e}_1$ and $\bm{e}'_1$ are the first columns of identity matrices of different sizes, respectively. As a result, the shared component is identified as $\pm \bm{s}/\|\bm{s}\|_2$.

For the $K>1$ case, let $\bm{B}^{(q)}=\A^{(q)}\left((\A^{(q)})^\top\A^{(q)}\right)^{-1}\bm{R}^{(q)}$. Similarly, define
\begin{align*}
    \bm{R}=\begin{bmatrix}
    \bm{R}^{(1)}(1:K,:)-\bm{R}^{(2)}(1:K,:) \\
    \bm{R}^{(1)}(K+1:\text{end},:) \\
    -\bm{R}^{(2)}(K+1:\text{end},:)
\end{bmatrix}\in\mathbb{R}^{(K+R_1+R_2)\times K}
\end{align*}
where $\bm{R}^{(1)}(1:K,:)$ denotes rows 1 to $K$ of $\bm{R}^{(1)}$. We have the following holds:
\begin{align*}
    \bm{TR}=\bm 0.
\end{align*}
If $\bm{T}$ has full column rank, then the solution is unique with $\bm{R}^{(1)}(1:K,:)=\bm{R}^{(2)}(1:K,:)=\bm \Theta$, $\bm{R}^{(1)}(K+1:\text{end},:)=\bm{0}$ and $\bm{R}^{(2)}(K+1:\text{end},:)=\bm{0}$. 
This means that the solution is
\[   \B^{(q)}\bm Y^{(q)} =\bm \Theta\S,~q=1,2, \]
which completes the proof.
\end{IEEEproof}

One remark here is that if $\bm{s}_\ell, \c^{(q)}_\ell$ are drawn from any jointly continuous distribution and the condition $N\geq R_1+R_2+K$ holds, then the matrix $\bm{R}$ has full column rank almost surely.

\section{The Ubiquitously Unanchored Condition and Cross-Derivatives}\label{app:uu}
The key consequence of having $v_i$'s to be ubiquitously unanchored is that the cross-derivatives such as $\partial v_i/{\partial v_j}$ are all zeros under this condition.

Recall that the derivative of a continuous function $\phi(x)$ is defined as follows:
\begin{align*}
    \varphi^\prime (x)= \lim_{\Delta x\rightarrow 0} \frac{\varphi(x+\Delta x)-\varphi(x)}{\Delta x}.
\end{align*}

Let us treat $v_j$ as a function of $v_i$ (since the two could be dependent). Consider the partial derivative as follows:
\begin{align*}
    \frac{\partial v_j}{\partial v_i}= \lim_{\Delta v_i\rightarrow 0}\frac{v_j(v_i+\Delta v_i)-v_j(v_i)}{\Delta v_i}
\end{align*}
where the notation $v_j(v_i)$ means that $v_j$ is a function of $v_i$. 

Suppose that the components of $\bm{v}$ are ubiquitously unanchored. Hence, for any $v_i$, any fixed $\bar{v}_j$ can appear together with it in the domain of ${\bm v}$. Let us denote ${\cal V}(v_i,\bar{v}_j)\subseteq {\cal V}$ as the set of $\bm v$ such that $[\bm v]_i = v_i$ and $[\bm v]_j=\bar{v}_j$. Over this subset, we have
\begin{align*}
    v_j(v_i+\Delta v_i)=\bar{v}_j,\quad v_j(v_i)=\bar{v}_j,~\forall \bm v\in {\cal V}(v_i,\bar{v}_j)
\end{align*}
for all possible $\Delta v_i$. Hence, the above leads to
\begin{align*}
    \frac{\partial v_j}{\partial v_i} &= \lim_{\Delta v_i\rightarrow 0}\frac{v_j(v_i+\Delta v_i)-v_j(v_i)}{\Delta v_i}\\
    & = \lim_{\Delta v_i\rightarrow 0}\frac{\bar{v}_j-\bar{v}_j}{\Delta v_i}\\
    & = 0,\quad ~\forall \bm v\in {\cal V}(v_i,\bar{v}_j),
\end{align*}
for all $j\neq i$. 
Repeating the above for all possible $v_i$ and $\bar{v}_j$, one can reach the conclusion that $ \frac{\partial v_j}{\partial v_i} =0$.

\end{document}